\documentclass[manuscript,screen]{acmart}

\usepackage{graphicx}
\usepackage{booktabs}
\usepackage{multirow}
\usepackage{amsmath}
\usepackage{bbold}
\usepackage{enumitem}
\usepackage{hyperref}
\usepackage{subcaption}

\DeclareGraphicsExtensions{.png,.pdf,.eps}

\newcommand{\Quote}[1]{``\textit{#1}''}

\AtBeginDocument{%
  \providecommand\BibTeX{{%
    \normalfont B\kern-0.5em{\scshape i\kern-0.25em b}\kern-0.8em\TeX}}}

\begin{document}

\title{A Survey on Graph Counterfactual Explanations: Definitions, Methods, Evaluation, and Research Challenges}

\author{Mario Alfonso Prado-Romero}
\email{marioalfonso.prado@gssi.it}
\orcid{0000-0002-0491-3515}
\affiliation{
  \institution{Gran Sasso Science Institute}
  \city{L'Aquila}
  \country{Italy}
  \postcode{67100}
}

\author{Bardh Prenkaj}
\email{prenkaj@di.uniroma1.it}
\orcid{0000-0002-2991-2279}
\affiliation{
    \institution{Sapienza University of Rome}
    \city{Rome}
    \country{Italy}
    \postcode{00198}
}

\author{Giovanni Stilo}
\orcid{0000-0002-2092-0213}
\email{giovanni.stilo@univaq.it}
\affiliation{
    \institution{University of L'Aquila}
    \city{L'Aquila}
    \country{Italy}
    \postcode{67100}
}

\author{Fosca Giannotti}
\orcid{0000-0003-3099-3835}
\email{fosca.giannotti@sns.it}
\affiliation{
    \institution{Scuola Normale Superiore}
    \city{Pisa}
    \country{Italy}
    \postcode{56126}
}


\begin{abstract}
Graph Neural Networks (GNNs) perform well in community detection and molecule classification. Counterfactual Explanations (CE) provide counter-examples to overcome the transparency limitations of black-box models. Due to the growing attention in graph learning, we focus on the concepts of CE for GNNs. We analysed the SoA to provide a taxonomy, a uniform notation, and the benchmarking datasets and evaluation metrics. We discuss fourteen methods, their evaluation protocols, twenty-two datasets, and nineteen metrics. We integrated the majority of methods into the GRETEL library to conduct an empirical evaluation to understand their strengths and pitfalls. We highlight open challenges and future work.
\end{abstract}



\maketitle

\section{Introduction}\label{sec:intro}
Explaining predictions is crucial for enabling trusted decision-making in sensitive domains for both users and service providers \cite{guidotti2018survey}. However, the prevalent use of deep neural networks in generating predictions has given rise to the issue of the \Quote{black box} problem \cite{petch2021opening}. These complex models hinder the comprehension of the decision-making process employed for forecasting outcomes. Due to their reliance on nonlinear activations for learning feature representations, deep neural networks remain opaque to users, resulting in limited adoption in critical domains like health and finance.
In contrast, there are \Quote{white box} or transparent models, which allow for easy inspection of their inner processes \cite{8882211}. Although white-box models are preferred for decision-making \cite{verenich2019predicting}, black-box models outperform them in scenarios involving high-dimensional data, such as protein relation prediction \cite{madeddu2020feature}, student dropout prediction \cite{prenkaj2020reproducibility,prenkaj2021hidden}, and trend forecasting \cite{verma2022temporal}.

Recently, Graph Neural Networks (GNNs) \cite{scarselli2008graph} have emerged as a promising solution to various graph mining tasks, such as vertex classification \cite{pareja2020evolvegcn,wang2021semi,zhao2021graphsmote} and link prediction \cite{welling2016semi,you2021identity}. GNNs take as input a graph composed of vertices and edges. Examples of graphs are Facebook's user friendship network and PPIs that collect proteins and their relations. In these scenarios, the vertices are users (proteins), and the edges are the friendships (chemical relations). For instance, in the case of Facebook's user friendship network, GNNs\footnote{Facebook has a friend suggestion functionality that takes information from already-established friendships between two users and suggests other similar profiles to make plausible new friendship connections. Because Facebook’s prediction algorithms remain proprietary, we can assume that GNNs would benefit from this particular task due to their intrinsic characteristics of extrapolating information from the neighbourhoods of the graph vertices.} can be used to predict if two users will become friends in the future, by leveraging information from their neighbourhoods in the graph. GNNs operate by transforming the input graph, using learned latent features for both vertex and edge attributes. The connectivity patterns induced by the edges enforce the relationships in the latent vector space of the learned features (see Sec. A.1 of the supplementary material). However, as black-box models, GNNs are not suitable as decision support systems (DSS) in critical domains.

The literature has received many contributions in interpreting black-box models via feature-based analysis \cite{guidotti2019factual} and counterfactual explanations \cite{guidotti2022counterfactual,lucic2022cf}. Models that perform feature analysis to produce explanations are denoted \textit{feature-based} explainers. However, explanations that reveal a prediction's \textit{why} need to be more comprehensive to understand how to change the outcome of a specific model. Thus, explainers - namely \textit{counterfactual explainers} - that provide examples of what input features to change to obtain a different prediction are necessary to describe cause-effect relationships between the data and the outcome \cite{byrne2019counterfactuals}. Counterfactual explainability could help, for example, obtain the following suggestions:

\noindent\textit{User banning DSS} --- Suppose a user within a fictitious social network posts content attempting to sell drugs illicitly. Such conduct directly violates the network's terms of use, resulting in potential legal consequences and the suspension of the user's profile. To address this, our Decision Support System (DSS) takes action by blocking the user's account and offers a counterfactual explanation, stating that \Quote{had the user refrained from posting about drug sales, her account would not have been banned}. The provision of such explanations aids auditors in classifying banned users based on the seriousness of their violations, thereby contributing to a safer online environment for other users.

\begin{figure}[t]
    \centering
    \makebox[\textwidth]{\includegraphics[width=0.9\textwidth]{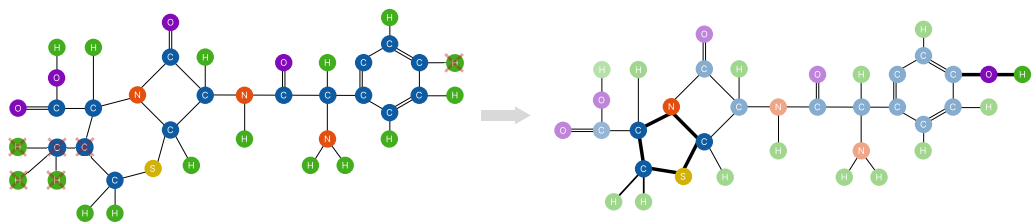}}
    \caption{Drug repurposing mechanism. Passing from cephalexin (left) to amoxicillin (right) by removing atoms and reforming hexagonal connections into pentagons. Here, the counterfactual explanation consists of the highlighted edges and vertices.}
    \label{fig:drug_repurpusing}
\end{figure}

\noindent\textit{Drug repurposing DSS} --- Assume that a drug laboratory aims to cure bacterial infections (e.g., dental abscess) with cephalexin\footnote{\url{https://pubchem.ncbi.nlm.nih.gov/compound/cephalexin}}. When treating such diseases with this drug, our DSS gives a negative result with the explanation, \Quote{if we modified cephalexin's molecular structure as in Fig. \ref{fig:drug_repurpusing}, then we would be able to treat the diseases}. This explanation is useful because it would give sprout to the aminopenicillins class of antibiotics, particularly the amoxicillin\footnote{\url{https://pubchem.ncbi.nlm.nih.gov/compound/amoxicillin}} drug known for its faster treatments and lower side effect risks.

Counterfactual explanations enhance the analysis of black-box models, offering transparency and feedback to non-experts. This fosters trust in critical domains like health and finance and helps identify biases. For example, in the user banning DSS, race-related bias in account suspension can be detected \cite{colmenarejo2022fairness}. Addressing biases based on ethical principles rather than purely performance-driven concepts becomes feasible through counterfactual explainability. However, eliminating bias from data remains a separate and challenging endeavour.

From a market's perspective, developing a new drug is expensive, costing between \$314 million and \$2.8 billion as of 2018, with a low success rate of only 12\% during clinical development \cite{wouters2020estimated}. Clinical trial complexity, larger trial sizes, changes in protocol design, and testing on comparator drugs contribute to this price hike. Clear counterfactual methods can help decipher the causes of outcomes, particularly in scenarios such as drug repurposing, potentially reducing costs and increasing approval rates. Therefore, pharmaceutical companies are interested in adopting innovative explainable graph learning approaches to comprehend trial scenarios without costly tests.

\subsection{Contributions of this survey}\label{sec:differences_survey}\label{sec:intro:contrib}

Graph Counterfactual Explainability (GCE) is still in its early stages since significant attention towards graph learning has risen sharply in the last decade.

This survey addresses GCE, going beyond mere feature-based (factual) explainability in graphs. We provide an in-depth discussion and a systematic literature review of GCE. Specifically, we organise existing counterfactual methods according to a uniform formal notation to facilitate easy comparison of their strengths and pitfalls (See Sec. \ref{sec:unifying_gce}). Additionally, we propose a formalisation of GCE for multi-class prediction problems, deriving the global minimal counterfactual example on a specific graph black-box prediction model (See Sec. \ref{sec:our_gce_definition}). We summarise the strengths and weaknesses of counterfactual explainer methods present in the literature, classifying them according to several dimensions, making it easy for readers to identify alternative methodologies that better suit their scenario (See Sec. \ref{sec:methods}). 
In Sec. \ref{sec:eval}, we discuss the benefits of the evaluation protocol of GCE methods, including the benchmark datasets and measures used in the literature. We integrated the majority of the methods into the GRETEL library, thus conducting an empirical evaluation among them (see Sec. \ref{sec:gretel_exp}). 
In Sec. \ref{sec:privacy}, we give insights on how to adapt privacy-preserving counterfactual explainers and briefly discuss oracle fairness. Open challenges and promising directions in GCE are summarised in Sec. \ref{sec:open_challenges}, and concluding remarks are presented in Sec. \ref{sec:conclusion}. Supplementary material containing background concepts and notation (Sec. A), detailed dataset descriptions (Sec. B), evaluation framework comparisons and GRETEL overview(Sec. C), and the employed hyperparameters adopted in the empirical evaluation (Sec. D).

Nevertheless, we analysed surveys on graph factual explainability \cite{amara2022graphframex,yuan2022explainability} and counterfactual explainability on other data structures for completeness  \cite{artelt2019computation,guidotti2022counterfactual,verma2020counterfactual} since they are the two closest areas. A recent survey \cite{guo2023survey} has emerged on Counterfactual Learning that can be useful to have a broader overview of \textit{Trustworthy AI} applications. Still, ours delves much deeper into the formal notation, the methods' analysis, and the benchmarking datasets and metrics, as it is possible to notice in Table \ref{tab:survey_comparison}. In addition to it, we discuss in depth the evaluation protocols of each work. We present an empirical evaluation of the methods and suggest possible strategies for privacy-preserving counterfactual explainers.

We analysed the existing surveys according to seven dimensions (see Table \ref{tab:survey_comparison}), reflecting the scope, the domains and the typical workflow of GCE's research area. \textit{Domain} depicts the considered graph structure. \textit{Prediction Task} represents the goal of the explainable methods. \textit{Definition} depicts formal and uniform definitions adopted for counterfactual explainability. \textit{Explainer Adaptation} illustrates the 
extension of feature-based explainers for counterfactual generation purposes. \textit{Evaluation} illustrates the evaluation of methods via some established evaluation protocol and designated metrics. \textit{Privacy of Explainers, and Fairness of Oracles} considers the importance of privacy maintenance at the explainers' level and the fairness of the underlying prediction models. We argue that these dimensions are a better framework for categorising and comparing existing surveys.

\begin{table}[!ht]
\centering
\caption{Qualitative comparison of the surveys present in the literature according to the dimensions reflecting the scope and workflow of GCE approaches. $\cdot$ depicts a missing aspect, \checkmark a covered aspect, $\sim$ a partially covered aspect.}
\label{tab:survey_comparison}
\resizebox{\textwidth}{!}{%
\begin{tabular}{@{}lllccccccc@{}}
\toprule
\multicolumn{3}{l}{\multirow{4}{*}{}} & \multicolumn{7}{c}{Surveys on Explainability} \\ \cmidrule(l){4-10} 
\multicolumn{3}{l}{} & \multicolumn{4}{c|}{Considering graphs data} & \multicolumn{3}{c}{Considering other data types} \\ \cmidrule(l){4-10} 
\multicolumn{3}{l}{} & \multicolumn{2}{c|}{Counterfactual} & \multicolumn{2}{c|}{Factual} & \multicolumn{3}{c}{Counterfactual} \\
\multicolumn{3}{l}{} & [this] & \multicolumn{1}{c|}{Guo et al. \cite{guo2023survey}} & Amara et al. \cite{amara2022graphframex} & \multicolumn{1}{c|}{Yuan et al. \cite{yuan2022explainability}} & \multicolumn{1}{l}{Artlet and Hammer \cite{artelt2019computation}} & \multicolumn{1}{l}{Guidotti \cite{guidotti2022counterfactual}} & \multicolumn{1}{l}{Verma et al.\cite{verma2020counterfactual}} \\ \midrule
\multirow{4}{*}{Domain} & \multicolumn{2}{l}{Synthetic networks} & \checkmark & \multicolumn{1}{c|}{\checkmark} & \checkmark & \multicolumn{1}{c|}{\checkmark} & $\cdot$ & $\cdot$ & $\cdot$ \\
 & \multicolumn{2}{l}{Social networks} & \checkmark & \multicolumn{1}{c|}{\checkmark} & \checkmark & \multicolumn{1}{c|}{\checkmark} & $\cdot$ & $\cdot$ & $\cdot$ \\
 & \multicolumn{2}{l}{Molecular networks} & \checkmark & \multicolumn{1}{c|}{\checkmark} & $\cdot$ & \multicolumn{1}{c|}{\checkmark} & $\cdot$ & $\cdot$ & $\cdot$ \\
 & \multicolumn{2}{l}{Omics Networks} & \checkmark & \multicolumn{1}{c|}{\checkmark} & $\cdot$ & \multicolumn{1}{c|}{$\cdot$} & $\cdot$ & $\cdot$ & $\cdot$ \\ \midrule
\multirow{4}{*}{Prediction Task} & \multicolumn{2}{l}{Vertex classification} & \checkmark & \multicolumn{1}{c|}{\checkmark} & $\cdot$ & \multicolumn{1}{c|}{\checkmark} & $\cdot$ & $\cdot$ & $\cdot$ \\
 & \multicolumn{2}{l}{Link (edge) prediction} & \checkmark & \multicolumn{1}{c|}{\checkmark} & $\cdot$ & \multicolumn{1}{c|}{$\cdot$} & $\cdot$ & $\cdot$ & $\cdot$ \\
 & \multicolumn{2}{l}{Graph classification} & \checkmark & \multicolumn{1}{c|}{\checkmark} & $\cdot$ & \multicolumn{1}{c|}{\checkmark} & $\cdot$ & $\cdot$ & $\cdot$ \\
 & \multicolumn{2}{l}{Graph-pair affinity} & \checkmark & \multicolumn{1}{c|}{$\cdot$} & $\cdot$ & \multicolumn{1}{c|}{$\cdot$} & $\cdot$ & $\cdot$ & $\cdot$ \\ \midrule
\multicolumn{3}{l}{Definition} & \checkmark & \multicolumn{1}{c|}{\checkmark} & $\cdot$ & \multicolumn{1}{c|}{$\cdot$} & \checkmark & \checkmark & \checkmark \\ \midrule
\multicolumn{3}{l}{Explainer Adaptation} & \checkmark & \multicolumn{1}{c|}{\checkmark} & $\cdot$ & \multicolumn{1}{c|}{\checkmark} & \checkmark & $\cdot$ & $\cdot$ \\ \midrule
\multirow{25}{*}{Evaluation} & \multirow{2}{*}{Datasets} & Synthetic & \checkmark & \multicolumn{1}{c|}{\checkmark} & \checkmark & \multicolumn{1}{c|}{\checkmark} & $\cdot$ & $\cdot$ & $\cdot$ \\
 &  & Real & \checkmark & \multicolumn{1}{c|}{\checkmark} & \checkmark & \multicolumn{1}{c|}{\checkmark} & $\cdot$ & \checkmark & \checkmark \\ \cmidrule(l){2-10} 
 & \multicolumn{2}{l}{Reproducibility} & \checkmark & \multicolumn{1}{c|}{$\cdot$} & $\sim$ & \multicolumn{1}{c|}{$\sim$} & $\cdot$ & $\sim$ & $\cdot$ \\
 & \multicolumn{2}{l}{Extensibility} & \checkmark & \multicolumn{1}{c|}{$\cdot$} & $\cdot$ & \multicolumn{1}{c|}{$\cdot$} & $\cdot$ & $\cdot$ & $\cdot$ \\
 & \multicolumn{2}{l}{Supported Methods} & \checkmark & \multicolumn{1}{c|}{$\cdot$} & \checkmark & \multicolumn{1}{c|}{\checkmark} & $\cdot$ & \checkmark & $\cdot$ \\
 & \multicolumn{2}{l}{Evaluation Protocol} & \checkmark & \multicolumn{1}{c|}{$\cdot$} & $\cdot$ & \multicolumn{1}{c|}{$\cdot$} & $\cdot$ & $\cdot$ & $\cdot$ \\ \cmidrule(l){2-10} 
 & \multirow{19}{*}{Metrics} & Runtime & \checkmark & \multicolumn{1}{c|}{$\cdot$} & \checkmark & \multicolumn{1}{c|}{$\cdot$} & $\cdot$ & \checkmark & \checkmark \\
 &  & Oracle Calls & \checkmark & \multicolumn{1}{c|}{$\cdot$} & $\cdot$ & \multicolumn{1}{c|}{$\cdot$} & $\cdot$ & $\cdot$ & $\cdot$ \\
 &  & Oracle Accuracy & \checkmark & \multicolumn{1}{c|}{$\cdot$} & $\cdot$ & \multicolumn{1}{c|}{$\cdot$} & $\cdot$ & $\cdot$ & $\cdot$ \\
 &  & Correctness & \checkmark & \multicolumn{1}{c|}{$\cdot$} & $\sim$ & \multicolumn{1}{c|}{$\cdot$} & $\cdot$ & $\cdot$ & \checkmark \\
 &  & Sparsity & \checkmark & \multicolumn{1}{c|}{\checkmark} & $\cdot$ & \multicolumn{1}{c|}{\checkmark} & $\cdot$ & $\cdot$ & \checkmark \\
 &  & Fidelity & \checkmark & \multicolumn{1}{c|}{\checkmark} & \checkmark & \multicolumn{1}{c|}{\checkmark} & $\cdot$ & $\cdot$ & $\cdot$ \\
 &  & Robustness & \checkmark & \multicolumn{1}{c|}{\checkmark} & $\cdot$ & \multicolumn{1}{c|}{\checkmark} & $\cdot$ & \checkmark & $\cdot$ \\
 &  & Explainer Accuracy & \checkmark & \multicolumn{1}{c|}{\checkmark} & \checkmark & \multicolumn{1}{c|}{\checkmark} & $\cdot$ & $\cdot$ & $\cdot$ \\
 &  & Prediction  Distance & \checkmark & \multicolumn{1}{c|}{$\cdot$} & $\cdot$ & \multicolumn{1}{c|}{$\cdot$} & $\cdot$ & $\cdot$ & $\cdot$ \\
 &  & Causality & \checkmark & \multicolumn{1}{c|}{$\cdot$} & $\cdot$ & \multicolumn{1}{c|}{$\cdot$} & $\cdot$ & $\cdot$ & \checkmark \\
 &  & Diversity & \checkmark & \multicolumn{1}{c|}{$\cdot$} & $\cdot$ & \multicolumn{1}{c|}{$\cdot$} & $\cdot$ & \checkmark & \checkmark \\
 &  & Actionability & \checkmark & \multicolumn{1}{c|}{$\cdot$} & $\cdot$ & \multicolumn{1}{c|}{$\cdot$} & $\cdot$ & \checkmark & $\cdot$ \\ \cmidrule(l){3-10} 
 &  & GED & \checkmark & \multicolumn{1}{c|}{\checkmark} & $\cdot$ & \multicolumn{1}{c|}{$\cdot$} & $\cdot$ & $\cdot$ & $\cdot$ \\
 &  & Explanation Size & \checkmark & \multicolumn{1}{c|}{$\cdot$} & $\cdot$ & \multicolumn{1}{c|}{$\cdot$} & $\cdot$ & $\sim$ & $\sim$ \\
 &  & Tanimoto Similarity & \checkmark & \multicolumn{1}{c|}{$\cdot$} & $\cdot$ & \multicolumn{1}{c|}{$\cdot$} & $\cdot$ & $\cdot$ & $\cdot$ \\
 &  & MEG Similarity & \checkmark & \multicolumn{1}{c|}{$\cdot$} & $\cdot$ & \multicolumn{1}{c|}{$\cdot$} & $\cdot$ & $\cdot$ & $\cdot$ \\ \cmidrule(l){3-10} 
 &  & Recourse Cost & \checkmark & \multicolumn{1}{c|}{$\cdot$} & $\cdot$ & \multicolumn{1}{c|}{$\cdot$} & $\cdot$ & $\cdot$ & $\cdot$ \\
 &  & Coverage & \checkmark & \multicolumn{1}{c|}{$\cdot$} & $\cdot$ & \multicolumn{1}{c|}{$\cdot$} & $\cdot$ & $\cdot$ & $\cdot$ \\
 &  & Compactness & \checkmark & \multicolumn{1}{c|}{$\cdot$} & $\cdot$ & \multicolumn{1}{c|}{$\cdot$} & $\cdot$ & $\cdot$ & $\cdot$ \\ \midrule
\multicolumn{3}{l}{Privacy of Explainers} & $\sim$ & \multicolumn{1}{c|}{$\cdot$} & $\cdot$ & \multicolumn{1}{c|}{$\cdot$} & $\cdot$ & $\cdot$ & $\cdot$ \\ \midrule
\multicolumn{3}{l}{Fairness of Oracles} & $\sim$ & \multicolumn{1}{c|}{\checkmark} & $\cdot$ & \multicolumn{1}{c|}{$\cdot$} & $\cdot$ & $\cdot$ & $\cdot$ \\ \bottomrule
\end{tabular}%
}
\end{table}

\noindent\textbf{Domain} --- We identified three\footnote{Notice that synthetic datasets have been adopted in GCE, and we include them in the table. Alas, they do not constitute a category of real-world datasets.} main categories of graphs: i.e., social, molecular, and -omics networks. Other surveys concentrate on social \cite{amara2022graphframex,yuan2022explainability,guo2023survey} and molecular networks \cite{yuan2022explainability,guo2023survey} since they represent traditional structures in deep graph learning. Besides extensively discussing the first three graph types, we also shed light on the structure of -omics networks \cite{madeddu2022deep}, useful for analysing the interconnection of biological processes.
    
\noindent\textbf{Prediction Task} --- In this context, we have identified four types of tasks, namely vertex classification, graph classification, link prediction, and graph-pair affinity. These tasks rely on different concepts and have unique characteristics. Vertex classification aims to predict the role of each vertex in a graph by analysing the role of its neighbouring nodes. For example, the prediction task in Zachary's karate club is to classify whether a member will switch to another club. Link prediction, on the other hand, predicts relationships between vertices. For instance, in an image whose entities are vertices, the task is to predict which of them share an edge. Graph classification predicts the property of an entire graph, such as whether a molecule - represented as a graph - binds to a receptor involved in a specific disease. Finally, graph-pair affinity involves the prediction of the similarity between pairs of instances, such as a drug-target binding for drug discovery or drug repurposing.
    
\noindent\textbf{Definition} --- Although graph neural networks \cite{scarselli2008graph,DBLP:conf/iclr/VelickovicCCRLB18} and counterfactual explainability \cite{artelt2019computation,guidotti2022counterfactual,verma2020counterfactual,guo2023survey} have been extensively defined, this survey is the first contribution in the literature which provides a uniform GCE definition. First, we agglomerate the literature's graph counterfactual definitions according to a single notation, and then, we provide a formalisation of the GCE problem. Thus, future researchers can use it to align their work with the SoA.

\noindent\textbf{Explainer Adaptation} --- Some methods in GCE use factual explainability as a starting point and change the features of the factual to produce the counterfactual \cite{yuan2022explainability,guo2023survey}. Here, we analyse the counterfactual technique as an adaptation similar to what has been done in factual explanation approaches or as an extension of image-based explainability.
    
\noindent\textbf{Evaluation} ---  A crucial aspect of evaluating methods in the literature is a uniform benchmarking system that uses standardised metrics and datasets. While other surveys \cite{amara2022graphframex,verma2020counterfactual,yuan2022explainability,guo2023survey} mention the characteristics and potential issues of synthetic and real datasets, only Guidotti \cite{guidotti2022counterfactual} and Yuan et al. \cite{yuan2022explainability} provide in-depth explanations of the reproducibility\footnote{Notice that Amara et al. \cite{amara2022graphframex} briefly discuss their explainability but do not show what hyperparameters to use for reproducibility purposes.} of their evaluation framework by sharing links to repositories that contain the code base. These frameworks have several explanation methods from the literature that are readily available. However, they \cite{amara2022graphframex,guidotti2022counterfactual,yuan2022explainability} are challenging to adapt and reuse for experiments in different scenarios than those already included. This survey focuses in empirically assessing the performance of SoA explainers using GRETEL, a modular framework for evaluating GCE. GRETEL has an advantage when it comes to reproducibility (see Sec. C.2). Additionally, we provide the reader with the evaluation protocol used in each paper to shed light on various counterfactuality scenarios, thus, the first survey to give such detail. We included several metrics proposed in the literature to evaluate the explainers' goodness, such as oracle accuracy, sparsity, and accuracy. The surveys in \cite{amara2022graphframex,guidotti2018survey,verma2020counterfactual} also cover the runtime of the explanation technique besides the three previous ones. Besides using only graph edit distance (GED) as in \cite{guo2023survey}, we include two classes of metrics - i.e., minimality evaluation (\textit{GED}, \textit{Explanation Size},  \textit{Tanimoto Similarity}, and \textit{MEG Similarity}) and global metrics (\textit{Recourse Cost}, \textit{Coverage}, and \textit{Compactness})   useful for future investigations.

\noindent\textbf{Privacy of Explainers and Fairness of Oracles} 
--- An important issue regarding trustworthiness is the interplay between privacy, explainability, and fairness.  We observed that there is very little attention posed in the GCE domain.
Specifically, all the other surveys fail to address the privacy infringement that explainers could introduce when presenting counterfactual examples to end-users, given the vast amounts of sensitive information generated by social networks. 
On the other hand, counterfactual explanations on graphs can be employed as a mechanism to understand and cope with the oracle's bias. For this reason, Sec. \ref{sec:privacy} proposes how to construct privacy-preserving explanation strategies. We discuss how explainability can be harnessed to tackle the unfairness of the underlying model by using generated counterfactuals as an intervention mechanism to cope with such situations. \newline

Table \ref{tab:survey_comparison} highlights several limitations of factual surveys on graphs. Firstly, they rely solely on synthetic, social, and molecular networks. Secondly, they fail to explain the unique characteristics of link prediction tasks and how they differ from vertex and graph classification tasks. Additionally, these surveys provide a list of explanation techniques without analysing their primary limitations. While some explainers are suitable for all prediction tasks, others may only be effective in certain scenarios. Contrarily, Guo et al. \cite{guo2023survey} incorporate these aspects. Nevertheless, they fail to address evaluation metrics such as minimality evaluation and global measures, which are useful to provide a list of counterfactual examples to explain an entire dataset (or set of instances). They also fail to provide a benchmark of the current SoA: they only list the available frameworks in the literature and do not compare them according to a reproducible, replicable, generalisable, and robustness point-of-view (see Sec. C.1).

\section{Defining counterfactual explanations in graphs}\label{sec:gce_def}

Here, we discuss the GCE definitions used in the literature. Notice that not all the works have provided a formal graph explanation. Most of them present a loss function, typically based on the distance between the prediction of the input instance and its counterfactual. Hence, we provide a uniform definition as a cornerstone in this research area, discussing the (dis)similarities of what is proposed in the literature (see Sect \ref{sec:unifying_gce}). Then, we provide our definitions of multi-class and global minimal counterfactuals (see Sec. \ref{sec:our_gce_definition}). We refer the reader to Sec. A.1 for details on the notations and background knowledge useful to follow this paper.

\subsection{Unifying GCE definitions in the literature}\label{sec:unifying_gce} 
The literature in GCE has tried to provide a formalisation for the problem definition and, eventually, the way a counterfactual example is defined. Although the definitions do not follow a rigorous formalisation, the literature states that a counterfactual example satisfies the following equation:
\begin{equation}
    \Phi(x) \neq \Phi(x')
\end{equation}
where $x$  is the input instance (i.e., graph, vertex, or edge), $x'\in X'$ is the counterfactual example, $X'$ is a set of possible counterfactuals, and $\Phi(\cdot)$ is the prediction of the oracle $\Phi$. For example, in the case of graph classification $x=G$ and its counterfactual $x'=G'$ where $G=(V,E)$ is a graph with vertices $V$ and edges $E$.  Nevertheless, we believe that the following equation is a more inclusive way to define a counterfactual for both classification and regression tasks. 
\begin{equation}
    \mathcal{D}_{pred}(\Phi(x),\Phi(x')) \geq t
\end{equation}
where $\mathcal{D}_{pred}(\Phi(x),\Phi(x'))$ is the distance function between the prediction of the original instance and the counterfactual one, and $t$ is a threshold. The multi-class and the multi-label classification tasks can be addressed by a specialised $\mathcal{D}_{pred}$.

In this survey, we generalise the generation of a counterfactual via the minimisation of a loss function because the majority of the works follows an optimisation process. Hence, the generation of a counterfactual in GCE can be expressed as a combination of a minimisation objective and a regularisation term, constrained by the $\gamma$ term: 
\begin{equation}\label{eq:gce_unification}
    \arg \min_{\gamma} \alpha \cdot \mathcal{L}_{pred} + \beta \cdot \mathcal{L}_{inst}
\end{equation}
where $\mathcal{L}_{pred}$ is an optimisation function over the prediction outcome of the oracle $\Phi$, $\mathcal{L}_{inst}$ is a regularisation term over the original and counterfactual instances, $\alpha, \beta \in [0,1]$ account for the contribution of each term in the overall optimisation function.

Subsequently, we discuss the (dis)advantages and (dis)similarities of each GCE definition in the literature w.r.t to Eq. \ref{eq:gce_unification} by specialising them according to the following notation.  

We use $\mathcal{S}_{pred}(\Phi(x),\Phi(x'))$ to indicate a similarity function between the outcome of the original instance $\Phi(x)$ and the counterfactual one $\Phi(x')$. Additionally, we denote with $\mathcal{S}_{inst}(x,x')$ the similarity function between the original instance $x$ and the counterfactual one $x'$ and  $\mathcal{D}_{inst}(x,x')$ is its distance counterpart.

According to Lucic et al. \cite{lucic2022cf}, $\mathcal{L}_{pred}$ is the negative log-likelihood (NLL), $\alpha = - \mathbb{1}[\Phi(x) = \Phi(x')]$ which produces $-1$ if $\Phi(x) = \Phi(x')$ and $0$ otherwise, $\mathcal{L}_{inst}$ is the distance between $x$ and $x'$, $\beta$ remains a free hyperparameter, and $\gamma = x' \in X'$. Hence, we can rewrite this optimisation function as follows:
\begin{equation}\label{eq:lucic_new}
    \arg \min_{x' \in X'} \mathbb{1}[\Phi(x) = \Phi(x')] \cdot \mathcal{D}_{pred}(\Phi(x),\Phi(x')) + \beta \cdot \mathcal{D}_{inst}(x,x')
\end{equation}
Notice that the chosen formulation for $\mathcal{L}_{pred}$ admits a non-counterfactual example as a possible generated explanation example, including the original instance that minimises the overall loss. In other words, if $x' = x$, then $\mathcal{D}_{pred}(\Phi(x), \Phi(x')) = \mathcal{D}_{inst}(x,x') = 0$. In this case, the original instance $x$ is also a minimal counterfactual example. Similarly to Eq. \ref{eq:gce_multiclass_sim}, Eq. \ref{eq:lucic_new} supports multi-class classification since the prediction from the oracle $\Phi(x)$ is not enticed to a binary scenario. Nevertheless, Eq. \ref{eq:lucic_new} cannot support the generation of multiple counterfactual examples $x'$ belonging to different classes w.r.t. the original instance $x$ since it tries to find a global explanation.

Wellawatte et al \cite{wellawatte2022model} do not consider $\mathcal{L}_{pred}$ explicitly. Rather, they embed the search for valid counterfactuals in $\gamma$, making them robust towards the pitfall of considering the original instance $x$ the minimal counterfactual, as seen in Eq. \ref{eq:lucic_new}. We can rewrite the original formulation  as follows:
\begin{equation}\label{eq:wellatte_new}
    \arg\min_{x' \in X' \; | \; \Phi(x) \neq \Phi(x')} \mathcal{D}_{inst}(x,x')
\end{equation}
where $\alpha$ and $\mathcal{L}_{pred}$ can be omitted, $\beta = 1$, $\mathcal{L}_{inst}$ is the distance between $x$ and $x'$, and $\gamma = x' \in X' \;|\; \Phi(x) \neq \Phi(x')$. In the original paper, the authors rely on the Tanimoto index as a similarity measure between $x$ and $x'$. To remain compliant with Eq. \ref{eq:gce_unification}, we minimise the negated Tanimoto index instead of maximising it. Eq. \ref{eq:wellatte_new} supports multi-class classification scenarios as in Eq. \ref{eq:gce_multiclass_sim}, however it cannot produce a set of counterfactuals for the same input instance $x$. Furthermore, notice that this optimisation function is the same as Eq. \ref{eq:minimal_gce}.

Similarly to \cite{wellawatte2022model}, Abrate and Bonchi \cite{abrate2021counterfactual} search for a counterfactual graph $x'$ such that the symmetric difference between $x$ and $x'$ is minimised. Only the constraint $\gamma$ is changed to $\Phi(x) = 1 - \Phi(x')$, which confines finding valid counterfactuals to only binary classification scenarios. Hence, Eq. \ref{eq:wellatte_new} becomes the following:
\begin{equation}\label{eq:bonchi_new}
    \arg\min_{x' \in X' \; | \; \Phi(x) = 1- \Phi(x')} \mathcal{D}_{inst}(x,x')
\end{equation}

Numeroso and Bacciu \cite{numeroso2021meg} optimise a distance-based loss function for molecules that incorporates their structural information. They exploit both a distance between the outcomes of the original molecule $x$ and the counterfactual one, $x'$, and a similarity between the structure of the two molecules. According to the original paper, when performing a classification, $\Phi$ emits a probability distribution for a certain set of classes $C$. Hence, given a molecule in input $x$ and its outcome $\arg \max_{c \in C} \Phi(x)$, the authors produce counterfactuals $x'$ that maximise the outcome of classes different from $c$, and the similarity between $x$ and $x'$. The final form of the proposed optimisation function maximises the negation of the probability of outputting a class equal to $c$ and the structural similarity between $x$ and $x'$ controlled by the weight $\alpha$ and $\beta = 1-\alpha$, respectively. Maximising the negation of the probability of producing $c$ is fancy wording for maximising the distance between the outcomes on the original instance $x$ and the counterfactual $x'$. It is important to highlight that the counterfactual can be generated by adding and removing both vertices and edges. Therefore, the counterfactual graph is not necessarily a sub-graph of the original one. Before uniforming what is described here in the same form of Eq. \ref{eq:gce_unification}, it is useful to review the original optimisation function as follows:
\begin{equation}\label{eq:numeroso_intermediate}
    \arg \max_{x' \in X'} \alpha \cdot \mathcal{D}_{pred}(\Phi(x),\Phi(x')) + (1-\alpha) \cdot \mathcal{S}_{inst}(x,x')
\end{equation}
We can easily translate Eq. \ref{eq:numeroso_intermediate} by considering its dual function (i.e., minimisation) as follows:
\begin{equation}\label{eq:numeroso_new}
      \arg \min_{x' \in X'} (1-\alpha) \cdot \mathcal{S}_{pred}(\Phi(x), \Phi(x')) + \alpha \cdot \mathcal{D}_{inst}(x,x')
\end{equation}
Notice that the condition $\gamma$ remains the same. Nevertheless, $\mathcal{L}_{pred}$ changes into its dual form (i.e., from maximising distance of predicted classes $\mathcal{D}_{pred}(\Phi(x),\Phi(x'))$, to minimising their similarity $\mathcal{S}_{pred}(\Phi(x), \Phi(x'))$). A similar reasoning is applied to $\mathcal{L}_{inst}$. Naturally, the weights $\alpha$ and $\beta=1-\alpha$ are specular. Notice that this formalisation is a generalisation of Eq. \ref{eq:wellatte_new} and \ref{eq:bonchi_new} since it introduces the minimisation of $\mathcal{S}_{pred}(\Phi(x), \Phi(x'))$ instead of the constraints $\Phi(x) \neq \Phi(x')$ and $\Phi(x) = 1 - \Phi(x')$, respectively. Similarly, Ma et al. \cite{ma2022clear} rely on Eq. \ref{eq:numeroso_new} to produce valid counterfactuals. In the original paper, the authors include the minimisation of the KL divergence that makes the explainer learn to produce realistic counterfactuals w.r.t. the distribution of the original instances and the ground truth. However, producing realistic counterfactuals can be included as a component of $\mathcal{D}_{inst}(x,x')$. 

Nguyen et al. \cite{nguyen2022explaining} rely on oracles that predict drug-target pair interactions. In this scenario, $\Phi$ takes in input pairs of instance $(x_d, x_t) \in X_d \times X_t$ and produces an output, where $X_d$ is the set of drugs, and $X_t$ is the set of targets. Naturally, $X_d'$ and $X_t'$ are the counterfactual sets of $X_d$ and $X_t$, respectively. Hence, for each instance pair $(x_d, x_t)$, we obtain the minimal counterfactual pair $(x_d',x_t')$ by optimising the following function:
\begin{equation}\label{eq:nguyen_new}
    \arg \min_{x_d' \in X_d'\;,\; x_t' \in X_t'} \alpha \cdot \mathcal{S}_{pred}({\Phi(x_d,x_t), \Phi(x_d',x_t')}) + \beta \cdot (\mathcal{D}_{inst}(x_d,x_d') + \mathcal{D}_{inst}(x_t,x_t'))
\end{equation}

Although Huang et al. \cite{huang2023global} provide a set of counterfactual explanations for the oracle $\Phi$, we can easily translate their optimisation according to Eq. \ref{eq:gce_unification}. To this end, we rely on the power set $\mathcal{P}(X')$ of all possible counterfactuals, $X'$. In detail, the authors maximise the coverage of the counterfactual set $X^* \in \mathcal{P}(X')$ w.r.t. the size of the set of instances, $|X|$:
\begin{equation}\label{eq:kosan_intermediate}
    \arg \max_{X^* \in \mathcal{P}(X')\;|\; |X^*| = k} \frac{\big|x \in X\;|\; \min_{x'\in X^*} \mathcal{D}_{inst}(x,x') \leq \theta\big|}{\big|X\big|}
\end{equation}
where $\theta$ is an upper-bound for the distance between the original instance $x$ and the counterfactual $x'$, and $k$ is the size of the counterfactual set $X^*$ drawn from $\mathcal{P}(X')$. Notice that the fraction in Eq. \ref{eq:kosan_intermediate} is in $[0,1]$, meaning that we can rewrite it as a minimisation function:
\begin{equation}\label{eq:kosan_new}
    \arg \min_{X^* \in \mathcal{P}(X')\;|\; |X^*| = k} 1 - \frac{\big|x \in X\;|\; \min_{x'\in X^*} \mathcal{D}_{inst}(x,x') \leq \theta\big|}{\big|X\big|}
\end{equation}
Notice that Eq. \ref{eq:kosan_new} is compliant with the notation of Eq. \ref{eq:gce_unification} where $\gamma = X^* \in \mathcal{P}(X') \;|\; |X^*| = k$, $\mathcal{L}_{pred}$ is not specified since $\gamma$ implies only searching for valid counterfactuals, and $\mathcal{L}_{inst}$ is the negated coverage function with $\beta = 1$.

Bajaj et al. \cite{bajaj2021robust} are the only ones that provide a formal definition for GCE instead of optimising a loss function. Hence, given a model $\Phi$ trained on a set of graphs, for an input graph $G = \{V, E\}$, the authors explain why $G$ is predicted as $\Phi(G)$ by identifying a small subset of edges $S \subseteq E$, such that (1) removing the set of edges in $S$ from $G$ changes the prediction on the remainder $\{V, E-S\}$ of $G$ significantly; and (2) $S$ is stable w.r.t. slight changes on the edges of $G$ and the feature representations of the vertices of $G$. According to this definition, the authors cannot always change $\Phi$'s prediction only relying on edge removals. Generally, it is not always possible to find a set $S \subseteq E$ that maintains the stability w.r.t. changes on $G$. Thus, generating a counterfactual example is impossible. Additionally, the set $E-S$ contains the factual edges for which $\Phi(G = (V,E-S)) \neq \Phi(G=(V,E))$. Without loss of generality, all methods that generate counterfactual examples by only removing edges suffer from the same phenomenon.

The works in \cite{tan2022learning,liu2021multi} are factual-based methods that can be used to derive counterfactuals. Generally, the counterfactual example can be considered as the remainder of the original graph when the factual explanation (e.g., usually a sub-graph) gets eliminated. Furthermore, by concentrating first in identifying the subset of edges, vertices, and vertex attributes to form a sub-graph for the factual explanation, these works do not guarantee that the counterfactual example is minimal. We do not formalise these methods due to their complete disalignment with Eq. \ref{eq:gce_unification}.

\subsection{Minimal graph counterfactual explanations}\label{sec:our_gce_definition}
Considering the variety of the proposed GCE definitions in the literature, it is necessary to define a comprehensive one which, on the one hand, encompasses all of them and, on the other, ensures their correctness and provides enough flexibility to embed future definitions. 
First, we present a general setting of our GCE definition considering the instance classification problem (see Def. \ref{def:our_gce}). 

\begin{definition}\label{def:our_gce}
\textit{Multi-class minimal counterfactual examples}:  Let $\Phi$ be a prediction model that classifies $x$ into a class $c \in C$ from a set of classes $C$. Let $X'$ be the set of possible counterfactual examples $x'$ and $\mathcal{S}_{inst}(x,x')$ be a similarity measure that tells how similar $x'$ is to $x$. Then, we define the set of counterfactual examples w.r.t. $\Phi$ as follows:
\begin{equation}\label{eq:gce_multiclass_sim}
\begin{gathered}
    s(c',x) \coloneqq  \max_{x' \in X' , x \neq x'} \{\mathcal{S}_{inst}(x,x')\;  | \; \Phi(x') = c'\}\\
    \mathcal{E}_\Phi(x) = \bigcup_{c' \in C - \{c\}} \{  x' \in X' \;|\; x \neq x', \mathcal{S}_{inst}(x,x') = s(c',x)\}
\end{gathered}
\end{equation}
\end{definition}
According to Eq. \ref{eq:gce_multiclass_sim}, $\mathcal{E}_\Phi(x)$ has the maximally similar counterfactual examples $x' \in X'$ to the original graph $x$ for each class $c'$ that is different to the prediction on $x$ (i.e.,  $c' \in C - \{c\}$ where $c = \Phi(x)$). More specifically, we find those counterfactual examples that must be different from the original instance $x$ such that they maximise a similarity function w.r.t. the original instance $\mathcal{S}_{inst}(x,x')$. Accordingly, we can refer to the counterfactual examples of a specific class $c' \in C - \{c\}$ as $\mathcal{E}_{c',\Phi}(x) = \{x' \in X'\;|\;x\neq x', \mathcal{S}_{inst}(x,x')=s(c',x)\}$.

The definition above has two main advantages w.r.t. the ones provided in the literature: i.e., (1) it supports all the graph-based tasks in a multi-class scenario, and (2) it contains all the minimal counterfactual examples for each class $c \in C$. Furthermore, differently from the majority of the formalisations in Sec. \ref{sec:unifying_gce}, we found that using a similarity function $\mathcal{S}_{inst}(x,x')$ is more beneficial because of its flexibility since it might consider both the attributes and the structure of the graph (i.e vertices, vertex/edge attributes, and edges). 
Finally, we constrain the counterfactual example to be different from the original instance (i.e., $x' \neq x$). 

\begin{definition}\label{def:minimal_gce}
\textit{Global minimal counterfactual example}: Let $\Phi$ be a prediction model that classifies $x$ into a class $c \in C$. Let $X'$ is the set that contains all the possible counterfactual examples $x'$. We define the global minimal counterfactual example $\mathcal{E}^*_{\Phi}(x)$ of $x$, as follows:
\begin{equation}\label{eq:minimal_gce}
    \begin{gathered}
    \mathcal{E}^*_{\Phi}(x) = \arg\max_{x' \in X'}  \mathcal{S}_{inst}(x,x')
    \end{gathered}
\end{equation}
\end{definition}

\noindent As anticipated at the beginning of this section, we can extend Def. \ref{def:our_gce} to consider also vertex and edge classification tasks. The only component that changes is the prediction model $\Phi$. Recall that we denote with $x$ an instance now it can be a vertex or an edge belonging to the original graph $G$. In this way, the prediction model $\Phi$ takes in input the instance $x$ and $G$ to produce a class $c$ (i.e., $\Phi(x,G) = c$). Hence, for a particular counterfactual example graph $G'$, $\Phi(x,G') = c'$. It is advisable that the generated counterfactual $G'=(V',E')$ contains the original instance $x$: i.e., $x \in V'$ for vertex prediction and $x \in E'$ for edge prediction. In this way, it is possible to understand how $x$'s relations (vertices or edges) in its vicinity have changed in $G'$ w.r.t. $G$.

\section{Methods}\label{sec:methods}

Here, we present the Graph Counterfactual Explanations (GCE) methods present in the literature gathered following this process. We collected the works from Google Scholar on a bimonthly basis from September 2021 according to: {\footnotesize
    \begin{flalign*}
      & \text{\texttt{intitle}:\textit{counterfactual} \texttt{AND}}\\
        & \text{{[ ( \texttt{intitle}:\textit{graphs} \texttt{OR} \texttt{intitle}:\textit{graph} \texttt{OR} \texttt{intitle}:\textit{gnn} \texttt{OR} \texttt{intitle}:\textit{drug} \texttt{OR} \texttt{intitle}:\textit{molecular} \texttt{OR} \texttt{intitle}:\textit{molecules} )}}\\
        & \text{\texttt{AND} }\text{( \texttt{intitle}:\textit{explainable} \texttt{OR} \texttt{intitle}:\textit{explanations} \texttt{OR} \texttt{intitle}:\textit{explaining} \texttt{OR} \texttt{intitle}:\textit{explainer} ) ]}
    \end{flalign*}
}
The searches produced a total of 21 results, which we reviewed according to the following inclusion/exclusion criteria: we kept only works on graphs as input data evaluating their quality and their in-scope; we excluded Theses and non-original (reproducibility) papers; we excluded works that treat agents' explainability. After the selection, we obtained an initial set of 10 papers. We examined the papers cited and the papers which cite this initial set of papers according to the previously defined inclusion/exclusion criteria. We also monitored the literature with google scholar alerts. Finally, we collected fourteen papers (fifteen methods) \cite{abrate2021counterfactual,wu2021counterfactual,numeroso2021meg,bajaj2021robust,liu2021multi,sun2021preserve,lucic2022cf,wellawatte2022model,tan2022learning,ma2022clear,nguyen2022explaining,cai2022probability,huang2023global,chen2022grease}  that are at the base of this survey.

In the following, we show the gathered methods  organised within the GCE taxonomy that we identified (Sec. \ref{sec:gce_in_general_explanations}). In Sec. \ref{sec:gce_classification}, we summarise all the works in Table \ref{tab:methods} accordingly to  the ten chosen dimensions that we present. Finally, in Sec. \ref{sec:graph_counterfactual_explanations}, we discuss the methods by describing them and showing how they fit in each dimension. 

\subsection{Graph Counterfactual Explainability Taxonomy}\label{sec:gce_in_general_explanations} 

\begin{figure}[!t]
  \centering
  \includegraphics[width=\textwidth]{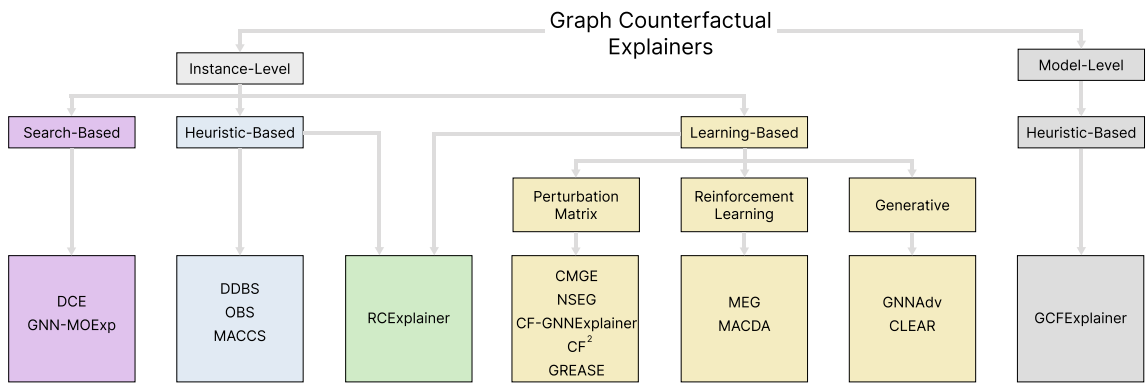}
  \caption{Taxonomy of Graph Counterfactual Explanation (GCE) methods.}
   \label{fig:ge_taxonomy}
\end{figure}

Upfront, we need to clarify the main differences between graph counterfactuality and that on other data structures. First, GCE methods consider the graph structure when generating counterfactuals. This means that changes to the structure (e.g., edge additions/removals) can be used to generate the explanation. Generally, CE methods are strongly dependent on the type of data they have to explain (e.g., tabular data or images). Thus, Applying an explainer not specifically designed for graphs would not be able to exploit the graph structure properly. For instance, a  tabular explainer might modify all the edges. Likewise, an image explainer might modify solely edges between contiguous vertices in the graph's adjacency matrix (due to the fact that vertices do not have a fixed order by definition). Second, graphs are highly complex, with many vertices, edges, and attributes associated with both. This specific complexity poses different challenges for generating counterfactuals compared to other data structures. Additionally, interpreting the explanations is a challenging task requiring expertise due to the wide range of domain applications. Finally, when providing counterfactuals for graphs, one should rely on graph-based explainers (e.g., GNNs, GCNs, GATs) that can extract/embed features from neighbouring vertices (see Sec. A). Thus, having a specialised explainer is fundamental, as it happened with the specialisation of neural networks when the fully connected network (successfully applied to tabular data)  evolved in the convolutional one (specialised for images data) and lastly was evolved in the GNN (specialised for graphs data).

Here, we describe the taxonomy of GCE methods - see Fig. \ref{fig:ge_taxonomy} shaped by considering \cite{guidotti2022counterfactual,yuan2022explainability,amara2022graphframex,artelt2019computation,verma2020counterfactual} which, on the other hand, were focused on counterfactual explainability in general. The first level analyse the scope of the GC explainers by dividing them into instance-level and model-level. The former focuses on providing the reasons that make a black-box model reach a specific decision on a particular input instance. Whereas the latter entails producing an explanation of the overall logic of the black-box model assuming that the provided explanation is complete and valid for any instance. In other words, model-level explainability on graphs engages in providing the boundaries of the decision space of explained outcomes. Therefore, having a group of graph instances, model-level explainers aim to determine their collocation in the decision space.

The following level discusses the high-level approach adopted by the explainers dividing them into search-based, heuristic-based, and perturbation methods. Notice that each of these classes are not mutually exclusive (see RCExplainer \cite{bajaj2021robust}) meaning that methods can merge two different strategies into a single one to produce counterfactuals. Search-based methods rely on a specific criterion (e.g., similarity between instances) to search for a counterfactual within the dataset for a given input instance \cite{faber2020contrastive,liu2021multi}. Heuristic-based methods rely on a rigorous policy of modifying the input graph until they reach a valid counterfactual \cite{abrate2021counterfactual,wellawatte2022model}. Without loss of generality, learning methods study the output variations w.r.t. input changes. All methods follow a similar high-level pipeline consisting of 1) generating masks that indicate features of interest given a specific input graph $G$; 2) combining the mask with $G$ to obtain a new graph $G'$ such that the features of interest remain unchanged; 3) feeding $G'$ to the prediction model $\Phi$ and updating the mask according to the outcome $\Phi(G')$. We have identified three sub-classes of learning-based methods: i.e., perturbation matrix \cite{wu2021counterfactual,lucic2022cf,tan2022learning,cai2022probability}, reinforcement learning \cite{numeroso2021meg,nguyen2022explaining}, and generative \cite{ma2022clear,sun2021preserve}. 
Perturbation matrix methods take the desired counterfactual class, the input instance, and the oracle weights and learn a soft-mask that represents which graph features should be present and which removed to get a counterfactual instance. Reinforcement learning methods employ agents that take actions to produce counterfactuals that maximise a user-defined cumulative reward function. Generally, this kind of approach fixes the number of possible actions that the agent can take that lead towards a valid counterfactual (e.g., see MEG \cite{numeroso2021meg}) in a well-defined domain. Generative methods \cite{ma2022clear,sun2021preserve} sample counterfactuals from a learned latent space. 

Lastly, since model-level explainers are recent (see GCFExplainer \cite{huang2023global}), there is only one class that is currently explored: i.e., heuristic-based. Hence, we invite researchers to explore this class of explainers due to its promising future directions (see Sec. \ref{sec:open_challenges}). Note that almost all the methods perform a generation based on perturbing the input instance. Thus, in the taxonomy, we highlight the approach employed by each method, and this must not be confounded with the methods' properties that we discuss in the following section.

\subsection{GCE literature classification}\label{sec:gce_classification}

\begin{table}[ht]
\centering
\caption{Comparison of GCE methods. $\cdot$ depicts a missing aspect; \checkmark depicts a covered aspect; $\sim$ depicts a partially covered aspect; $E$, $V$, and $F$ indicate edge, vertex, and vertex/edge attribute sets, respectively; $+$ and $-$ indicate, respectively, adding and removal operations over the set that precedes them; $*$ indicates perturbation over the set it precedes.}
\label{tab:methods}
\resizebox{\textwidth}{!}{%
\begin{tabular}{@{}lcccccccccl@{}}
\toprule
Method &
  \begin{tabular}[c]{@{}c@{}}Model\\ Agnosticism\end{tabular} &
  \begin{tabular}[c]{@{}c@{}}Model\\ Access\end{tabular} &
 \begin{tabular}[c]{@{}c@{}} Factual-Based\\Explanations\end{tabular} &
  \begin{tabular}[c]{@{}c@{}}Minimal\\ CE\end{tabular} &
  \begin{tabular}[c]{@{}c@{}}Domain\\ Agnosticism\end{tabular} &
  \begin{tabular}[c]{@{}c@{}}Training Data\\ Accessibility\end{tabular} &
  \begin{tabular}[c]{@{}c@{}}Explanation\\ Level\end{tabular} &
  \begin{tabular}[c]{@{}c@{}}Classification\\ Task\end{tabular} &
  \begin{tabular}[c]{@{}c@{}}Generation\\ Type\end{tabular}
   & Approach\\ \midrule
DDBS \cite{abrate2021counterfactual} & \checkmark & $\cdot$   & $\cdot$    & \checkmark & \checkmark & \checkmark & Instance & $G$          & $E(+,-)$               & Heuristic               \\
OBS \cite{abrate2021counterfactual} & \checkmark & $\cdot$   & $\cdot$     & \checkmark & \checkmark & $\cdot$    & Instance & $G$          & $E(+,-)$               & Heuristic    \\
RCExplainer \cite{bajaj2021robust}  & \checkmark & \checkmark & \checkmark & $\cdot$    & \checkmark & \checkmark & Instance & $G$, $V$     & $E(-)$                 & \begin{tabular}[c]{@{}l@{}}Heuristic  \\ \& Learning   \end{tabular}    \\
GNN-MOExp \cite{liu2021multi}       & \checkmark & $\cdot$    & \checkmark & $\cdot$    & \checkmark & $\cdot$    & Instance & $V$          & sub-graph              & Search                \\
MEG \cite{numeroso2021meg}          & \checkmark & \checkmark & $\cdot$    & \checkmark & $\cdot$    & $\sim$     & Instance & $G$          & $E(+,-)$, $V(+,-)$     & Learning   \\
GNNAdv \cite{sun2021preserve}       & \checkmark & \checkmark & $\cdot$    & \checkmark & \checkmark & $\cdot$    & Instance & $G$          & $E(+,-)$               & Learning   \\
CMGE \cite{wu2021counterfactual}    & $\cdot$    & \checkmark & \checkmark & $\cdot$    & $\cdot$    & \checkmark & Instance & $G$          & $E(+,-)$, $V(-)$       & Learning   \\
NSEG \cite{cai2022probability}      & \checkmark & \checkmark & \checkmark & $\cdot$    & \checkmark & $\cdot$    & Instance & $G$, $V$     & $E(-)$, $F(*)$         & Learning\\
CF-GNNExplainer \cite{lucic2022cf}  & \checkmark & \checkmark & $\cdot$    & \checkmark & \checkmark & $\cdot$    & Instance & $V$          & $E(-)$                 & Learning\\
CLEAR \cite{ma2022clear}            & \checkmark & $\cdot$    & $\cdot$    & \checkmark & \checkmark & \checkmark & Instance & $G$          & $E(+,-)$, $F(*)$       & Learning  \\
MACDA \cite{nguyen2022explaining}   & \checkmark & $\cdot$    & $\cdot$    & \checkmark & $\cdot$    & $\sim$     & Instance & $(G_1, G_2)$ & $E(+,-)$, $V(+,-)$     & Learning \\
CF$^2$ \cite{tan2022learning}       & \checkmark & $\cdot$    & \checkmark & $\cdot$    & \checkmark & $\cdot$    & Instance & $G$, $V$     & $E(-)$, $V(-)$, $F(-)$ & Learning\\
MACCS \cite{wellawatte2022model}    & \checkmark & $\cdot$    & $\cdot$    & \checkmark & $\cdot$    & $\cdot$    & Instance & $G$          & $E(+,-)$, $V(+,-)$     & Heuristic \\
GREASE \cite{chen2022grease} & \checkmark & $\cdot$    & \checkmark & \checkmark & $\cdot$    & $\cdot$    & Instance & $E$          & $E(-)$                 & Learning\\
GCFExplainer \cite{huang2023global} & \checkmark & $\cdot$    & $\cdot$    & $\cdot$    & \checkmark & $\cdot$    & Model    & $G$          & $E(+,-)$, $V(+,-)$     & Heuristic\\
\bottomrule
\end{tabular}%
}
\end{table}

Here, we provide the reader with a classification of the GCE methods by identifying more dimensions w.r.t. the other surveys \cite{guidotti2018survey,guidotti2022counterfactual,arrieta2020explainable}, as some of these are related to the graph domain. For each dimension, we provide a brief description and then delve into further detail (see Sec. \ref{sec:graph_counterfactual_explanations}). Table \ref{tab:methods} reports the methods characterised within the chosen dimensions.

\noindent\textbf{Model Agnosticism} - A GCE method relies on a black-box model $\Phi$ to generate counterfactual examples by maximising Eq. \ref{eq:gce_multiclass_sim} or \ref{eq:minimal_gce}. Notice that the desiderata for a counterfactual explainer is to decouple the process of producing counterfactuals $G'$ over an input instance $G$ from the prediction $\Phi(G')$ s.t. $\Phi(G) \neq \Phi(G')$. In other words, model-agnostic counterfactual explainers can be used to explain the outcome of any prediction model $\Phi$. Contrarily, model-specific explainers are intertwined on a particular class of prediction models (e.g., attention-based methods).

\noindent\textbf{Model Access} - GCE methods might require different levels of access to the underlying oracle $\Phi$. According to \cite{verma2020counterfactual}, model access can be embedding- and gradient-wise. Embedding-wise access suggests that the explainer can get the embedding of graph $G$ at a specific layer of $\Phi$. Gradient-wise access consists of obtaining the gradients of $G$ at any layer of $\Phi$. Contrarily to gradient-wise, embedding-wise access does not restrict the explainer to rely only on neural network oracles.

\begin{figure}
    \centering
    \includegraphics[width=0.8\textwidth]{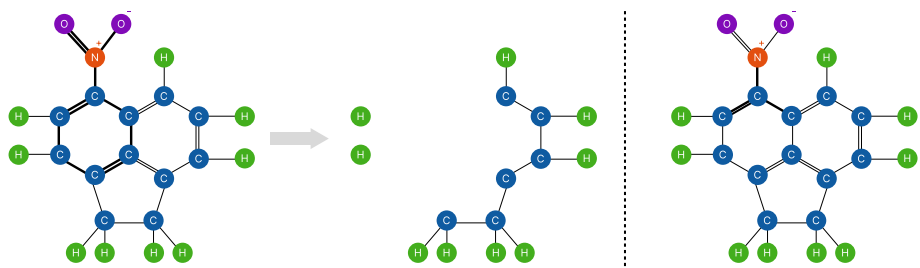}
    \caption{An example of a 5-Nitroacenaphthene molecular structure. (left) Factual example, represented as bold connections, as a Nitrobenzene molecule. (middle) Derived counterfactual example by eliminating the Nitrobenzene molecule. (right) Minimal counterfactual example highlighting the three connections that, if removed, break the Nitrobenzene structure avoiding the molecule being a carcinogen.}
    \label{fig:factual_counterfactual}
\end{figure}

\noindent\textbf{Factual-based explanations} - Some GCE methods use a factual explainer to generate a factual example of $G$ producing the most important features w.r.t. the prediction $\Phi(G)$. Then, it is possible to build a counterfactual example $G'$ by changing the most important features that change the prediction $\Phi(G')$. Notice that these explainers do not guarantee to produce a minimal counterfactual because they do not use an optimisation function to minimise the distance between the generated example $G'$ and the input $G$. Figure \ref{fig:factual_counterfactual} illustrates three different explanations. On the left corner, we depict a factual example - highlighted connections - representing the most important characteristics of the Nitrobenzene molecule. The middle molecular structure illustrates a derived counterfactual example from the factual one on the left. It is not guaranteed that the derived counterfactual example entails a valid molecule. Without loss of generality, derived counterfactuals can only be produced if the original graph was perturbed by eliminating vertices/edges (see Sec. \ref{sec:methods}). For completeness purposes, we illustrate the non-minimality of factual-based counterfactual explainers (compare the molecule in the middle with the one on the right). Here, we can clearly see that the minimal counterfactual has the three connections highlighted that, if removed, would break the Nitrobenzene structure shown on the left, thus, avoiding it to be a carcinogen.

\noindent\textbf{Minimal Counterfactual Example (MCE)} - According to \cite{guidotti2022counterfactual}, generating counterfactuals with minimal changes from the input is paramount. Providing a MCE is useful when giving the end-user tangible explanations. Recall the loan approval example with counterfactual \textit{if the customer were a millionaire, then her loan would be granted}. Notice that this particular example is a useless counterfactual since it does not provide a meaningful explanation. Minimal counterfactual examples ensure that the change in the outcome is guaranteed when the structure of the input slightly differs. In the loan approval example, a minimal counterfactual explanation would be \textit{if the customer had \$1,000 more in her bank account, then her loan would be granted}, and the minimality is represented as the additional \$1,000.

\noindent\textbf{Domain Agnosticism} - Domain-agnostic counterfactual explainers are transferable from one domain to the other. In particular, this kind of explainer can work with heterogeneous graph data (e.g., passing from molecular graphs to human social networks). Contrarily, domain-specific explainers are specialised in generating counterfactuals on only a subclass of graph data (e.g., protein-protein interactions). These explainers, when given in input a graph from a different domain, fail to generalise or even output a valid counterfactual example. The desiderata for robust counterfactual explainers is to be domain-agnostic and decouple from properties valid only in specific cases.

\noindent\textbf{Training Data Accessibility} - GCE methods can access training data to generate meaningful examples or be oblivious and rely only on the prediction $\Phi(G)$ while maximising the similarity between $G'$ and $G$. When trying to preserve privacy in critical domains, accessing further data is not feasible. In these scenarios, we can estimate how the prediction model works by randomly perturbating $G$ to produce a new graph $G'$, and verify whether $\Phi(G') \neq \Phi(G)$. In this way, it is possible to construct an ad-hoc synthetic dataset of graph instances and understand $\Phi$'s functioning while maintaining data privacy. Hence, being oblivious of any additional information besides $G$ is desired for GCE methods. 

\noindent\textbf{Explanation Level} - It refers to the granularity of the explanation. Most explainers are focused on explaining the decision of $\Phi$ for a single instance $G$. On the other hand, model-level \cite{huang2023global} or multiple single-instance explanation methods \cite{prado2022ensemble} provide counterfactuals for a set of input instances to have a high-level understanding of $\Phi$'s decision boundaries.

\noindent\textbf{Classification Task} - There are three classification tasks in graph learning: i.e., vertex, edge, and graph classification. In the literature, graph classification is the most common task, while edge classification has only been explored in \cite{chen2022grease}. Recently, Liu et al. \cite{liu2021multi} provide explanations for graph-pair affinity prediction tasks. In other words, given a drug-target $(G_1,G_2)$ instance, the task is to generate counterfactual $(G'_1, G'_2)$ for both the drug and the target graphs such that $\Phi(G_1,G_2) \neq \Phi(G'_1,G'_2)$. 

\noindent\textbf{Generation Type} - The literature focuses on two main strategies of generating counterfactual examples: i.e,. search-based and perturbation-based generation. Search-based methods find the most similar instance $G'$ from the original dataset\footnote{Having access to either the training data or the embeddings/gradients of the model $\Phi$, makes them unfeasible for privacy preservation.} w.r.t. the input graph $G$ s.t. $\Phi(G) \neq \Phi(G')$ without guarantee that the produced explanation is minimal. Contrarily, perturbation-based approaches perturb $G$ to generate $G'$. Many perturbation-based methods minimise an optimisation function, trying to produce a minimal counterfactual explanation. Perturbations can happen on vertices and edges by either adding or removing them (denoted in the table by + and - respectively) and by changing vertex/edge attributes (denoted by $F(*)$ in the table).

\noindent\textbf{Approach} - We report the high-level approach adopted by the method as in the taxonomy section \ref{sec:gce_in_general_explanations} included in the table \ref{tab:methods} for convenience. While in the future description of each method, we did not report it explicitly to avoid redundancy since the method will be thoughtfully described.

\subsection{GCE Methods}\label{sec:graph_counterfactual_explanations}

Here, we provide a detailed description of the methods in GCE and a discussion on how they cope with the dimensions provided in Table \ref{tab:methods}. In particular, we first describe the method and how it produces counterfactuals. Then, we provide, for each dimension, its advantages and limitations. We report the methods according to their publication year breaking ties according to the alphabetical order of the authors' surnames.

\noindent\textbf{DDBS} and \textbf{OBS} \cite{abrate2021counterfactual} are two heuristic explanation methods designed for brain networks. The brain can be represented as a graph where the vertices represent well-established regions of interest (ROIs) and the edges are links between two co-activated ROIs. Both methods rely on a bidirectional search heuristic. First, they perturb the edges of the input $G$ until they reach a counterfactual $G'$. Second, they rollback some perturbations done in the first stage such that the distance between $G$ and $G'$ decreases while maintaining the counterfactuality condition. OBS chooses to perturb the edges randomly, whereas DBS queries the dataset for the most common edges for each class and uses this information for perturbation purposes. (1)  \textit{Model agnosticism and model access}: Both methods do not access the oracle since they only need to know whether the produced graph $G'$ is a valid counterfactual. Furthermore, since they treat the oracle as a black-box model, they work with any kind of oracle. (2) \textit{Factual-based explanations and minimal CE}: OBS and DBS do not rely on any existing factual explanation method. Both methods focus on finding minimal counterfactuals since, in their second stage, they try to minimise the distance between $G$ and the produced counterfactual. Notice that, given that these methods are heuristics, it is not guaranteed to converge to a minimal counterfactual example. (3) \textit{Domain agnosticism and training data accessibility}: Both methods are designed for adaptability across various domains. However, their original implementation assumes binary classification and uniform data instances with the same vertices throughout the dataset. Fixing the first limitation is feasible with minor tweaks, but the second requires significant changes, affecting performance. OBS relies solely on the input graph $G$, while DDBS needs labeled instances to extract common edges for each class.

(4)  \textit{Classification task and generation type}: Both methods rely on oracles for graph classification tasks to generate counterfactuals via edge addition/removal operations. These edge perturbations are carried on for a user-defined number of times (i.e., upper-bound).

\noindent\textbf{RCExplainer} \cite{bajaj2021robust} generates robust counterfactuals. Firstly, it devises a set of decision regions containing linear decision boundaries obtained by an underlying GNN. RCExplainer exploits an unsupervised approach to find the decision regions for each class. In this way, the linear boundaries of these regions capture the commonalities of the graph instances therein, thus avoiding overfitting due to possible noises/peculiarities of specific instances. Therefore, the produced counterfactuals are robust to noise. Secondly, RCExplainer uses a loss function based on the obtained linear boundaries to train a network that produces a small subset of the edges $E^*$ of the original graph $G$. RCExplainer ensures that the graph $G^* = (V^*,E^*)$ induced by $E^*$ lies on the same decision region as $G$ (i.e., have the same class). Therefore, $G' = G-G^*$ (i.e., removing the edges $E^*$ from $G$) should lie outside this decision region. Hence, $G'$ can be considered a counterfactual. (1) \textit{Model agnosticism and model access}: RCExplainer assumes that the oracle is a GCN for it to access its last convolutional layer to obtain a vector representation of the input graph and, thus, represent the oracle's decision regions/boundaries. However, RCExplainer is adaptable to work with other types of oracles that provide dense graph representations. (2) \textit{Factual-based explanations and minimal CE}: RCExplainer finds a subset of edges $E^*$ such that $G^* = (V^*, E^*)$ induced on $E^*$ is a factual explanation of the input $G$. Consequently, $G' = G - G^*$ is a counterfactual example of $G$. As noticed with CF-GNNExplainer, only removing edges from the input graph does not necessarily produce minimal counterfactuals. (3) \textit{Domain agnosticism and training data accessibility}: The explanation method is not dependent of specific domain knowledge and can be used in diverse application domains. RCExplainer needs a collection of labeled instances to determine the decision regions governing each of the classes predicted by the oracle. (4)  \textit{Classification task and generation type}: RCExplainer explains the oracles' decision on graph classification tasks by removing the edges of $G$ to produce $G'$. However, as with CF-GNNExplainer, RCExplainer does not guarantee to produce valid counterfactual explanations. For instance, assume that an oracle decides whether a graph has cycles or not. If an input $G$ is a tree (i.e., it does not have cycles), RCExplainer would fail to produce a counterfactual $G'$ that has cycles only by removing the edges of $G$. Notice that transforming a tree into a graph with cycles is only possible via edge additions.

\noindent\textbf{GNN-MOExp} \cite{liu2021multi}  is a multi-objective factual-based explanation method for GNN predictions. Although a factual method, GNN-MOExp enforces counterfactual relevance to its factual explanation subgraphs (e.g., CF$^\text{2}$). It searches for a subgraph in the original instance that optimises both the factual and counterfactual features. GNN-MOExp comes with several limitations that limit the expressiveness of the produced counterfactual: i.e., the factual subgraphs are required to be acyclic; the explanation size is specified a priori; it restricts counterfactuals to be subgraphs of the factual ones. (1)  \textit{Model agnosticism and model access}: GNN-MOExp assumes the oracles are black boxes and does not require access to their internal representations. (2)  \textit{Factual-based explanations and minimal CE}: GNN-MOExp is a factual explanation method. It encourages the generation of small factual explanations. However, this is not enough to optimise towards minimal GCEs. Furthermore, it is affected by its impossibility to generate GCEs containing cycles. (3)  \textit{Domain agnosticism and training data accessibility}: The method does not make any particular assumption about the dataset and does not perform training. (4)  \textit{Classification task and generation type}: GNN-MOExp focuses on oracles for vertex classification. For this reason, it searches for the factual explanation subgraphs in the neighbourhood of the current vertex, ranking them according to a the desired metrics. However, this generation mechanism cannot perform edge/vertex additions.

\noindent\textbf{MEG} \cite{numeroso2021meg} is a reinforcement learning approach that produces counterfactuals for a given input molecule. The reward function of the procedure incorporates a task-dependent regularisation term that affects the policy of choosing the next action to perturb the input. Since MEG strives to produce valid counterfactuals, its policy is designed to only chose those actions that lead towards the generation of \Quote{new} valid molecules \cite{zhou2019optimization}. (1)  \textit{Model agnosticism and model access}: MEG accesses the vector representation of molecules used by the oracle to calculate the similarity between instances (i.e., input and produced molecules). Since MEG's original formulation requires its underlying oracle to be a GNN, this method is not model agnostic. However, one can argue, as done with RCExplainer, that MEG can use any type of oracle that can provide learned representations of molecules. (2) \textit{Factual-based explanations and minimal CE}: MEG perturbs the input molecule directly to produce valid counterfactuals, hence, it is not a factual-based approach. Since MEG employs a policy to choose the most promising action to perform on the input molecule, this policy needs to be calibrated according to a reward function that maximises the similarity between the input instance and the produced counterfactual. Thus, MEG produces a minimal counterfactual example. (3)  \textit{Domain agnosticism and training data accessibility}: MEG is designed for the molecular domain and enforces that the graphs are valid molecules. It uses domain knowledge to provide meaningful molecular counterfactuals, thus losing its adaptation flexibility in other domains. Since MEG is a method that is trained/optimised per instance, then it requires access to all the instances one-by-one. However, notice that MEG does not use other instances in the dataset besides the current one to decide the next step to take w.r.t. the reward function that brings to the generation of a new, potentially counterfactual, molecule (hence the $\sim$ in Table \ref{tab:methods}). (4)  \textit{Classification task and generation type}: MEG provides explanations for oracles focusing on the graph classification task. However, the method can also be used to explain the decisions in regression tasks. Additionally, MEG is one of the few methods that employ additions and removals of both vertices and edges in its generation process.

\noindent\textbf{GNNAdv} \cite{sun2021preserve} is the first strategy that generates counterfactual explanations via adversarial attacks. In particular, GNNAdv uses a Topology Attribution Map (TAM) defined with the help of two variables that summarise edge manipulations. Then, it optimises a sparsity-promoting problem over the perturbation variables and exploits a GNN oracle to optimise the TAM. Then, the TAM is sampled according to a Bernoulli distribution to produce the adjacency matrix of the counterfactual explanation.

(1)  \textit{Model agnosticism and model access}: GNNAdv can work with different models as long as they are GNNs. This limitation results from the explainer's need to access the gradients of the oracle to update the TAM.
(2)  \textit{Factual-based explanations and minimal CE}: GNNAdv relies on a reward function that encourages smaller explanations without guaranteeing minimality.
(3)  \textit{Domain agnosticism and training data accessibility}: The explainer does not need direct access to training data. It has also been tested in datasets from different domains.
(4)  \textit{Classification task and generation type}: GNNAdv works on graph classification tasks and generates counterfactuals by only adding/removing edges from the TAM.

\noindent\textbf{CMGE} \cite{wu2021counterfactual} works only for Electronic Health Records (EHRs). First, the authors transform the medical data into a hierarchical graph structure that encodes the relationship between the different types of records. Then, they train a learnable soft-mask matrix to mask the features of vertices/edges in the graph while keeping the decision unaltered. The remaining features - i.e., those that have not been masked - can be considered as supportive to the decision representing a particular health diagnosis. The authors rely on Graph Attention Networks (GATs) as the explainer of their approach despite the fact that they are not intended as interpretable models \cite{jain2019attention}. When generating counterfactual examples, the authors focus on graph classification.
(1) \textit{Model agnosticism and model access}:  Because CMGE relies on the attention mechanism as an explanation method, the explainer and the oracle cannot be decoupled, thus becoming an oracle-specific explanation method. Additionally, CMGE requires complete access to the model logic to verify the attention weights of each feature.
(2) \textit{Factual-based explanations and minimal CE}: Despite the authors claiming to do counterfactual reasoning, CMGE is a perturbation-based factual explanation method. Here, by relying on the attention mechanism, the authors return the top-$k$ most important features, which is not guaranteed to be the smallest possible set to engender counterfactuals (minimality violation).
(3) \textit{Domain agnosticism and training data accessibility}: The authors analyse CMGE's performances only on EHRs. However, we argue that GATs can be adopted to other domains as long as the input data can be represented as graphs. Thus, CMGE is partially domain agnostic. Additionally, CMGE is oracle-specific since it relies on the attention mechanism used by the oracle. Notice that attention mechanisms need to access the network's weights to produce the contribution scores for each input feature. Hence, CMGE accesses the training data to produce counterfactuals.
(4) \textit{Classification task and generation type}: The authors produce counterfactuals via vertex/edge perturbations for graph classification. Recall that CMGE is a factual-based counterfactual explainability method. Therefore, for vertex perturbations on $G$, the authors can only remove vertices because their goal is to find a sub-graph $G^*$ s.t. $\Phi(G) = \Phi(G^*)$. Then, by eliminating $G^*$ from $G$, the authors produce the (possibly) counterfactual $G' = G - G^*$. For edge perturbations, the authors use both adding and removal operations.

\noindent\textbf{NSEG} \cite{cai2022probability} generates necessary and sufficient generations in a similar manner as CF$^\text{2}$. While CF$^\text{2}$ heuristically determines a trade-off between the probability of necessity and that of sufficiency, NSEG maximises their joint lower bound. For this reason, NSEG leverages a continuous mask with a specific sampling strategy, thus producing a discrete adjacency matrix to optimise this lower bound.
(1)  \textit{Model agnosticism and model access}: NSEG can work with different explainers. However, it needs to access the gradients of the overall loss w.r.t. to the masks because it uses them to update the masks and obtain the necessary and sufficient explanation.
(2)  \textit{Factual-based explanations and minimal CE}: NSEG is primarily a factual explanation method that can be used to generate counterfactual explanations, given its emphasis on obtaining explanations with a high probability of necessity. This kind of approach, in general, cannot generate minimal counterfactual explanations.
(3)  \textit{Domain agnosticism and training data accessibility}: The explainer is not tied to perform in any specific domain and does not requires access to training data.
(4)  \textit{Classification task and generation type}: NSEG can be used with oracles targeting both graph and vertex classification tasks. The perturbation method used to generate the mask is based on  \cite{ying2019gnnexplainer}, so NSEG can only remove edges. However, for vertex features, it performs feature value permutation values by only considering the values of the features that exist in other vertices in the input graph.

\noindent\textbf{CF-GNNExplainer} \cite{lucic2022cf} finds a binary perturbation matrix that sparsifies the adjacency matrix of $G$. To find the perturbation matrix, the authors rely on \cite{srinivas2017training} to train sparse neural networks. Additionally, the authors' goal is to remove edges by zeroing the adjacency matrix. CF-GNNExplainer produces counterfactual examples whose distance is the smallest w.r.t. $G$.
(1) \textit{Model agnosticism and model access}: CF-GNNExplainer is model-agnostic and can be used with any oracle. In this way, the explainer does not access any information regarding the inner workings of the oracle. Instead, it just relies on the input instance and the oracle's decision.
(2)  \textit{Factual-based explanations and minimal CE}: CF-GNNExplainer is not based on existing factual explanation techniques. Moreover, it is designed to tackle the minimal GCE problem, returning, from the generated counterfactual examples, the closest one to the input instance. However, notice that it is not always possible to find the minimal counterfactual $\mathcal{E}_\Phi^*(G)$ due to the limited perturbation operations performed on $G$.
(3) \textit{Domain agnosticism and training data accessibility}: CF-GNNExplainer does not use any domain knowledge. Hence, it can be employed on different application domains. Additionally, it does not require labeled data for training purposes.
(4)  \textit{Classification task and generation type}: CF-GNNExplainer produces counterfactuals for oracles trained for the vertex classification task. Hence, the explainer gets a vertex $x$ in input and its ego-network. While other methods of this kind perturb the vertex features, CF-GNNExplainer perturbs $x$'s ego-network until $\Phi$ changes decision. In other words, CF-GNNExplainer strives to change the relationships between instances (e.g., connected vertices) rather than the instances themselves. CF-GNNExplainer can only remove edges from the original graph which is its main limitation since it is not always possible to obtain a valid counterfactual.

\noindent\textbf{CLEAR} \cite{ma2022clear} is a generative GCE method. It relies on a variational autoencoder (VAE) where the encoder maps each input graph $G$ into a latent representation ${Z}$, and the decoder generates the counterfactual based on ${Z}$. The counterfactuals are complete graphs with stochastic weights on the edges\footnote{CLEAR cuts edges according to a Bernoulli distribution over the stochastic adjacency matrix to produce a counterfactual graph.} where the vertex features and graph structure are similar to $G$. The generation of the counterfactuals is conditioned on $G$ and a desired class $c \neq \Phi(G)$. It is important to highlight that, during the decoding process of generating the counterfactual $G'$, the order of the vertices of $G$ is not the same as that in $G'$. Thus, a graph matching step between $G$ and $G'$ is necessary.
(1)  \textit{Model agnosticism and model access}: CLEAR does not require access to the oracle internal representation and can be used with different kinds of oracles.
(2)  \textit{Factual-based explanations and minimal CE}: The method is not a factual-based explanation method. CLEAR's loss function enables the generation of counterfactual graphs that are the closest to the original instance.
(3)  \textit{Domain agnosticism and training data accessibility}: The method is not domain specific and was tested in social networks and molecular datasets. CLEAR requires having access to training data and performing gradient descent on it to learn how to generate counterfactuals on unseen graphs.
(4)  \textit{Classification task and generation type}: CLEAR is designed to explain oracle predictions on graph classification tasks. To generate the explanations, the method performs edge removals/additions and feature value perturbations.

\noindent\textbf{MACDA} \cite{nguyen2022explaining} produces counterfactuals consisting of drug-target pairs for the drug-target affinity (DTA) prediction task relying on multi-agent reinforcement learning. Given in input a pair of instances $(G_1, G_2)$, where $G_1$ is a drug and $G_2$ is a target graph, MACDA generates the counterfactual pair $(G'_1, G'_2)$ such that $G'_1$ is a counterfactual to $G_1$ and $G'_2$ to $G_2$ (i.e., $\Phi(G_1) \neq \Phi(G'_1) \;\land\; \Phi(G_2) \neq \Phi(G'_2)$). Notice that the counterfactual pair is produced simultaneously.
(1)  \textit{Model agnosticism and model access}: The method assumes the oracle is a black-box and only accesses its output.
(2)  \textit{Factual-based explanations and minimal CE}: MACDA is designed to generate counterfactual explanations, and the loss function of the agents encourages the method to produce counterfactuals close to the original instance.
(3)  \textit{Domain agnosticism and training data accessibility}: The framework is designed to work with molecular graphs and specifically for the drug-target affinity prediction task, not easily adaptable to other domains. As in MEG, MACDA needs to generate the next possible actions w.r.t. a specific reward function for each instance. Therefore, it does not need to access training data to generate a counterfactual for the current instance $G$, but it optimises the actions to take for $G$ specifically.
(4)  \textit{Classification task and generation type}: The DTA prediction task is a classification task that consists of predicting the strength (binding affinity) of a drug molecule and a target protein. Usually, the binding affinity is categorised as zero, low, medium, and high. As with MEG, MACDA adds/removes both vertices and edges in its generation process.

\noindent\textbf{CF$^\text{2}$} \cite{tan2022learning} produces factual explanations by balancing factual and counterfactual reasoning. It solves a multi-objective optimisation problem where the generated counterfactuals need to adhere to specific constraints. According to the proposed factual reasoning, the factual graph is a subgraph of the input instance. Hence, the counterfactual explanation is the input graph without the factual subgraph, as in the case of RCExplainer. CF$^\text{2}$ takes into consideration the simplicity of the counterfactual (i.e., the smaller the explanation size, the better). Despite being a factual method, we include CF$^\text{2}$ because it inherently has a counterfactual property of eliminating the factual subgraph.
(1) \textit{Model agnosticism and model access}: CF$^\text{2}$ does not access the oracle's latent representations to produce counterfactuals, and it can be used with any kind of oracle.
(2) \textit{Factual-based explanations and minimal CE}: CF$^2$ is a perturbation-based factual method. However, it requires the explanations to comply with the counterfactual property, meaning that removing the factual subgraph from the original graph must produce a change in the prediction. As mentioned for RCExplainer, subtracting a factual subgraph from the original graph does not guarantee producing a minimal counterfactual explanation.
(3)  \textit{Domain agnosticism and training data accessibility}: The authors of CF$^2$ provide experimental results demonstrating the effectiveness of their method on synthetic, citation, and molecular datasets. In this way, CF$^2$ is adaptable to any domain as long as it encompasses graph data. Additionally, the explainer does not require labeled data to produce a counterfactual example since the explanation produced is a mere derivation of the removal of the factual subgraph from the original instance.
(4)  \textit{Classification task and generation type}: CF$^\text{2}$ can to explain oracle predictions on graph and vertex classification tasks. It does edges, vertex, and vertex attribute removals on the original graph to generate counterfactuals. A limitation of this approach is that edge and vertex additions are not considered because the factual explanations are a subset of the input graph, while, in general, counterfactuals are not.

\noindent\textbf{MACCS} \cite{wellawatte2022model} works in the molecular domain. The method takes in input a molecular graph represented as SELFIES (\textit{SELF}-referenc\textit{I}ng \textit{E}mbedded \textit{S}trings) \cite{krenn2020self}. MACCS builds on top of the STONED protocol \cite{nigam2021beyond}, which rapidly explores the chemical space without relying on pre-trained generative models or reaction rules. The STONED protocol consists of string insertion, deletion, and modification steps that can generate valid perturbed molecules that are close in the chemical space w.r.t. the input molecule. After expanding the chemical space around the original molecular graph, MACCS identifies similar counterfactuals with a changed prediction, selecting a small number of these using clustering and the Tanimoto similarity. By clustering the counterfactual examples and selecting, for each cluster, the closest counterfactual to the original molecule, MACCS returns multiple counterfactuals that are different from each other.
(1) \textit{Model agnosticism and model access}: MACCS is model agnostic. Neither the STONED method nor the clustering/similarity processes need access to the oracle's internal information to produce counterfactuals. 
(2)  \textit{Factual-based explanations and minimal CE}: MACCS is not based on factual explanation approaches. The explainer returns multiple counterfactual examples ordered by their proximity w.r.t. the original molecular graph. Furthermore, MACCS has an upper-bound on the modifications the counterfactual molecules can have to ensure they are located nearby in the chemical space.
(3)  \textit{Domain agnosticism and training data accessibility}: By employing the STONED protocol, MACCS can generate counterfactuals without needing labeled training data. Contrarily, MACCS is domain-specific, being restricted to the molecular domain. This poses an important limitation to the adoption of MACCS on other domains since instances (molecules) are represented as SELFIES strings. Consequently, the mutations (i.e., changes in the molecular structure) are restricted to an alphabet specific to the molecular domain.
(4)  \textit{Classification task and generation type}: MACCS is designed to explain the decisions of oracles on molecular graph classification. Since STONED can token deletions, replacements, and insertion on the SMILES representation on molecules, MACCS can, thus, modify vertices and edges on the original graph.

\noindent \textbf{GREASE} \cite{chen2022grease} is designed to explain user-item recommendations, which is equivalent to explaining the existence of edges in a bipartite graph composed by user and item vertices. The authors train a surrogate model to find an optimal perturbation mask without accessing the original black-box model.
(1) \textit{Model agnosticism and model access}: Thanks to the surrogate model, no knowledge about the original model is necessary (i.e., no gradient access).
(2) \textit{Factual-based explanations and minimal CE}: The method performs factual and counterfactual explanation generation separately. However, using a common definition for both processes limits the counterfactual examples to be a subgraph of the original graph. The number of changes performed to the original instance to produce the counterfactual explanation is considered, thus encouraging minimal explanations.
(3) \textit{Domain agnosticism and training data accessibility}: GREASE is designed for explainability in recommender systems and not in general graph data. The method does not require training data.
(4) \textit{Classification task and generation type}: The classification task could be equivalent to edge classification but also takes into account the rankings used by the recommender system. The method only performs edge removals during the explanation generation process.

\noindent\textbf{GCFExplainer} \cite{huang2023global} is a model-level GCE method. It takes in input a set of graphs $\mathcal{G}$ belonging to a class $c$, an oracle $\Phi$, a distance boundary $\theta$, and a budget $k$. It, then, produces a set of counterfactual graphs $\mathcal{G}'$ with $|\mathcal{G}'| = k$ s.t. it maximises the number of graphs $G \in \mathcal{G}$ that have a counterfactual $G' \in \mathcal{G}'$ within an edit distance lower than $\theta$ (see Eq. \ref{eq:kosan_new}). In other words, GCFExplainer maximises the coverage of the explanation set. First, GCFExplainer organises the search space as a meta-graph where the vertices are all the graphs that can be obtained by performing no more than $\theta$ edits on any input $G$, and edges connect graphs that are at one edit distance. After building this search space, GCFExplainer relies on vertex-reinforced random walks \cite{pemantle1992vertex} to obtain a set of counterfactuals prioritising those that most instances can reach. Lastly, GCFExplainer uses a greedy algorithm to select a set $\mathcal{G}'$ of size $k$ that maximises $\mathcal{G}$'s coverage.
(1)  \textit{Model agnosticism and model access}: GCFExplainer does not require access to the oracle internal representation. Furthermore, it can be used with different kinds of oracles as it only needs access to the prediction.
(2)  \textit{Factual-based explanations and minimal CE}: GCFExplainer is specifically designed to produce counterfactual explanations and it is not based on any existing factual explainer. The method does not search for a minimal counterfactual explanation. However, it permits manually setting the maximum allowed edit distance $\theta$ to produce counterfactuals.
(3)  \textit{Domain agnosticism and training data accessibility}: The method was tested only on molecular datasets. However, there are no domain-specific restrictions that prevent its application to different domains. The explainer does not perform any kind of training, thus, it does not need access to training data.
(4)  \textit{Classification task and generation type}: GCFExplainer is designed to explain the oracle prediction on graph classification tasks. To generate the explanations, the method performs vertex and edge removals/additions.

\section{Evaluation}\label{sec:eval}
A fundamental aspect of successful research is to provide evidence of the effectiveness of the proposed solution. This evaluation is done through quantitative assessments, including standardised tests and various measurements, following a fixed experimental protocol. In the case of GCE methods, a general evaluation protocol involves the following steps. First, the oracle is trained on a selected dataset. Second, the explainer, to be evaluated, takes the trained oracle and a selected instance as inputs and produces an explanation. Third, the produced explanation and the runtime traces are evaluated using several evaluation measures. This process can be repeated for different datasets, explainers, and oracles, aggregating the resulting measures in tables and plots. 

In Sec. \ref{sec:datasets}, we describe the datasets adopted in the literature. We refer the reader to Sec. C for a detailed description of the characteristics of each of them. In Sec. \ref{sec:evaluation_metrics}, we summarise the evaluation metrics used in GCE and argue that multiple metrics should be used for fair performance evaluations. Lastly, Sec. \ref{sec:eval:startegies} illustrates the evaluation protocol adopted in each SoA work.

\begin{table}[!t]
\centering
\caption{The datasets used in the literature alongside their domains, the link to their repository, and the papers they are used by.}
\label{tab:datasets}
\resizebox{.95\textwidth}{!}{%
\begin{tabular}{@{}llll@{}}
\toprule
Dataset &
  Domain &
  Publicly Available Repository (Data or Code)&
  Used by \\ \midrule
Tree-Cycles \cite{ying2019gnnexplainer} &
  synthetic &
  \url{https://github.com/RexYing/gnn-model-explainer} &
  \cite{bajaj2021robust,tan2022learning,lucic2022cf,cai2022probability} \\
Tree-Grid \cite{ying2019gnnexplainer} &
  synthetic &
  \url{https://github.com/RexYing/gnn-model-explainer} &
  \cite{bajaj2021robust,lucic2022cf,cai2022probability} \\
Tree-Infinity & 
  synthetic & \url{https://github.com/MarioTheOne/GRETEL}& \cite{prado2022gretel}
   \\
BA-Shapes \cite{ying2019gnnexplainer} &
  synthetic &
  \url{https://github.com/RexYing/gnn-model-explainer} &
  \cite{bajaj2021robust,tan2022learning,lucic2022cf,cai2022probability} \\
BA-Community \cite{ying2019gnnexplainer} &
  synthetic &
  \url{https://github.com/RexYing/gnn-model-explainer} &
  \cite{bajaj2021robust} \\
BA-2motifs \cite{luo2020parameterized} &
  synthetic &
  \url{https://github.com/flyingdoog/PGExplainer} &
  \cite{bajaj2021robust,tan2022learning} \\
ADHD \cite{brown2012ucla} &
  -omics &
  \url{https://github.com/MarioTheOne/GRETEL/tree/main/data/datasets/adhd} &
  \cite{abrate2021counterfactual} \\
ASD \cite{craddock2013neuro,lanciano2020explainable} &
  -omics &
  \url{https://github.com/MarioTheOne/GRETEL/tree/main/data/datasets/autism/asd} &
  \cite{abrate2021counterfactual} \\
BBBP \cite{martins2012bayesian} &
  molecular &
  \url{https://www.kaggle.com/datasets/mmelahi/cheminformatics?select=bbbp.zip} &
  \cite{wellawatte2022model} \\
HIV \cite{hiv2019dataset,hiv2019kaggle,riesen2008iam} &
  molecular &
\url{https://www.kaggle.com/datasets/mmelahi/cheminformatics?select=hiv.zip} &
  \cite{wellawatte2022model,huang2023global} \\
Ogbg-molhiv \cite{hu2020open} & molecular & \url{https://huggingface.co/datasets/OGB/ogbg-molhiv} & \cite{ma2022clear} \\
Mutagenicity \cite{kazius2005derivation} &
  molecular &
  \url{https://ls11-www.cs.tu-dortmund.de/people/morris/graphkerneldatasets/Mutagenicity.zip} &
  \cite{bajaj2021robust,tan2022learning,huang2023global} \\
NCI1 \cite{wale2008comparison} &
  molecular &
  \url{https://ls11-www.cs.tu-dortmund.de/people/morris/graphkerneldatasets/NCI1.zip} &
  \cite{bajaj2021robust,tan2022learning,huang2023global} \\
  TOX21 \cite{kersting2016benchmark} &
  molecular &
  \url{https://tripod.nih.gov/tox21/challenge/data.jsp} &
  \cite{numeroso2021meg} \\
ESOL \cite{wu2018moleculenet} &
  molecular &
  \url{https://github.com/deepchem/deepchem} &
  \cite{numeroso2021meg,tan2022learning} \\
Proteins \cite{borgwardt2005protein} &
  molecular &
  \url{https://chrsmrrs.github.io/datasets/docs/datasets/} &
  \cite{huang2023global} \\
Davis \cite{davis2011comprehensive} &
  molecular &
  \url{http://staff.cs.utu.fi/~aatapa/data/DrugTarget/} &
  \cite{nguyen2022explaining} \\
PDBBind \cite{wang2004pdbbind} &
  molecular &
  \url{http://www.pdbbind.org.cn/} &
  \cite{nguyen2022explaining} \\
CiteSeer \cite{giles1998citeseer} &
  social &
  \url{https://linqs.org/datasets/} &
  \cite{tan2022learning,liu2021multi} \\
IMDB-M \cite{yanardag2015deep} & 
  social & 
  \url{https://virginia.app.box.com/s/941v9pwh83lfw5vnwfbgcertlsoivg5j} & 
  \cite{ma2022clear} \\
CORA \cite{mccallum2000automating} & 
  social & 
  \url{https://relational.fit.cvut.cz/dataset/CORA} & 
  \cite{liu2021multi} \\
Musae-Facebook \cite{rozemberczki2019multiscale} & 
  social & 
  \url{https://www.kaggle.com/datasets/rozemberczki/musae-facebook-pagepage-network} & 
  \cite{liu2021multi} \\
LastFM \cite{rozemberczki2020characteristic} & 
  social & 
  \url{https://github.com/gusye1234/LightGCN-PyTorch/tree/master/data/lastfm} & 
  \cite{chen2022grease} \\
Yelp \cite{wang2019neural} & 
  social & 
  \url{https://github.com/gusye1234/LightGCN-PyTorch/tree/master/data/yelp2018/} & 
  \cite{chen2022grease} \\
  \bottomrule
\end{tabular}%
}
\end{table}

\subsection{Datasets adopted in the literature}\label{sec:datasets}
Due to the absence of standardised and well-established benchmarks in the literature, comparing the approaches presented in Sec. \ref{sec:methods} is difficult. The surveyed studies typically compare themselves with simple baselines rather than SoA solutions by adopting a heterogeneous set of synthetically generated ad-hoc datasets not part of an established benchmark. We provide Table \ref{tab:datasets} as an index of available datasets to bridge this gap for future researchers. Thus, researchers can adopt them to evaluate and compare their performances with those proposed in the literature.

Synthetic datasets are structured and generated through specific algorithms\footnote{Usually, synthetic graph datasets are generated via programmed processes that rely on specific algorithms and constraints.} and constraints, making them easier to control and modify. They allow for easy evaluation of performance since properties are well-known. Additionally, they enable the identification of the minimum counterfactual example\footnote{Recall that the minimality of a counterfactual example does not strictly depend on the explainer. Rather, the oracle classifying the generated examples is important in considering them counterfactual. In cases where the explainer performs well, but the oracle lacks, the literature has proposed to assess the oracle's performance by measuring its accuracy.} for a given input instance and the computing of the minimal distance between the two. This approach can be applied to classification tasks such as distinguishing cyclic and acyclic graphs. The generation of synthetic datasets follows \cite{ying2019gnnexplainer}, and it involves three steps presented in its binary version for readability: (1) generate a base graph with specific characteristics, including the number of vertices and edge; (2) generate well-known motifs or use handcrafted ones if preferred, ensuring that the base graph does not already contain the motif to be added; (3) connect the chosen motifs to the base graph while controlling for additional similar motifs. The resulting dataset can have two classes, graphs generated solely by the first step labeled 0 and those following all three steps labeled 1. Repeat the second step for the desired number of classes for multi-class scenarios. Researchers can choose the number of instances and their distribution among classes.
\textit{Tree-Cycles} \cite{ying2019gnnexplainer}, \textit{Tree-Grid} \cite{ying2019gnnexplainer}, \textit{Tree-Infinity} \cite{prado2022gretel}, \textit{BA-Shapes} \cite{ying2018graph}, \textit{BA-2motifs} \cite{luo2020parameterized}, \textit{BA-Community} \cite{ying2019gnnexplainer} are the synthetic datasets adopted in the literature. Besides these synthetic datasets, the literature relies on real datasets. The real datasets can be divided into three domains: i.e., -omics (\cite{brown2012ucla,craddock2013neuro,lanciano2020explainable}), molecular (\cite{wellawatte2022model,hiv2019dataset,hiv2019kaggle,riesen2008iam,hu2020open,kazius2005derivation,kersting2016benchmark,wale2008comparison,wu2018moleculenet,borgwardt2005protein}), and social networks (\cite{giles1998citeseer,yanardag2015deep}). We point the reader to Sec. B for detailed descriptions of the datasets provided in Table \ref{tab:datasets} and their characteristics.

\subsection{Evaluation Metrics}\label{sec:evaluation_metrics}

Table \ref{tab:metrics} summarises the metrics used in the literature to evaluate the performance of a GCE method. Since multiple factors can affect the quality of explanations, it is better to employ many simple metrics instead of a complex ones, such as fidelity, to fully understand the different facets of the explainer's behaviour.
Moreover, for completeness purposes, we group some metrics according to two classes -  i.e., Minimality Evaluation and Global Metrics - and report two ideal metrics (Diversity and Actionability) that are not used in the literature. We believe these metrics can be useful to uncover more insight about the performances of GCE.
\begin{table}[!ht]
\centering
\caption{Evaluation metrics used in the literature. \checkmark indicates that the measure is used by that paper, $\cdot$ if it is not used.}
\label{tab:metrics}
\resizebox{\textwidth}{!}{%
\begin{tabular}{@{}l@{\hspace{.2\tabcolsep}}ll@{\hspace{.2\tabcolsep}}cccccccccccccc@{}}
\toprule
\multicolumn{3}{l}{} & \multicolumn{1}{l}{DDBS \& OBS \cite{abrate2021counterfactual}} & \multicolumn{1}{l}{RCExplainer \cite{bajaj2021robust}} & \multicolumn{1}{l}{GNN-MOExp \cite{liu2021multi}} & \multicolumn{1}{l}{MEG \cite{numeroso2021meg}} & \multicolumn{1}{l}{GNNAdv \cite{sun2021preserve}} & \multicolumn{1}{l}{CMGE \cite{wu2021counterfactual}} & \multicolumn{1}{l}{NSEG \cite{cai2022probability}} & \multicolumn{1}{l}{CF-GNNExplainer \cite{lucic2022cf}} & \multicolumn{1}{l}{CLEAR \cite{ma2022clear}} & \multicolumn{1}{l}{MACDA \cite{nguyen2022explaining}} & \multicolumn{1}{l}{CF$^2$ \cite{tan2022learning}} & \multicolumn{1}{l}{MACCS \cite{wellawatte2022model}} & \multicolumn{1}{l}{GREASE \cite{chen2022grease}} & \multicolumn{1}{l}{GCFExplainer \cite{huang2023global}} \\ \midrule
\multicolumn{3}{l}{Runtime}            & $\cdot$    & \checkmark & $\cdot$    & $\cdot$ & $\cdot$    & $\cdot$    & $\cdot$    & $\cdot$    & \checkmark & $\cdot$    & $\cdot$    & $\cdot$ & $\cdot$    & \checkmark \\
\multicolumn{3}{l}{Oracle Calls}       & \checkmark & $\cdot$    & $\cdot$    & $\cdot$ & $\cdot$    & $\cdot$    & $\cdot$    & $\cdot$    & $\cdot$    & $\cdot$    & $\cdot$    & $\cdot$ & $\cdot$    & $\cdot$ \\
\multicolumn{3}{l}{Oracle Accuracy}    & $\cdot$    & $\cdot$    & $\cdot$    & $\cdot$ & $\cdot$    & $\cdot$    & $\cdot$    & $\cdot$    & $\cdot$    & $\cdot$    & $\cdot$    & $\cdot$ & $\cdot$    & \checkmark \\
\multicolumn{3}{l}{Correctness}        & $\cdot$    & $\cdot$    & \checkmark & $\cdot$ & $\cdot$    & $\cdot$    & \checkmark & $\cdot$    & \checkmark & $\cdot$    & \checkmark & $\cdot$ & \checkmark & $\cdot$ \\
\multicolumn{3}{l}{Sparsity}           & $\cdot$    & $\cdot$    & $\cdot$    & $\cdot$ & $\cdot$    & $\cdot$    & $\cdot$    & \checkmark & $\cdot$    & $\cdot$    & $\cdot$    & $\cdot$ & $\cdot$    & $\cdot$ \\
\multicolumn{3}{l}{Fidelity}           & $\cdot$    & \checkmark & $\cdot$    & $\cdot$ & $\cdot$    & $\cdot$    & \checkmark & \checkmark & $\cdot$    & $\cdot$    & $\cdot$    & $\cdot$ & $\cdot$    & $\cdot$ \\
\multicolumn{3}{l}{Robustness}         & $\cdot$    & \checkmark & \checkmark & $\cdot$ & $\cdot$    & $\cdot$    & $\cdot$    & $\cdot$    & $\cdot$    & $\cdot$    & $\cdot$    & $\cdot$ & $\cdot$    & $\cdot$ \\
\multicolumn{3}{l}{Explainer Accuracy} & $\cdot$    & $\cdot$    & $\cdot$    & $\cdot$ & \checkmark & \checkmark & \checkmark & \checkmark & $\cdot$    & $\cdot$    & \checkmark & $\cdot$ & $\cdot$    & $\cdot$ \\
\multicolumn{3}{l}{Prediction Distance}& $\cdot$    & $\cdot$    & $\cdot$    & $\cdot$ & $\cdot$    & $\cdot$    & $\cdot$    & $\cdot$    & $\cdot$    & \checkmark & $\cdot$    & $\cdot$ & $\cdot$    & $\cdot$ \\
\multicolumn{3}{l}{Causality}          & $\cdot$    & $\cdot$    & $\cdot$    & $\cdot$ & $\cdot$    & $\cdot$    & $\cdot$    & $\cdot$    & \checkmark & $\cdot$    & $\cdot$    & $\cdot$ & $\cdot$    & $\cdot$ \\
\multicolumn{3}{l}{Diversity}          & $\cdot$    & $\cdot$    & $\cdot$    & $\cdot$ & $\cdot$    & $\cdot$    & $\cdot$    & $\cdot$    & $\cdot$    & $\cdot$    & $\cdot$    & $\cdot$ & $\cdot$    & $\cdot$ \\
\multicolumn{3}{l}{Actionability}      & $\cdot$    & $\cdot$    & $\cdot$    & $\cdot$ & $\cdot$    & $\cdot$    & $\cdot$    & $\cdot$    & $\cdot$    & $\cdot$    & $\cdot$    & $\cdot$ & $\cdot$    & $\cdot$ \\ 
\midrule
\multirow{4}{*}{\rotatebox{90}{Minimality}} & \multirow{4}{*}{\rotatebox{90}{Evaluation}} & GED & \checkmark & $\cdot$ & $\cdot$ & \checkmark & $\cdot$  & $\cdot$    & $\cdot$    & $\cdot$ & \checkmark & \checkmark & $\cdot$ & \checkmark & $\cdot$ & $\cdot$ \\
 &  & Explanation Size & $\cdot$ & $\cdot$ & $\cdot$ & $\cdot$ & $\cdot$ & $\cdot$ & $\cdot$ & \checkmark & $\cdot$ & $\cdot$ & $\cdot$ & $\cdot$ & \checkmark & $\cdot$ \\
 &  & Tanimoto Similarity & $\cdot$ & $\cdot$ & $\cdot$ & $\cdot$ & $\cdot$ & $\cdot$ & $\cdot$ & $\cdot$ & $\cdot$ & $\cdot$ & $\cdot$ & \checkmark & $\cdot$ & $\cdot$ \\
 &  & MEG Similarity & $\cdot$ & $\cdot$ & $\cdot$ & \checkmark & $\cdot$ & $\cdot$ & $\cdot$ & $\cdot$ & $\cdot$ & $\cdot$ & $\cdot$ & $\cdot$ & $\cdot$ & $\cdot$ \\ \midrule
\multirow{3}{*}{\rotatebox{90}{Global}} & \multirow{3}{*}{\rotatebox{90}{Metrics}} & Recourse Cost & $\cdot$ & $\cdot$ & $\cdot$ & $\cdot$ & $\cdot$ & $\cdot$ & $\cdot$ & $\cdot$ & $\cdot$ & $\cdot$ & $\cdot$ & $\cdot$ & $\cdot$ & \checkmark \\
 &  & Coverage & $\cdot$ & $\cdot$ & $\cdot$ & $\cdot$ & $\cdot$ & $\cdot$ & $\cdot$ & $\cdot$ & $\cdot$ & $\cdot$ & $\cdot$ & $\cdot$ & $\cdot$ & \checkmark \\
 &  & Compactness & $\cdot$ & $\cdot$ & $\cdot$ & $\cdot$ & $\cdot$ & $\cdot$ & $\cdot$ & $\cdot$ & $\cdot$ & $\cdot$ & $\cdot$ & $\cdot$ & $\cdot$ & \checkmark \\ 
 \bottomrule
\end{tabular}%
}
\end{table}

\noindent\textbf{Runtime} measures the time taken by the explainer to produce the counterfactual example. It provides an efficient way of measuring the explainer's efficiency, including the execution time of the oracle. The runtime can discriminate a \textit{good} explainer based on the sluggishness of its oracle. This measure must be performed in isolation on the same hardware and software platform.

\noindent \textbf{Oracle Calls} \cite{abrate2021counterfactual} measures the number of times the explainer must ask the oracle to produce a counterfactual. This metric is similar to the runtime, but it is useful to evaluate the explainer's computational complexity performances, especially in a distributed system, avoiding considering latency and throughput, which are critical exogenous factors of the measurement.

\noindent \textbf{Oracle Accuracy} evaluates the oracle's reliability in predicting outcomes. The accuracy of the employed oracle significantly influences the quality of explanations, as the explainer aims to elucidate the model's behaviour. Only accurate model predictions will result in meaningful explanations provided to the user. Mathematically, for a given input $x$ and its true label $y_x$, accuracy is defined as $\chi(x) = \mathbb{1}[\Phi(x) = y_x]$. 

\noindent \textbf{Correctness} (Validity) \cite{guidotti2022counterfactual,prado2022gretel} indicates whether the explainer is capable of producing a valid counterfactual explanation (i.e., the example has a different classification from the original instance). More formally, given the original instance $x$, the produced example $x'$, and oracle $\Phi$, the correctness is an indicator function $\Omega(x,x') = \mathbb{1}[\Phi(x) \neq \Phi(x')]$.
 
Correctness can also be referred as \textit{Probability of Necessity} when adopted for multiple counterfactual explanations \cite{tan2022learning} for a given instance, as $\frac{\sum_{x' \in X'}\mathbb{1}[\Phi(x') \neq \Phi(x)]}{|X'|}$.

\noindent
\textbf{Sparsity} \cite{yuan2022explainability} measures the similarity between the input instance and its counterfactual according to the input attributes. Recall that an instance $x$ can be a vertex, edge, or graph, with its associated counterfactual $x'$ and an attribute set of length $|x|$. 

Assuming that $\mathcal{D}_{inst}(x,x')$ is the distance between $x$ and $x'$, we adapt the original definition of sparsity to be $\frac{\mathcal{D}_{inst}(x,x')}{|x|}$ for graphs.

\noindent\textbf{Fidelity} \cite{yuan2022explainability} measures how faithful the explanations are to the oracle considering their correctness. Given the input $x$, its true label $y_x$, and its counterfactual $x'$, fidelity is defined as $\Psi(x,x') = \chi(x) - \mathbb{1}[\Phi(x') = y_x]$. Notice that $\Psi(x, x')$ can assume three values. A value of $\mathbf{1}$ entails that both the explainer and oracle are working correctly. It is trivial to verify that $\chi(x)$ needs to produce a 1 (i.e. $\Phi(x) = y_x$) and the indicator a 0 (i.e. $\Phi(x') \neq y_x$). $\mathbf{0}$ and $\mathbf{-1}$ describe something wrong with the explainer or the oracle. However, we cannot attribute the incorrect function to the explainer or the oracle. This is a shortcoming of fidelity since it bases the assessment of the correctness on the ground truth label $y_x$ instead of the prediction $\Phi(x)$.

\noindent The \textbf{Robustness} \cite{bajaj2021robust} of an explanation quantifies its resistance to changes induced by adding noise to the input graph. A perturbed graph $G_p$ is produced to measure robustness by adding random noise to the vertex features of $G$ and randomly adding or deleting edges while ensuring that the oracle prediction remains unchanged. The extent of change in the explanations $G'$ and $G_p'$ is then computed. Bajaj et al. \cite{bajaj2021robust} consider the top $k$ edges of $G'$ as the ground truth and compare $G_p'$ against them. They evaluate the robustness using AUC-ROC. It is worth noting that their approach assumes that the explainer assigns importance scores to the graph's edges. One can use different functions to assess robustness to calculate the similarity between $G'$ and $G_p'$. The metric is also referred to as \textit{Instability} \cite{guidotti2022counterfactual}.

\noindent\textbf{Explainer Accuracy} is the proportion of explanations that are ``correct''. Works focused on explaining node classifiers \cite{lucic2020does,tan2022learning} follow the evaluation idea used in \cite{ying2019gnnexplainer} for nodes that are originally labeled as being part of a specific motif. Then, an explanation is considered correct if it only includes edges inside the motif. Accuracy can only be computed for instances where a ground truth explanation is known.

\noindent\textbf{Prediction Distance} \cite{nguyen2022explaining} is the distance between the predicted outcome of $x$ and that of $x'$, i.e., $|\mathcal{D}_{pred}(\Phi(x), \Phi(x'))|$.

\noindent \textbf{Causality } \cite{ma2022clear} measures if the changes from $x$ to $x'$ are consistent with the underlying structural causal model that describes the causal relations among different variables (e.g., node features, degree, etc.) present in the data.

\noindent \textbf{Diversity} \cite{guidotti2022counterfactual} quantifies how diverse a set of counterfactuals is, i.e., $Div(X') = \frac{1}{|X'|} \sum_{x' \in X'} \sum_{x'' \in X'} \mathcal{D}_{inst}(x', x'')$.

\noindent\textbf{Actionability} \cite{guidotti2022counterfactual} accounts for the counterfactuals $x' \in X'$ that are feasible in practice for a specific instance $x$, given the set $\mathcal{A}$ of actionable features. Feasible counterfactuals are those that do not consider features (e.g., race, gender) outside of $\mathcal{A}$. Actionability is defined as $Act(x, X') = \frac{|\{x' \in X'\;|\; \mathbb{1}(\mathcal{A}(x',x))\}|}{|X'|}$ where $\mathbb{1}(\mathcal{A}(x',x))$ gives 1 if $x'$ is actionable w.r.t. the list of actionable features $\mathcal{A}$, and 0, otherwise.

\noindent\textbf{Minimality Evaluation} studies the minimality of the counterfactual examples (cf. Def. \ref{def:minimal_gce}). Notice that, to provide a minimal counterfactual example, one needs to define metrics of similarity/distances between the generated counterfactual $G'$ and the input graph $G$. Here, we report each metric in this category (i.e., \textit{Graph Edit Distance (GED)}, \textit{Explanation Size},  \textit{Tanimoto Similarity}, and \textit{MEG Similarity }), and comment on them accordingly. \textbf{Graph Edit Distance (GED)} quantifies the structural distance between the original graph $G$ and its counterfactual $G'$. The distance is evaluated based on a set of actions $\{p_1,p_2,\dots,p_n\} \in \mathcal{P}(G,G')$, representing a path to transform $G$ into $G'$ over all possible paths of actions $\mathcal{P}(G,G')$. The path consists of adding or removing vertices or edges, and each action $p_i$ in the path is associated with a cost $\omega(p_i)$. Given $G$, $G'$, and the set of actions, the GED is computed as 
$$\min_{\{p_1,\dots,p_n\} \in \mathcal{P}(G,G')} \sum_{i = 1}^n \omega(p_i)$$
Typically, when presented with two counterfactual examples, we prefer one closer to the original instance $G$, as it provides shorter action paths on $G$ to change the oracle's output. Notice that GED yields a global measure, and a relative metric that considers the instance size, such as sparsity, can be employed to evaluate the explainer's performance over multiple instances. \textbf{Explanation Size} \cite{lucic2022cf} depicts the difference between the original graph $G$ and the counterfactual $G'$. Similarly to GED, it considers counterfactual explanation as a set of edit actions $\{p_1,\dots,p_n\} \in \mathcal{P}(G, G')$ to perform on $G$ to transform it into $G'$. The explanation size is equal to the number of edit actions, i.e., $|\{p_1,\dots,p_n\}| = n$. \textbf{Tanimoto Similarity } \cite{bajusz2015tanimoto} calculates the similarity between two molecule graphs represented as binary vectors. Hence, given $G = (V,B,E)$ and $G'=(V,B',E')\text{ s.t. } B,B' \in \{0,1\}^{|V|}$, $\tau(G, G') = \frac{\sum_{i=1}^{n} B_i \cdot B'_i}{\sum_{i=1}^{n} B_i + \sum_{i=1}^{n} B'_i - \sum_{i=1}^{n} B_i \cdot B'_i}$. Lastly, \textbf{MEG Similarity} \cite{numeroso2021meg} is a convex combination of $\tau(G,G')$ and the cosine similarity of the graphs $G$ and $G'$.

\noindent \textbf{Global Metrics} measure the explainer's performance at the dataset level (global explanations).  Thus, the metrics discussed here assume that there is a single explanation set $X'$ for the entire dataset.  {\textbf{Recourse Cost}}  \cite{huang2023global} evaluates global explanations considering the distance between the original instances $x \in X$ and the produced counterfactuals $x' \in X'$, i.e., $Cost(X') = f_{x \in X}( \min_{x' \in X'} \mathcal{D}_{inst}(x, x'))$ where $f$ is a permutation-invariant aggregation function. Note that, if $|X| = |X'| = 1$, the recourse cost can be rewritten as the previous minimality metrics by appropriately defining $\mathcal{D}_{inst}(x,x')$. {\textbf{Coverage}} \cite{huang2023global} measures the quality of the counterfactual explanations as the proportion of input instances $x \in X$ that have close counterfactuals in $X'$ under a given distance threshold $t$, i.e., $Cov(X,X') = \frac{1}{X} \cdot |\{x \in X\;|\;\min_{x' \in X'} \mathcal{D}_{inst}(x, x') \leq \theta \}|$. \textbf{Compactness}, like \textit{Interpretability} \cite{huang2023global}, accounts for the number of counterfactual instances included in the explanation set $X'$. The larger the number of counterfactual graphs in the solution, the harder it gets for a human to understand the explanation. Compactness returns a value in the interval $(0,1]$ where values closer to 1 are preferred. We define the metric as $Comp(X') = \frac{1}{|X'|}$.

\subsection{Adopted evaluation protocols for GCE}\label{sec:eval:startegies}

Here, we report the evaluation protocol adopted in the surveyed works. To the best of our knowledge, this is the first survey that provides the reader with a detailed description of the way GCE is evaluated in the literature. Knowing how different works evaluate their performance, not only via the metrics adopted but also via the datasets used, compared baselines, and evaluation scenarios, permits future researchers to have a deep understanding of the challenges in this field.

\noindent\textbf{DCE} \cite{faber2020contrastive}, used as a baseline in \cite{abrate2021counterfactual}, refers to explanations that belong to the same data distribution as the input instance. It searches for a counterfactual instance $G^*$ in the dataset such that $G^* = \arg\min_{G' \in \mathcal{G},\;\Phi(G) \neq \Phi(G')} \mathcal{D}_{inst}(G, G')$.

\noindent\textbf{DDBS} and \textbf{OBS} \cite{abrate2021counterfactual} are originally proposed in the domain of -omics networks. Both methods are compared against a simple Data Search (DS) method that looks for a counterfactual instance in the dataset whose distance is minimal w.r.t. the original one. They are tested on the ASD and ADHD datasets. Since these methods are search-based, they exploit a simple white-box classifier that produces a 2-dimensional embedding for the input graph and pass this embedding to a linear classifier to produce the counterfactual. The authors of this work rely on the GED between the counterfactual example and the original instance and the Oracle Calls to evaluate the performance.

\noindent\textbf{RCExplainer} \cite{bajaj2021robust}, as a factual-based explanation method, is compared to GNNExplainer \cite{ying2018graph}, PGExplainer \cite{luo2020parameterized}, PGM-Explainer \cite{vu2020pgm}, and CF-GNNExplainer. The baselines that produce a set of vertices $V'$ as an explanation were modified to induce a subgraph $G' = (V', E')$ on $V'$ whose edges, $E'$, are the counterfactuals produced. Similarly, the baselines that identify a subgraph $G' = (V',E')$ were modified to output $E'$ only. As seen in Table \ref{tab:methods}, the performance is assessed on graph classification and vertex classification, both relying on a GNN oracle. RCExplainer is evaluated on BA-2motifs, Mutagenicity, and NCI1 for graph classification. For vertex classification, it is evaluated on BA-Shapes, BA-Community, Tree-Cycles, and Tree-Grid. Fidelity, Robustness, and Runtime are used for comparing RCExplainer with SoA methods.

\noindent\textbf{GNN-MOExp} \cite{liu2021multi} is only compared to SoA factual methods. The metrics used for evaluation are simulatability, probability of necessity, and robustness. Notice that simulatability measures whether a subgraph explanation preserves the classification of the original instance. Therefore, since counterfactuality aims to change the original class, simulatability is not suitable in GCE. GNN-MOExp is tested in different social graphs for vertex classification, including CiteSeer, CORA, and Musae-Facebook.

\noindent\textbf{MEG} \cite{numeroso2021meg} is not compared to other SoA methods due to its inherent design for molecular graphs. TOX21 and ESOL are two benchmark datasets used to assess MEG's performances. Additionally, MEG incorporates a molecule sanitisation check to filter out invalid instances. Hence, TOX21 remains with $1756$ equally distributed samples, while ESOL has $1129$ compounds. A split of $80\%:10\%:10\%$ is used for training, validation, and test sets for both datasets. During the counterfactual generation procedure, MEG finds $10$ counterfactuals for each input molecule ranked according to its multi-objective scoring function (reward). The counterfactuals are evaluated according to the MEG similarity function.

\noindent\textbf{GNNAdv} \cite{sun2021preserve} provides factual and counterfactual explanation methods. The provided counterfactual explainer is not compared against any other SoA counterfactual method. Furthermore, the GCE method is only tested on MS–COCO \cite{lin2014microsoft}, a dataset for image classification, considering the co-occurrence of object labels within the images as graphs. Accuracy is the only metric used to evaluate the counterfactual explainer because the authors consider counterfactual explanations as adding noise to the original instance and thus expect the accuracy of the oracle to decrease when counterfactuals are provided instead of the original instances.

\noindent\textbf{CMGE} \cite{wu2021counterfactual} is based on particular knowledge graphs based on Electronic Medical Records (EMRs) of hospitalised patients. First, the authors extract three main features from Electronic Medical Records (EMRs) and submit them for human evaluation. The employed dataset, \textit{not publicly available}, is composed of Chinese EMRs of lymphedema patients. The authors also use MIMIC-III-50 \cite{mullenbach2018explainable} to assign multiple International Classification of Diseases (ICD) codes to EMRs. The proposed method is tested for link prediction and its explanation capabilities are not directly compared against other counterfactual methods.

\noindent\textbf{NSEG} \cite{cai2022probability} is compared against other factual explanation methods and CF$^\text{2}$ \cite{tan2022learning}. The metrics used in the evaluation are versions of the fidelity metric, namely \textit{Fidelity+} and \textit{Fidelity-} to quantify necessity and sufficiency, respectively, and \textit{charact score}, which is a combination of both. Furthermore, the authors use top-k accuracy and  AUCROC, which are commonly used to evaluate factual explainers. NSEG's performance is evaluated on BA-Shapes, Tree-Cycles, Tree-Grid \cite{ying2019gnnexplainer}, Mutagenicity \cite{kazius2005derivation} and MSRC-21 \cite{morris2020tudataset}.

\noindent\textbf{CF-GNNExplainer} \cite{lucic2022cf} can only remove edges from the original instance. The authors compare their performance with Random, Only-1hop, Rm-1hop, and GNNExplainer \cite{ying2019gnnexplainer}. Random is a method used for sanity checking and randomly removes edges from the original graph. Only-1hop and Rm-1hop are based on the 1-hop neighbourhood of the vertex - its ego-graph. Only-1hop keeps all the edges in the ego-graph, while Rm-1hop removes them. The employed datasets are Tree-Cycles, Tree-Grids, and BA-Shapes for vertex classification. The metrics adopted to assess the performance of the explanations are Fidelity, Sparsity, and the explanation size. The experimental results show that CF-GNNExplainer can generate counterfactuals for the majority of the vertices in the test set by removing only a small number of edges.

\noindent\textbf{CLEAR} \cite{ma2022clear} is compared to a Random method that performs a fixed number of random perturbations on the input graph and two of its variants, i.e., one that removes edges and one that adds edges. CLEAR also compares against SoA methods, among whom CF-GNNExplainer, MEG, and GNNExplainer. Here, the authors modify GNNExplainer to remove the identified subgraph (factual instance) from the original graph to produce a possible counterfactual. CLEAR is tested on Community \cite{ma2022clear}, an ad-hoc synthetic dataset, IMDB-M, and Ogbg-molhiv, relying on metrics such as runtime, correctness, proximity, and causality. For the causality aspect reported in the original paper, the authors measure the ratio of counterfactuals that satisfy the causal constraints corresponding to a predefined relation of interest.

\noindent\textbf{MACDA} \cite{nguyen2022explaining} is compared to Joint-List and MaMEG. Joint-List chooses the top-10 drug and protein counterfactual instances that have the highest difference in predicted affinity and similarity w.r.t. the original instance. MaMEG is an adaptation of MEG to the drug-target counterfactual generation task. Ad-hoc metrics are used to measure MACDA's minimality of the generated counterfactual, i.e., average drug encoding similarity and average protein encoding similarity. Additionally, MACDA relies on the Prediction Distance. The method uses the Davis and PDBBind datasets.

\noindent\textbf{CF$^\text{2}$} \cite{tan2022learning} is a factual-based method compared to factual explanation methods, i.e., GNNExplainer, CF-GNNExplainer and GEM \cite{lin2021generative}. The datasets used are BA-Shapes, Tree-Cycles, Mutagenicity, NCI1, and CiteSeer. BA-Shapes, Tree-Cycles and CiteSeer are employed for vertex classification, while Mutagenicity and NCI1 for graph classification. Furthermore, BA-Shapes and Tree-Cycles have ground-truth motifs for explaining the classification since they are human-designed, while NCI1 and CiteSeer do not contain such motifs. The used metrics are Accuracy and the Probability of Necessity\footnote{The original paper also presents precision, recall, and F1 scores for the evaluation. However, we report only counterfactual-related evaluation protocols.}.

\noindent\textbf{MACCS} \cite{wellawatte2022model} is tied to the molecular domain, and, as such, it is not compared to other SoA methods. It is tested on three real-world datasets: i.e., BBBP, ESOL, and HIV. It relies on a Random Forest as oracle to produce counterfactuals on BBBP, a GRU on ESOL, and a GCN on HIV. The Tanimoto similarity is used to assess the performance on all datasets.

\noindent\textbf{GREASE} \cite{chen2022grease} is designed to explain user-item recommendations, and it is not compared against other SoA explanation methods. It is tested on two real-world datasets, namely LastFM and Yelp. The evaluation metrics used are Probability of Necessity and Explanation Cost.

\noindent\textbf{GCFExplainer} \cite{huang2023global} is compared against CF$^\text{2}$ and RCExplainer. It is also compared against a bespoke omniscient method that produces counterfactuals directly taken from the dataset for a particular desired class to explain. The datasets used are NCI1, Mutagenicity, AIDS, and Proteins \cite{borgwardt2005protein,dobson2003distinguishing}. Since this work produces global counterfactual explanations, the metrics used for evaluation are coverage, recourse cost, and compactness. Additionally, standard metrics such as runtime are also considered. According to the global scenario this work adopts, GCFExplainer demonstrates that producing model-level counterfactuals generalises better than producing multiple counterfactuals at the instance level.

\section{Empirical evaluation of GCE methods}\label{sec:gretel_exp}

A comprehensive benchmark of existing GCE methods is out of the scope of this work. However, here we use GRETEL \cite{prado2022gretel,prado2023developing} to assess the performance of several SoA methods in multiple domains\footnote{Notice that the repository is constantly updated and the number of explainers is periodically increased. Current reporting is done on the version of April 2023.}. We use one dataset for each scenario: i.e., Tree-Cycles (synthetic), ASD (-omics), and BBBP (molecular). Table \ref{tab:gr_datasets} reports the datasets' general statistics. Meanwhile, Table \ref{tab:soa_performance} depicts the performance of the methods on the test set (i.e., 10\% of $|\mathcal{G}|$). For each method, we report averages on 10-fold cross-validation. We use runtime, GED, the number of oracle calls, correctness, sparsity, fidelity, and the oracle's accuracy as metrics. Notice that all methods share the same folds\footnote{The same number of folds and the same exact splits of the data.}. In this way, we guarantee a fair comparison of all methods over the same view of the data instances. Moreover, methods that are trained (i.e., CLEAR, CF$^\text{2}$, MEG) or oracle-oblivious (i.e., RAND@k) do not need to access the oracle at inference (test) time. We were unable to adapt MACCS to Tree-Cycles and ASD as its optimisation function incorporates considering the produced counterfactuals as valid molecules, hence the dashes in the table. In Sec. D.1, we provide reasons to exclude from this evaluation some of the methods surveyed in Sec. \ref{sec:methods}. In Sec. D.2, we describe the explainer hyperparameters. All configuration files needed to reproduce the conducted empirical evaluation are available on GRETEL's GitHub repository\footnote{\url{https://github.com/MarioTheOne/GRETEL}}.

\begin{table}[!t]
\caption{The dataset characteristics. $|\mathcal{G}|$ is the number of instances; $\mu(|V|)$ and $\sigma(|V|)$ represent the mean and std of the number of vertices per instance; $\mu(|E|)$ and $\sigma(|E|)$ represent the mean and std of the number of edges per instance; $|C_i|$ is the number of instances in class $i \in \{0,1\}$. $|\text{Test set}|$ represents the number of instances evaluated in each fold.}
\resizebox{0.8\textwidth}{!}{\begin{tabular}{lrrrrrrrrr}
    \toprule
     &  $|\mathcal{G}|$ &  $\mu(|V|)$ & $\sigma(|V|)$ & $\mu(|E|)$ & $\sigma(|E|)$ &  $|C_0|$ &  $|C_1|$ & Class distr. & $|\text{Test set}|$\\
    \midrule
    Tree-Cycles & 500 & 32 & 0 & 31.54 &  0.62 & 263 & 237 & 0.526 : 0.474 & 50\\
    ASD &  101 &  116 &   0 &   655.62 &   7.29 & 52 & 49 & 0.515 : 0.485 & 10\\
    
    BBBP & 2039 & 24.06 & 10.58 & 25.95 & 11.71 & 479 & 1560 &   0.235 : 0.765 & 203\\
    \bottomrule
        \end{tabular}}
 
\label{tab:gr_datasets}
\end{table}

\begingroup
\setlength{\tabcolsep}{3pt} 
\begin{table}[ht]
\centering
\caption{Evaluation of the SoA for 10-fold cross validation. All oracles have been pre-trained. $\times$ depicts no convergence in two days; while $-$ means the method cannot be adapted to the domain at-hand.}
\label{tab:soa_performance}
\resizebox{.95\textwidth}{!}{%
\begin{tabular}{@{}clrrrrrrr@{}}
\toprule
Dataset & Method & Runtime $\downarrow$ & GED $\downarrow$ & Oracle Calls $\downarrow$ & Correctness $\uparrow$ & Sparsity $\downarrow$ & Fidelity $\uparrow$ & Oracle Accuracy $\uparrow$ \\ \midrule
\multirow{10}{*}{\rotatebox{90}{Tree-Cycles}} 
 & RAND@5 & $\mathbf{0.01 \pm 0.003}$ & $92.18 \pm 5.44$ & $0.00 \pm 0.00$ & $0.55 \pm 0.50$ & $1.45 \pm 0.09$ & $0.55 \pm 0.50$ & $1.00 \pm 0.00$ \\
 & RAND@10 & $0.02 \pm 0.006$ & $123.74 \pm 7.43$  & $0.00 \pm 0.00$ & $0.51 \pm 0.50$  & $1.94 \pm 0.12$  & $0.51 \pm 0.50$ & $1.00 \pm 0.00$ \\
 & RAND@15 & $0.01 \pm 0.004$ & $147.93 \pm 8.26$  & $0.00 \pm 0.00$ & $0.58 \pm 0.50$  & $2.33 \pm 0.13$  & $0.58 \pm 0.50$& $1.00 \pm 0.00$ \\
 & DCE & $0.13 \pm0.00$ & $50.36 \pm 0.00$ & $501.00 \pm 0.00$ & $\mathbf{1.00 \pm 0.00}$ & $0.79 \pm 0.00$ & $\mathbf{1.00 \pm 0.00}$ & $1.00 \pm 0.00$ \\
 & OBS & $0.07 \pm 0.01$ & $57.31 \pm0.03$ & $149.45 \pm21.13$ & $0.96 \pm0.01$ & $0.90\pm0.00$ & $0.96\pm0.01$ & $1.00 \pm 0.00$ \\
 & DDBS & $8.87 \pm0.10$ & $71.79 \pm0.24$ & $1342.62 \pm11.95$ & $0.59 \pm0.01$ & $1.13\pm0.00$ & $0.59\pm0.01$ & $1.00 \pm 0.00$ \\
 & MACCS & $-$ & $-$ & $-$ & $-$ & $-$ & $-$ & $-$ \\
 & CLEAR  & $2.47 \pm 0.08$ & $79.76 \pm 3.60$ & $0.00 \pm 0.00$ & $0.53 \pm 0.10$ & $1.26 \pm 0.06$ & $0.53 \pm 0.10$ & $1.00 \pm 0.00$ \\
 & CF$^\text{2}$ & $0.41 \pm 0.01$ & $\mathbf{31.54 \pm 0.12}$ & $0.00 \pm 0.00$ & $0.47 \pm 0.10$ & $\mathbf{0.50 \pm 0.00}$ & $0.47 \pm 0.10$ & $1.00 \pm 0.00$ \\
& MEG& $272.11 \pm 5.66$ & $159.70 \pm 1.34$ & $0.00 \pm 0.00$ & $0.53 \pm 0.00$ & $2.51 \pm 0.02$ & $0.53 \pm 0.00$ & $1.00 \pm 0.00$ \\
 \midrule
\multirow{10}{*}{\rotatebox{90}{ASD}}
 & RAND@5 & $1.45 \pm 0.46$ & $618.06 \pm 8.27$ & $0.00 \pm 0.00$ & $0.00 \pm 0.00$ & $0.80 \pm 0.01$ & $0.00 \pm 0.00$ & $0.79 \pm 0.08$ \\
& RAND@10 & $2.76 \pm 1.25$ & $1152.93 \pm 20.19$ & $0.00 \pm 0.00$ & $0.00 \pm 0.00$ & $1.49 \pm 0.02$ & $0.00 \pm 0.00$ & $0.79 \pm 0.08$ \\
& RAND@15 & $1.33 \pm 0.39$ & $1600.78 \pm 18.22$ & $0.00 \pm 0.00$ & $0.00 \pm 0.00$ & $2.08 \pm 0.03$ & $0.00 \pm 0.00$ & $0.79 \pm 0.08$ \\
& DCE & $\mathbf{0.09 \pm 0.02}$ & $1011.69 \pm 0.00$ & $102.00 \pm 0.00$ & $\mathbf{1.00 \pm 0.00}$ & $1.31 \pm 0.00$ & $\mathbf{0.54 \pm 0.00}$ & $0.79 \pm 0.08$ \\
 & OBS & $3.24 \pm 1.13$ & $\mathbf{9.89 \pm 0.11}$ & $347.73 \pm 15.11$ & $\mathbf{1.00 \pm 0.00}$ & $\mathbf{0.01 \pm 0.00}$ & $\mathbf{0.54 \pm 0.00}$ & $0.79 \pm 0.08$ \\
 & DDBS & $83.46 \pm 34.04$ & $11.79 \pm 0.29$ & $362.05 \pm 14.56$ & $\mathbf{1.00 \pm 0.00}$ & $0.02 \pm 0.00$ & $\mathbf{0.54 \pm 0.00}$ & $0.79 \pm 0.08$ \\
 & MACCS & $-$ & $-$ & $-$ & $-$ & $-$ & $-$ & $-$ \\
 & CLEAR  & $0.45 \pm 0.04$ & $1739.60 \pm 131.16$ & $0.00 \pm 0.00$ & $0.47 \pm 0.13$ & $2.25 \pm 0.17$ & $0.25 \pm 0.18$ & $0.79 \pm 0.08$ \\
 & CF$^\text{2}$ & $0.69 \pm 0.01$ & $655.49 \pm 2.87$ & $0.00 \pm 0.00$ & $0.46 \pm 0.09$ & $0.85 \pm 0.00$ & $0.37 \pm 0.15$ & $0.79 \pm 0.08$ \\ 
& MEG & $\times$ & $\times$ & $\times$ & $\times$ & $\times$ & $\times$ & $\times$ \\
   \midrule
\multirow{12}{*}{\rotatebox{90}{BBBP}}
 & RAND@5 & $\mathbf{0.01 \pm 0.03}$ & $30.98 \pm 33.27$ & $0.00 \pm 0.00$ & $0.85 \pm 0.35$ & $0.52 \pm 0.23$ & $0.62 \pm 0.69$ & $0.86 \pm 0.02$ \\
& RAND@10 & $\mathbf{0.01 \pm 0.03}$ & $52.98 \pm 58.96$ & $0.00 \pm 0.00$ & $0.86 \pm 0.35$ & $0.93 \pm 0.41$ & $0.65 \pm 0.66$ & $0.86 \pm 0.02$ \\
& RAND@15 & $0.02 \pm 0.12$ & $82.97 \pm 137.37$ & $0.00 \pm 0.00$ & $0.85 \pm 0.36$ & $1.32 \pm 0.70$ & $0.61 \pm 0.69$ & $0.86 \pm 0.02$ \\
& DCE & $37.51 \pm 5.21$ & $27.92 \pm 0.12$ & $2040.00 \pm 0.00$ & $\mathbf{1.00 \pm 0.00}$ & $0.59 \pm 0.00$ & $\mathbf{0.72 \pm 0.00}$ & $0.86 \pm 0.02$ \\
 & OBS & $2.92 \pm 0.07$ & $0.00 \pm 0.00$ & $314.61 \pm 0.00$ & $0.00 \pm 0.00$ & $\mathbf{0.00 \pm 0.00}$ & $0.61 \pm 0.00$ & $0.86 \pm 0.02$ \\
 & DDBS & $\times$ & $\times$ & $\times$ & $\times$ & $\times$ & $\times$ & $\times$ \\
 & MACCS & $31.35 \pm 0.97$ & $\mathbf{11.23 \pm 0.08}$ & $1221.33 \pm 0.22$ & $0.40 \pm 0.00$ & $0.19 \pm 0.00$ & $0.23 \pm 0.00$ & $0.86 \pm 0.02$ \\
 & CLEAR@1 & $213.21 \pm 5.64$ & $27056.29 \pm 9.69$ & $0.00 \pm 0.00$ & $0.87 \pm 0.02$ & $91.89 \pm 0.18$ & $0.64 \pm 0.03$ & $0.86 \pm 0.02$ \\
 & CLEAR@5 & $214.80 \pm 6.97$ & $26711.57 \pm 112.67$ & $0.00 \pm 0.00$ & $0.85 \pm 0.02$ & $90.71 \pm 0.37$ & $0.62 \pm 0.03$ & $0.86 \pm 0.02$ \\
 & CLEAR@15 & $251.93 \pm 36.01$ & $25986.66 \pm 170.41$ & $0.00 \pm 0.00$ & $0.85 \pm 0.01$ & $88.20 \pm 0.66$ & $0.62 \pm 0.06$ & $0.86 \pm 0.02$ \\
 & CF$^\text{2}$ & $84.41 \pm 49.06$ & $25.72 \pm 0.63$ & $0.00 \pm 0.00$ & $0.85 \pm 0.02$ & $0.09 \pm 0.00$ & $0.63 \pm 0.03$ & $0.86 \pm 0.02$ \\
& MEG & $90.66 \pm 29.51$ & $269.35 \pm 0.39$ & $0.00 \pm 0.00$ & $0.51 \pm 0.04$ & $0.91 \pm 0.00$ & $0.32 \pm 0.04$ & $0.86 \pm 0.02$ \\
   \bottomrule
\end{tabular}%
}
\end{table}
\endgroup

Oracles have been pre-trained on the entire dataset, as happens in production. Hence, because all explainers are tested on the same folds, thus the oracle accuracy is the same across the board. For Tree-Cycles we rely on an omniscient oracle that exploits a graph visit that verifies whether an already-visited vertex can be revisited, thus presenting a cycle. This oracle never fails to identify (a)cyclic graphs, thus guaranteeing a perfect accuracy of $1.00$. 
For ASD, we rely on a white-box classifier used in \cite{abrate2021counterfactual}.
This oracle is a rule-based classifier that looks at the co-activation of specific regions of interest in a brain graph.
 
For BBBP, we rely on a GCN with four graph convolutional layers interleaved with ReLU activation functions. The convolution is then aggregated via average pooling over the node features. This aggregation is finally passed to two dense layers with $[256, 1]$ neurons and a final sigmoid.

We also compare the SoA with a bespoke random explainer: RAND@k builds a counterfactual graph $G' = (V,E')$ of $G = (V,E)$ by randomly choosing k\% of the edges in $\binom{V}{2}$. First, RAND@k copies $E$ to $E'$. Then it skims through the sampled k\% edges. To this end, if a sampled edge $(v_i,v_j) \in E$, then $E' = E' - \{(v_i,v_j)\}$; contrarily, if $(v_i,v_j) \notin E$, then $E' = E' \cup \{(v_i,v_j)\}$. We use k $\in \{0.05, 0.10, 0.15\}$.

In Tree-Cycles, DCE has the highest correctness across the board. However, since DCE is search-based, it suffers from, potentially, a higher GED w.r.t. the other compared methods. As expected, RAND@k has correctness that is just above the chance level, which, itself, is a bar that most of the SoA does not reach (compare RAND@15 with CF$^\text{2}$, CLEAR, MEG, and DDBS). In detail, CF$^\text{2}$ has the lowest GED overall, but it fails to produce valid counterfactuals $\sim53$\% of the time. Contrarily, CLEAR, and DDBS, although reporting a higher correctness than CF$^\text{2}$, produce counterfactuals that are nearly twice as distant to the input graph. This phenomenon can also be noticed in sparsity, which is a scaled GED (see Sec. \ref{sec:evaluation_metrics}).  It is interesting to note that the correctness and fidelity are the same for all methods in Tree-Cycles. Recall that fidelity measures how faithful the counterfactual explanations are to the oracle considering their correctness. Because the oracle employed in this scenario is guaranteed to predict the correct class of the input, the fidelity is always faithful to the oracle's prediction. Thus, the $\chi(G)$ component of fidelity is always going to output 1, while $\mathbb{1}[\Phi(G') = y_G]$ depends on whether the explainer returns a valid counterfactual or not. In this scenario, since we know that the oracle is always right, we can attribute the misclassifications to the inability of the explainer to produce valid counterfactuals. We adapted MEG's action-choosing policy (agent environment) in this scenario by flipping the adjacency matrix of the input graph $G$ (i.e., an existing edge is removed; a non-existing edge is added). Therefore, for each cell $v_i,v_j$ in $G$'s adjacency matrix $A$, MEG can take an action to produce a counterfactual $G' = (V,E')$ s.t. $A'[v_i,v_j] = 1 - A[v_i,v_j]$. Hence, for each input $G$, there are $\binom{|V|}{2}$ possible actions that lead to a potential valid counterfactual. Now, notice that MEG has the highest runtime across the board. This happens because the number of possible actions per instance are $\binom{32}{2} = 496$. Additionally, because MEG's graph perturbation policy mentioned before is constrained to a single edge addition/removal to produce a counterfactual, its GED is the largest across the board. Notice also that MEG's oracle calls depend on the way the environment is implemented. In this scenario, it does not require any information from the oracle to reward certain actions more than others (see the discussion on BBBP). Finally, CLEAR and CF$^\text{2}$ access each instance once per epoch during training. Hence, CLEAR has $450 \times  600$ oracle cals, while CF$^\text{2}$ has only $450 \times 100$.

In ASD, it is interesting to notice that RAND@k performs poorly. Without loss of generality, we expect RAND@k to perform poorly in scenarios where the decision boundary between classes is defined at the vertex/edge feature space instead of being expressed in terms of connectivity patterns. Because RAND@k operates only on the adjacency matrix of the input graphs, its correctness, and fidelity suggest that it is unable to produce any valid counterfactual. Besides, its GED is among the worst across the board. All SoA methods surpass the random explainer, which leads us to believe that feature-based SoA methods need to be further investigated to tackle hard cases such as ASD. However, CLEAR and CF$^\text{2}$ do not have satisfactory correctness ($\sim 0.47$ and $\sim 0.46$, respectively) as it does not exceed the chance level of producing a valid counterfactual. Additionally, CLEAR has the highest GED across the board. We believe this is due to the graph-matching procedure that CLEAR has after it samples a counterfactual $G'$ from its latent space. This makes the sampled graph $G'$ lose the order of the vertices w.r.t. the original instance. It is interesting to notice for DCE, OBS, and DDBS that their correctness and fidelity are the same. Furthermore, since we know that these methods always produce a valid counterfactual - i.e., correctness is equal to 1 - we can state that oracle is not capable of differentiating between factual and counterfactual instances (see oracle's accuracy). Note that we adapt MEG to this scenario as described in Tree-Cycles. Therefore, due to the elevated number of possible actions per instance $\binom{116}{2} = 6670$, MEG is not able to generate a counterfactual after several days of execution. OBS and DDBS remain the best-performing methods in terms of correctness and GED. CF$^\text{2}$ has $91\times 100$ oracle accesses during training, and CLEAR $91 \times 600$.

In BBBP, notice that RAND@k has good performances in terms of correctness. This means that the employed oracle is sensitive to the connectivity patterns of the molecules rather than the feature set of each vertex (e.g., atom, valence, ionisation) in the molecule graph. Contrarily to Tree-Cycles, perturbing more than $5$\% of the edges does not have any benefits in this scenario. Notice that DCE has the highest correctness and a low GED, which suggests that the instances in the test set are similar to one another due to DCE's inherent counterfactual searching mechanism. Although MACCS and MEG are designed to work with molecules, they perform poorly, as expected, in terms of correctness with $\sim 0.4$ and $\sim x$, respectively, highlighting the necessity to have not only a data type-oriented explainers but also a domain-specific one. Meanwhile, their GED is better than DCE, which suggests that the counterfactual that DCE searches in the test set is not the least distant from the input molecule (DCE's GED is $27.92$, while MACCS's is $11.23$). CF$^\text{2}$ has one of the lowest GED, second to MACCS, even if it is not designed to work with molecular graphs specifically. Surprisingly, CF$^\text{2}$ reports correctness of $0.85$ (random) that exceeds domain-specific methods (i.e., MACCS and MEG). Notice that OBS fails to produce valid counterfactuals (correctness and sparsity equal to $0$). Because its GED is $0$, we can conclude that it returns the original instance. DDBS fails to produce a valid explanation for a single instance within $4$ hours\footnote{We ran experiments on an AMD Ryzen 7 4800HS, 2.90 GHz, 32 GB RAM.} of searching through the test set. CLEAR has $1836 \times 5$ oracle calls during training, while CF$^\text{2}$ has $1836 \times 100$. Finally, notice that CLEAR cannot compete with the SoA in BBBP although its correctness score is the second highest due to an extremely elevated GED and sparsity. Due to the size of the dataset, we only use $5$ epochs to train CLEAR, which might have affected its overall performance.

In conclusion, Table \ref{tab:soa_performance} paints a raucous picture of SoA methods. Most GCE methods fail to produce valid counterfactuals most of the time, even with an underlying omniscient and non-biased oracle (see Tree-Cycles). Besides, in the majority of scenarios, a random perturbation explainer outperforms the SoA, which entails the literature's need to perform an exhaustive evaluation with baselines and not only with other SoA explainers (notice BBBP). Generally, according to these empirical results, there is no silver bullet explanation method. The application domain and the particularities of the dataset influence explanation capabilities. Hence, it is important to use multiple explanation methods as a way to increase fairness and trustworthiness.

\section{Privacy of Explainers and Fairness of Oracles in GCE}\label{sec:privacy}
The European Union's vision for AI is to encourage excellence and trustworthiness, boost research and productivity, reinforce safety, and protect citizen rights \cite{european2020artificial}. To this end, the EU proposed a risk-based legal framework to regulate AI and address risks specifically created by AI applications \cite{european2021AIRegulation}. The framework aims to ensure AI systems' safety, privacy, fairness, and trustworthiness to create safer and more innovation-friendly digital environments.

The Commission has placed extensive regulations for AI's trustworthiness and the risk mitigation of its wide usage. The proposal points towards black-box models that are inherently non-explainable. In this context, interpretable models are seen as more compliant with the EU regulation for AI than their black-box counterparts. Despite that, on the one hand, good counterfactual explanations give domain laymen insight into the prediction of a specific input. On the other hand, GCE methods can pose a significant privacy risk, even when the explainer $\Xi$ does not have direct access to the original data. When $\Xi$ generates a counterfactual $G'$ for the input $G$ it can breach the privacy of sensitive information. For example, in a social network, $\Xi$ may include vertices that belong to sensitive groups, such as juveniles, without considering established privacy policies within the network.

\begin{figure}[!ht]
\centering
\subfloat[Workflow of generating counterfactual examples $G'$ of a graph instance $G$ with an embedded privacy-preservation strategy. Notice that step 0 can be done a priori to fetch specific privacy preservation rules indicated in the original data.]{
	\label{subfig:embedded_privacy}
	\includegraphics[width=.9\textwidth]{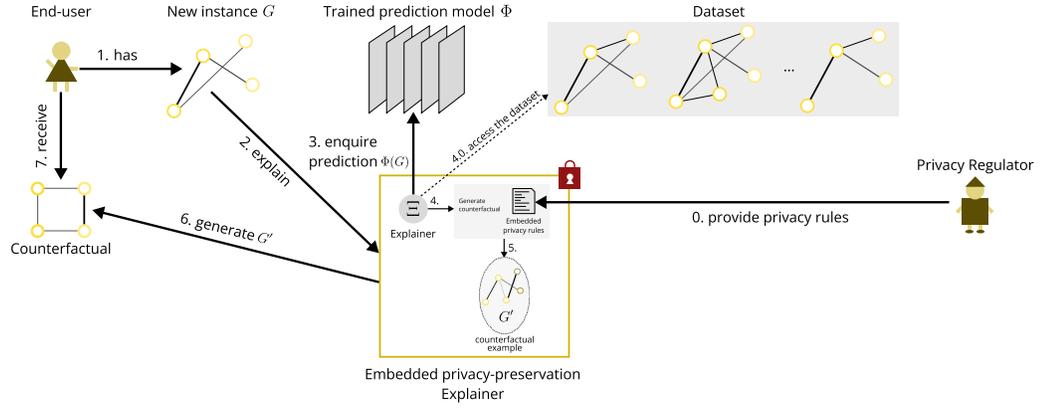}} 

\subfloat[Workflow of generating counterfactual examples $G'$ of a graph instance $G$ with a post-hoc privacy-preservation strategy. The explainer delegates the privacy compliance assessment to an external module.]{
	\label{subfig:post_hoc_privacy}
	  \includegraphics[width=.9\textwidth]{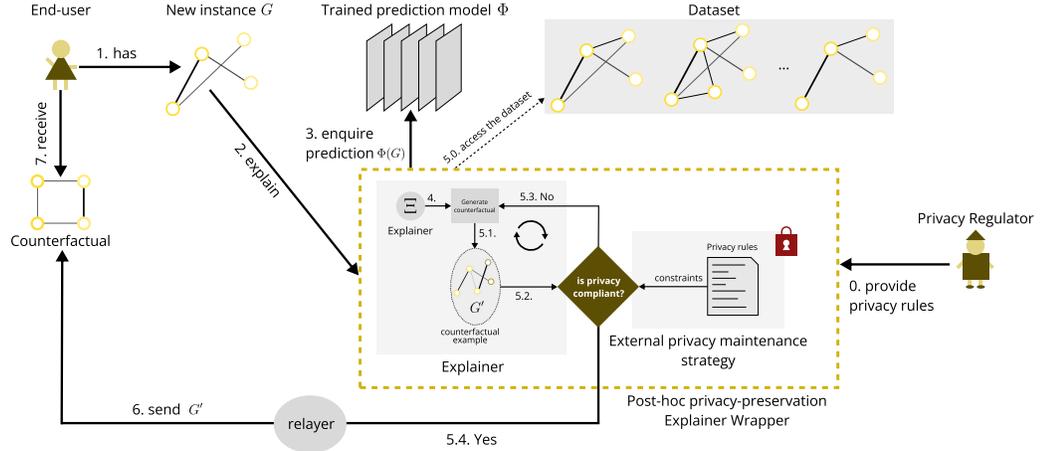} } 
\caption{Workflow of privacy-preserving counterfactual explainers.}
\label{fig:privacy_preserving_explainers}
\end{figure}

\subsection{Privacy of explainers}
The works surveyed here and the aspects they treat to provide counterfactuals to the end-user do not acknowledge any violation of privacy-related policies. For this reason, we propose two high-level solutions to tackle this aspect: i.e., embedded and post-hoc privacy preservation explainers. They differ in how the explainer $\Xi$ interacts with the original data and the end-user. Figures \ref{subfig:embedded_privacy} and \ref{subfig:post_hoc_privacy} depict embedded and post-hoc privacy preservation explainers, respectively. Figure \ref{subfig:embedded_privacy} shows an explainer with an embedded mechanism that constrains the way counterfactuals are generated. As shown in step $4$, $\Xi$ must either have full or partial access to the dataset. If partial access is granted, $\Xi$ would only have access to specific vertex/edge attributes while being barred from extracting sensitive information such as user profiles and other personal data. This access level is critical to generating a counterfactual example that complies with the privacy rules defined in step $0$. For example, in a social network where user vertices have attributes like age, gender, friendships, and profile pictures, $\Xi$ must adhere to privacy policies restricting the extraction of information about profile pictures and gender. Contrarily, Figure \ref{subfig:post_hoc_privacy} shows how $\Xi$ delegates the task of assessing privacy compliance to an external module: i.e., software exoskeleton, which mediates actions in the digital world according to the regulation. The privacy compliance module receives the privacy rules - step $0$ - and the explainer directly communicates with it without requiring them to be embedded within. Instead, $\Xi$ generates a counterfactual $G'$ arbitrarily and then uses a binary test to assess if $G'$ complies with the privacy rules. If the test indicates a violation, $\Xi$ repeats this procedure in the next iteration until a compliant $G'$ is generated (steps $5.1\text{-}5.3$). Once generated, $G'$ is presented to the end-user (steps $5.4\text{-}7$).

Note that generating a privacy-compliant counterfactual may require considerable effort in post-hoc privacy-preservation explainer strategies if the external privacy maintenance module is limited to providing a binary compliance test. Therefore, a more complex test that identifies the source of the compliance failure is necessary to improve the explainer's ability to regulate the generation procedure and mitigate privacy violations in the next iteration. In this case, we propose to use a wrapper to entice the interaction of the explainer with the privacy rules without exposing the rules to the outside. All interactions with the outer part go through the wrapper, not the components. Finally, embedded and post-hoc privacy-preservation explainers can access the dataset to generate more plausible explanations. However, if dataset access is mandatory, it should be handled via an encrypted and trusted channel.

\subsection{Fairness of Oracles}
Generally, fairness refers to whether the black-box model accurately represents and predicts incoming instances based on specific features without exhibiting discriminatory behaviour. For example, consider a toy dataset of banned users on a social network, with information about each individual's race. If the prediction model, $\Phi$, outputs a 1 (i.e., ban) only based on the user's race \cite{nowotny2021contribution}, irrespective of their connection with other users who post derogatory content or any previous terms of service violations, it is an instance of biased prediction. In such cases, counterfactual examples produced under a biased black-box predictor can be an auditing mechanism to detect unfair disparities. For example, \Quote{if the person had not been of a certain race, then they would not have been banned from the social network} is an unfair/biased counterfactual example highlighting $\Phi$'s underlying unfairness. Thus, GCE can be a monitoring strategy to detect risky and unfair predictors.
This auditing characteristic is even more crucial when the discriminated group corresponds to the minority group in the dataset, resulting in class imbalance. A predictor may perform well according to specific evaluation metrics, but it may still be unfair. Therefore, counterfactual examples play an essential role in questioning the fairness of the prediction model in any scenario. 

Moreover, while the literature has explored measuring bias and unfairness of the prediction model \cite{barocas2017fairness,mehrabi2021survey,pessach2022review}, measuring them for counterfactual explainers is an aspect that is yet to be covered. Lastly, the fairness of the oracles was widely investigated in the literature \cite{guo2023survey}.
However, the literature on counterfactual explainability has yet to investigate the fairness of the produced counterfactual examples. Nonetheless, the fairness of the prediction method provides insight into whether the explainer might generate biased explanations. Thus, the explainer's bias is an aspect that needs to be considered when evaluating the \textit{effectiveness} of the generated explanation.
\section{Open challenges and Future Works}\label{sec:open_challenges}

Here, we provide the reader with insight into the remaining open challenges yet to be treated in the area of graph counterfactual explainability so that researchers can concentrate on providing contributions therein.

According to the taxonomy presented in Sec. \ref{sec:methods}, there is a clear lack of methods that address model-level and global-level graph counterfactual explanation. Thus, we expect that in the near future, we will see a growing interest in this sub-domain with the proposition of methods based on learning and search.

Moreover, attentive readers have certainly noticed the complete absence of counterfactual methods designed for edge prediction tasks. This might be primarily imputed to the difficulty of the task and the dubious effectiveness that a counterfactual explanation could have in this case. Nonetheless, we expect a future exploration of this research field.

Even though, in this survey, we extend the definition of a counterfactual explanation to the multi-class setting, more work must be conducted to clarify it in the scenario of multi-label prediction. To the best of our knowledge, the research area also lacks any method dedicated to this task.

Similar to \cite{bias2022Manerba}, researchers could investigate how to measure the discrimination (i.e., unfairness/bias) of the explainers. In this way, explainability can be harnessed to exalt the unfairness of the underlying prediction model or to determine whether the explainer itself suffers from bias. Even though we present some ideas on privacy-preserving possibilities, the field must be explored further.
Therefore, new-generation explainers should be capable of complying with the regulation's guidelines, which are being continuously adopted in more countries.

Besides plain graph data, counterfactual explainability can be used to provide explanations on more complex structures such as temporal graphs, and manifolds. For temporal graphs, explanations should contain a temporal component that better provides counterfactual examples for a specific time interval. Contrarily, for manifolds\footnote{Usually manifolds are represented as meshes of triangles in a three-dimensional space.}, counterfactual examples might be considered as a \textit{surface} in the same vector space of the input instance.

We invite the reader to focus on the core of this survey, which emphasises explainability with GNNs for graph counterfactuals. In light of recent regulations on AI trustworthiness\footnote{\url{https://artificialintelligenceact.eu/the-act/}}, it becomes essential to consider a broader perspective on Trustworthy AI. For a more in-depth understanding of fairness, explainability, and bias within the trustworthiness umbrella term, we refer the reader to \cite{guo2023survey}.

Finally, as also highlighted in Sec. \ref{sec:gretel_exp}, this field of research might benefit from a systematic organisation of public competitions (never done until now) regarding the evaluation of counterfactual explainers as it happens with competitions in data mining (e.g., KDDCup15) that encourage a uniform evaluation benchmark with specific and well-formatted datasets.
\section{Conclusion}\label{sec:conclusion}
We presented a thorough survey on the methods and best-practices for graph counterfactual explainability accompanied by rigorous formalisations of a minimal counterfactual explanation, the evaluation protocols, including datasets and measurements, empirical evaluation, and means to construct privacy-compliant explainers. 

In detail, we provided the reader with an organisation of the literature according to a uniform formal notation, thus, simplifying potential comparisons w.r.t to the method's advantages and disadvantages. 
We emphasise that this is the first work to propose a formalisation of GCE under a multi-class prediction problem. Additionally, we provided a definition that encompasses the global minimal - i.e., the least distant counterfactual w.r.t. the original graph among all classes - for a chosen black-box prediction model.

We proposed a classification of the existing methods according to ten dimensions which aid the reader identify those methods that better suit their scenario of explainability. Besides this classification, we summarised the strengths and weaknesses of the surveyed methods. We also delve into shedding light on the benefits of a standardised evaluation protocol where we enlist synthetic and real datasets used in the literature, the adopted evaluation measures and the evaluation strategies of the surveyed methods.

Additionally, we argue that a fully-extensible and reproducible GCE evaluation framework is of paramount importance. Therefore, we illustrate an empirical evaluation, made with GRETEL which is an evaluation framework that concentrates on providing a highly modular architecture that permits the reader to plug-and-play with their ad-hoc explainer models, synthetic dataset generation, and evaluation metrics.

Finally, we discussed privacy and fairness in GCE which are necessary to comply with regulations being stabilised in more countries worldwide. As a concluding remark, we leave the reader with future directions and open challenges to be tackled in the future of this research area.

\begin{acks}
\begin{small}
This work is partially supported by the European Union - NextGenerationEU - National Recovery and Resilience Plan (Piano Nazionale di Ripresa e Resilienza, PNRR) - Project: SoBigData.it - Strengthening the Italian RI for Social Mining and Big Data Analytics - Prot. IR0000013 - Avviso n. 3264 del 28/12/2021, XAI: Science and technology for the eXplanation of AI decision - ERC Advanced Grant 2018 G.A. 834756 and by the HPC \& Big Data Laboratory of DISIM, University of L’Aquila (\url{https://www.disim.univaq.it/}).
\end{small}
\end{acks}

\bibliographystyle{ACM-Reference-Format}
\bibliography{bibliography}

\begin{appendix}
\section{A formal representation of the GCE problem}\label{sec:background}
Here, we provide the reader with the necessary building blocks and background concepts (see Section \ref{sec:background_concepts}) necessary to the field of Graph Counterfactual Explainability. Moreover, we briefly discuss the difference between factual and counterfactual explanations (see Section \ref{sub:factual_counter}).

\subsection{Background concepts}\label{sec:background_concepts}
Here we introduce definitions of the relevant entities that constitute the background on graph counterfactual explainability. We invite the reader to use Table \ref{tab:formal_notation} as a quick reference for the formal notation used throughout this survey.

\begin{table}[!ht]
    \centering
    \caption{Formal notation used in this survey.}
    \begin{tabular}{l|l}
        \toprule
         Notation & Meaning \\ \hline
         $G = (V,E)$ & A graph with the vertex set $V$ and edge set $E$ \\
         $v_i$ & A vertex $v_i \in V$\\
         $e = (v_i,v_j)$ & An edge $e \in E$ with its incident vertices $v_i$ and $v_j$\\
         $a(v_i)$ & The attributes of vertex $v_i$\\
         $\bar{a}(e)$ & The attributes of edge $e$\\
         $\tilde{a}(G)$ & The attributes of graph $G$ - dubbed global context\\
         $\mathcal{G}$ & The dataset comprising of a set of graphs $G_1,\dots,G_{|\mathcal{G}|}$\\
         $adj(v_i)$ & Set of adjacent vertices of $v_i$\\
         $co(v_i)$ & Set of edges with incident vertex $v_i$\\
         $d(v_i,v_j)$ & Distance of the shortest path between $v_i$ and $v_j$\\
         $A$ & The adjacency matrix of graph $G$\\
         $D$ & The degree matrix of graph $G$\\
         $F$ & The attribute matrix of the vertices of graph $G$\\
         $x$ & An instance to be examined\\
         $x'$ & A counterfactual instance\\
         $\mathcal{F}_x$ & The set of features for instance $x$\\
         $X$ & A set of original instances $\{x_1,\dots,x_{|X|}\}$\\
         $X'$ & A set of counterfactual instances $\{x'_1,\dots,x'_{|X'|}\}$\\
         $C$ & A set of classes $\{c_1,\dots,c_{|C|}\}$. In binary classification $|C| = 2$\\
         $\Phi$ & A black-box model/Oracle\\
        $\hat{y} = \Phi(x)$ & The decision $\hat{y}$ of the model on $x$\\
        $f_k(x,\Phi,X)$ & A counterfactual explainer producing at most $k$ examples\\
         \bottomrule
    \end{tabular}
    \label{tab:formal_notation}
\end{table}

\subsubsection{What is a graph?}
A graph $G = (V,E)$ is a data structure comprising of a set of vertices $V = \{v_1,...,v_n\}$ and a set of edges $E=\{(v_i,v_j)\;|\; v_i \land v_j \in V\}$. In other words, a vertex $v \in V$ depicts an object/concept, whereas an edge $e = (v_i,v_j) \in E$ depicts the relationship between its two incident vertices $v_i,v_j \in V$. Graphs can be either directed or undirected. A directed graph has an asymmetrical edge set, whereas an undirected graph depicts symmetrical relationships. In the undirected graph, $(v_i,v_j) = (v_j,v_i)\in E$. The neighbourhood of $v_i$, denoted by set $adj(v_i)$, contains all the vertices that are connected by an edge to $v_i$ formally $adj(v_i) = \{v_j \in V \;|\; (v_i,v_j) \in E \lor (v_j,v_i) \in E\}$. Similarly, let $co(v_i)$ be the set of edges having $v_i$ as an incident vertex, $co(v_i) = \{(v_i,v_j) \in E\} \cup \{ (v_j,v_i) \in E\}$. Notice that vertices and edges can have categorical and numerical attributes. We denote with $a(v)$ the set of the values of the attributes of vertex $v \in V$, and with $\bar{a}(e)$ the set of values of the attributes of edge $e \in E$. Additionally, we depict with
$\tilde{a}(G)$ the attributes of the entire graph $G$, also referred as global context. For convenience purposes, we define the adjacency matrix ${A}$, the degree matrix ${D}$, and the vertex attribute matrix ${F}$ for graph $G$ as follows:
\begin{minipage}{.33\linewidth}
\begin{equation*}
    A[v_i,v_j] = \begin{cases}
    1 & \text{if } v_j \in adj(v_i)\\
    0 & \text{otherwise}
    \end{cases}
\end{equation*}
\end{minipage}%
\begin{minipage}{.33\linewidth}
\begin{equation*}
    {D}[v_i,v_j] = \begin{cases}
        |adj(v_i)| & \text{if } v_i = v_j\\
        0 & \text{otherwise}
    \end{cases}
\end{equation*}
\end{minipage}%
\begin{minipage}{.33\linewidth}
\begin{equation*}
    {F}[v_i,k] = a(v_i)[k] 
\end{equation*}
\end{minipage}

\subsubsection{Graph Neural Networks}
Having formally defined the characteristics of a graph, we have the necessary tools to introduce graph neural networks (GNNs). In general, a neural network (NN) comprises of many interconnected layers of neurons that, upon receiving a firing signal, propagate information to the successive layers. We refer the reader to \cite{fiesler1994neural} for the formalisation of multiple-layered NNs. Similarly, a GNN is an optimisable transformation on all attributes (i.e., on nodes, edges, and global context) of a certain graph $G$, which typically maintains its proximity characteristics: i.e., two nodes $v_i$ and $v_j$ that have similar attributes and neighbourhoods are near each other in the target embedding space.

\begin{figure}[!t]
    \centering
    \includegraphics[width=\textwidth]{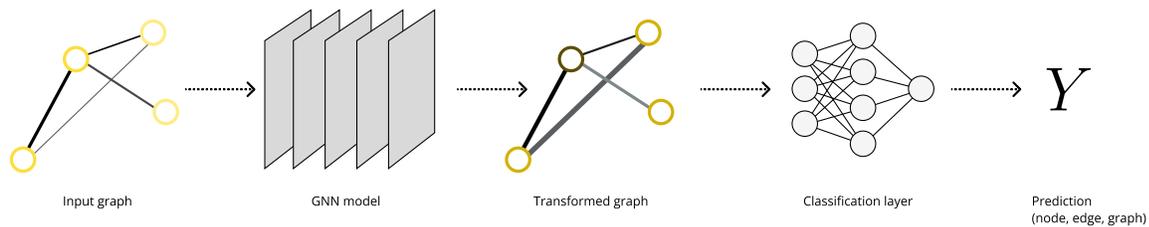}
    \caption{End-to-end prediction task of a GNN model.}
    \label{fig:gnn_general_workflow}
\end{figure}

Figure \ref{fig:gnn_general_workflow} illustrates the end-to-end prediction task of a GNN model. Generally, a GNN accept a graph $G$ in input with attributes associated to its vertices, edges, and a global context associated with $G$. Then, the model produces a transformed graph that gets fed to a classification layer - e.g., a deep neural network - producing a prediction $\hat{y}$. The simplest GNN learns a new embedding for all attributes while not exploiting $G$'s connectivity patterns. This kind of GNN uses a separate multilayer perceptron (MLP) on each component of the graph. For each node, the model applies an MLP and gets back an embedded set of attributes for each vertex. The same reasoning extends to edges and the global context. Recall that this simple GNN does not modify the structure of the graph. However, the transformed graph, shown in Figure \ref{fig:gnn_general_workflow}, contains embedded attributes for each of its components. To make predictions on the transformed graph, we can collect information via pooling and proceed as follows:
\begin{itemize}
    \item For each item to be pooled, we gather its embedding.
    \item We aggregate the gathered embeddings via a permutation-invariant operation (e.g., summation).
\end{itemize}
The pooling operation's functionality depends on the task that we want to perform (i.e., vertex, edge, graph classification).

To exploit the connectivity patterns in a graph $G$, we rely on three different concepts: i.e., random walks\footnote{Random walks can be considered as probabilistic graph visiting algorithms where each edge has a transition probability.}, message passing, and graph convolutional networks (GCNs). For completeness purposes, we provide a description for each of them and detail their similarities and differences.

\noindent
\textbf{Random Walks} are stochastic processes that delineate a path that consists of a succession of random steps on a particular mathematical space \cite{hughes1998random}. Generally, a random walk process in graphs consists of the following three steps:
\begin{enumerate}
    \item Given a graph $G = (V,E)$ and a starting vertex $v_i$, called the seed, select neighbour $v_j \in adj(v_i)$ with probability  $\frac{1}{|adj(v_i)|} = \frac{1}{D[v_i,v_i]}$.
    \item Move to the selected neighbour $v_j$, and repeat the same process as in the previous step until convergence.
    \item The random sequence of vertices visited composes the random walk throughout the graph $G$.
\end{enumerate}
\begin{figure}[!t]
    \centering
    \includegraphics[width=.4\textwidth]{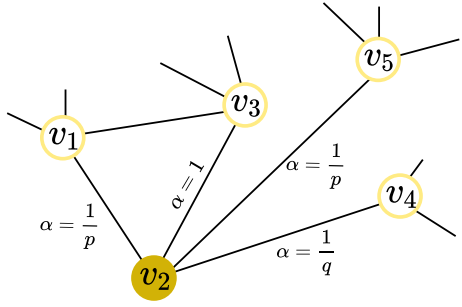}
    \caption{Illustration of a random walk procedure according to Node2Vec \cite{grover2016node2vec}. The walk transitioned from $v_1$ to $v_2$, and it needs to evaluate its next transition towards the vertices in $adj(v_2)$. The labels in the edges indicate the search biases $\alpha$.}
    \label{fig:random_walk}
\end{figure}
One of the most influential works in vertex embeddings based on random walks is Node2Vec \cite{grover2016node2vec}. According to the original proposal, Node2Vec builds on top of a biased random walk process that efficiently exploits neighbourhood exploration in a breadth-first and depth-first search jointly. Given a source vertex $v_i$, the authors simulate a random walk of a specific length $\ell$. The vertices in the walk are visited according to the following probability distribution:
\begin{equation}\label{eq:node2vec}
    P(v_j | v_i) = \begin{cases}
    \frac{\pi_{v_i,v_j}}{Z} &   \text{if } v_j \in adj(v_i)\\
    0 & \text{otherwise}
    \end{cases}
\end{equation}
where $\pi_{v_i,v_j}$ is the non normalised transition probability between vertices $v_i$ and $v_j$, and $Z$ is the normalising constant. To make the stochastic graph visiting described in Equation \ref{eq:node2vec}, Node2Vec adopts a second order random walk with two parameters $p$ and $q$. Figure \ref{fig:random_walk} illustrates a small view of a random walk in Node2Vec. Suppose that the walk just traversed the edge $(v_1,v_2)$ and is now deciding what path to traverse out of $v_2$ according to the probabilities $\pi_{v_2, v_j} \; \forall v_j \in adj(v_2)$. Node2Vec sets the non normalised transition probability from any vertex $v_i$ to $v_j$ as $\pi_{v_i,v_j} = \alpha_{p, q}(v_k,v_j) \cdot w(v_i,v_j)$ such that:
\begin{equation}
    \alpha_{p, q}(v_k, v_j) = \begin{cases}
        \frac{1}{p} & \text{if } d(v_k,v_j) = 0\\
        1 & \text{if } d(v_k,v_j) = 1\\
        \frac{1}{q} & \text{if } d(v_k,v_j) = 2
    \end{cases}
\end{equation}
where $v_k$ is the vertex visited immediately before $v_i$, $d(v_i,v_j)$ is the distance of the shortest path between $v_i$ and $v_j$, and $w(v_i,v_j)$ is the weight of the edge $e = (v_i,v_j)$. Parameters $p$ and $q$ control the velocity of exploring and leaving the neighbourhood of a starting vertex $v_i$. For a detailed discussion about the trade-off between $p$ and $q$, we refer the reader to the original paper \cite{grover2016node2vec}. Besides being able to generate vertex embeddings, Node2Vec can be exploited to learn edge embeddings. Given two vertices $v_i$ and $v_j$, Node2Vec relies on a binary operator over the attributes $a(v_i)$ and $a(v_j)$ to generate an embedded representation $g(v_i,v_j)$ for the edge\footnote{Notice that Node2Vec can also generate embeddings for non-existing edges in the original graph.} $e = (v_i,v_j)$. Other random-walk-based methods are LINE \cite{tang2015line}, DeepWalk \cite{perozzi2014deepwalk}, RW$^2$ \cite{madeddu2020feature}, HeteSpaceyWalk \cite{he2019hetespaceywalk}, and GloVeNoR \cite{kulkarni2020glovenor}.

\begin{figure}[!t]
    \centering
    \includegraphics[scale=.45]{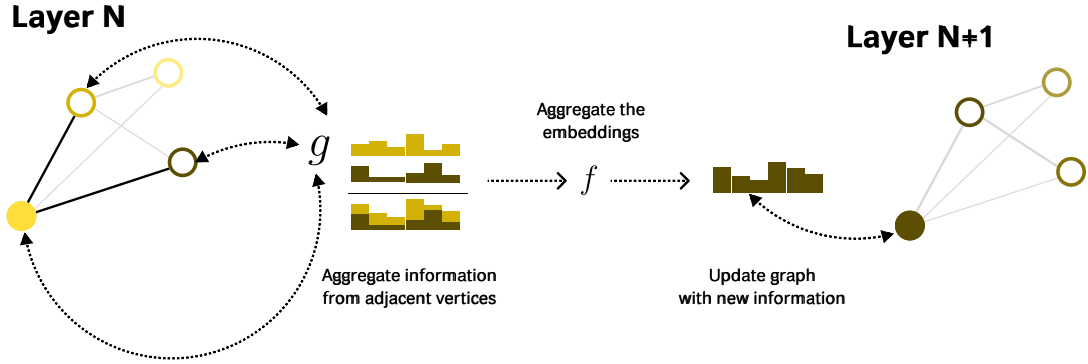}
    \caption{Message passing from the adjacent vertices of the filled vertex and the update of the embeddings in the next layer. For illustration purposes, the gathering function $g$ is a simple stacking operation, and $f$ is a summation.}
    \label{fig:gnn_message_passing}
\end{figure}

\noindent
\textbf{Message passing:} assumes that neighbouring vertices or edges exchange information and influence each other's updated embeddings. Without loss of generality, message passing works in three steps:
\begin{itemize}
    \item For each vertex $v_i \in V$, gather the vertex embeddings of $adj(v_i)$ according to a function $g$.
    \item Aggregate the embeddings (messages) via an aggregation function $f$ (e.g., summation).
    \item All pooled messages are passed through an update function (e.g., a trained NN).
\end{itemize}
Similar to the application of the pooling operation, message passing can occur between either vertices or edges. Figure \ref{fig:gnn_message_passing} depicts a similar operation w.r.t. standard convolution. In other words, message passing and convolutions are operations that aggregate information from the neighbours of a specific element and use this information to update the element's value. By stacking message passing GNN layers, a vertex can incorporate information from multiple hops of its neighbouring vertices: i.e., a two-layered message passing GNN permits a particular vertex $v_i \in V$ to gather embeddings of $adj(v_i)$ and $adj(v_j)\; \forall v_j \in adj(v_i)$. We refer the reader to \cite{gilmer2017neural} for a detailed formalisation of the message-passing framework. Some works that rely on message passing to generate embeddings are DimeNet \cite{klicpera2020directional}, HC-GNN \cite{zhong2020hierarchical}, LaMP \cite{lanchantin2019neural}, and PCAPass \cite{sadowski2022dimensionality}.

\begin{figure} 
    \centering
    \includegraphics[width=.5\textwidth]{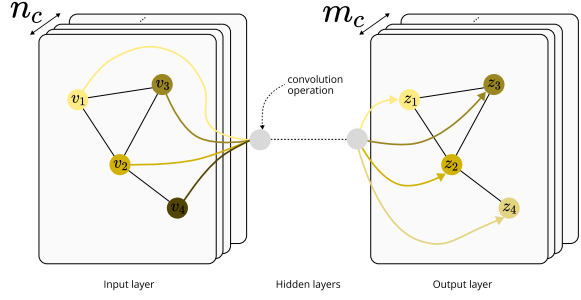}
    \caption{Illustration of a multi-layer GCN with $n_c$ input channels and $m_c$ feature maps in the output layer. The structure of the input graph (shown as the dotted lines) is shared in the hidden layers.}
    \label{fig:gcn}
\end{figure}

\noindent
\textbf{Graph Convolutional Networks (GCNs):} extends the concept of convolution operations from images\footnote{Images can be seen as graphs with a regular grid-like structure, where the individual pixels are vertices, and the RGB channel values at each pixel as the vertex attributes.} to more complex topologies such as graphs.  Recall that $A$ and $D$ are, respectively, the adjacency and the degree matrices of a certain graph $G$. We can define $G$'s non-normalised Laplacian as $L = D - A$. In this way, polynomials of the following form can be expressed:
$$
p_w(L) = w_0I_{|V|\times|V|} + w_1 L + w_2L^2 + \dots + w_dL^d = \sum_{i=0}^d w_iL^i
$$
which can be thought as filters of a vanilla CNN with weights $w = [w_0,\dots,w_d]$, where $I_{|V|\times|V|}$ is the identity matrix of dimensions $|V|\times|V|$. Having the vertex attribute matrix $F$, the convolution of $F$ w.r.t. the polynomial above can be defined as $F' = p_w(L) \cdot F$. Therefore, convolving $F$ with $p_w(L)$ of degree $d$ means that, for each vertex $v_i$, its embedding gets influenced by at most $d$ hops away. Thus, these polynomial filters are localised with $d$ as the degree of the localisation.
Defferrard et al. \cite{defferrard2016convolutional} extend the idea - hereafter ChebNet - of the Laplacian polynomial filters to consider degree-i Chebyshev polynomials of the first kind. Now, the d-degree Laplacian polynomial becomes $p_w(L) = \sum_{i=1}^d w_i T_i (\tilde{L})$ where $\tilde{L}$ is the normalised Laplacian according to the largest eigenvalue of L. By stacking ChebNet of layers with non-linear activation functions, one can perform a convolution over the input graph. In particular, if there are $K$ different polynomial filters - denoting the network layers - the k-th layer has its own learning weights $w^{(k)}$. Hence, starting with $h^{(0)}_v = F[v]$ one can iterate for $k=1,\dots,K$ and compute $h^{(k)}_v = \sigma(p_{w^{(k)}}(L) \cdot h^{(k-1)}_v)$ where $\sigma(\cdot)$ is a non linear activation function, and $h^{(k)}_v$ is the hidden representation of vertex $v$ in layer $k = 1,\dots,K$. Note that ChebNets reuse the same filter weights across different vertices, mimicking weight-sharing in CNNs. Let us focus on the convolutional operation of a specific vertex $v_i$ via the polynomial kernel $p_w(L) = L$:
\begin{equation*}
    \begin{gathered}
        (L\cdot F)[v_i] = L[v_i] \cdot F[v_i]\\
        = \sum_{v_j \in V} L[v_i,v_j] \times F[v_i]\\
        = \sum_{v_j \in V} (D[v_i,v_j] - A[v_i,v_j]) \times F[v_i]\\
        = D[v_i] \times F[v_i] - \sum_{v_j \in adj(v_i)} F[v_j] 
    \end{gathered}
\end{equation*} 
This is a 1-hop localised convolution. But more importantly, we can think of this convolution as arising of two steps: (1) aggregating over immediate neighbour attributes $F[v_j] \;\forall v_j \in adj(v_i)$, and (2) combining $v_i$'s attributes $F[v_i]$. These convolutions can be thought of as "message-passing" between adjacent nodes: after each step, every node receives some "information’ from its neighbours. By iteratively repeating the 1-hop localised convolutions K times - i.e., repeatedly "passing messages" - the receptive field of the convolution effectively includes all nodes up to K hops away. Figure \ref{fig:gcn} illustrates the general idea behind a GCN. Notice that the number of channels/filters can be different in the layers of the network, similarly as in a CNN. A GCN's layer-wise forward propagation works as follows:
\begin{equation*}
\begin{gathered}
 h_{v_i}^{(0)} = F_{v_i} \;\forall v_i \in V\\
 h_{v_i}^{(k)} = f^{(k)} \left(W^{(k)} \cdot \frac{\sum_{v_j \in adj(v_i)} h_{v_j}^{(k-1)}}{|adj(v_i)|} + B^{(k)}\cdot h_{v_i}^{(k-1)}\right)\;\forall k \in [1,K]
\end{gathered}
\end{equation*}
In other words, the embedding of vertex $v_i$ at layer $k$ is a linear combination of the mean of the embeddings of $v_i$'s neighbours and $v_i$'s own embedding at the previous layer, $k-1$. Predictions can be made at each vertex $v_i$ by using the final compute embedding, $h_{v_i}^{(K)}$. We refer the reader to \cite{kipf2016semi,zhang2019graph} for a detailed explanation of GCNs. Some works that rely on graph convolutions to generate embeddings are MoNet \cite{monti2017geometric}, Feastnet \cite{verma2018feastnet}, DeepInf \cite{qiu2018deepinf}, and PinSage \cite{ying2018graph}.

\subsubsection{Model interpretability}\label{sec:model_interpretability}
With the proliferation of deep learning and its use across various applications in society, trust has become a central issue. Given the black-box nature of these deep learning systems, there is a strong desire to understand the reasons behind their decisions. This has led to the sub-field of explainable AI (XAI)\cite{gunning2017explainable}. According to the works in \cite{lipton2018mythos,ras2022explainable}, explaining a model can (i) \textit{give insight into the model's training and generalisation} or (ii) \textit{give insight into the model's predictions}. The first category encompasses models that are inherently transparent - e.g., decision trees - and can provide indications of the path the model took to make its decision. However, the majority of interpretability strategies in deep learning fall into the second category: i.e., post-hoc interpretability. As suggested, post-hoc interpretability methods refer to the explanation of the outcome after the model has finished its training. Generally, models that are not intrinsically interpretable are denoted as black-boxes. Post-hoc interpretability 
engages in opening the black-box and explaining to domain laymen the reasons behind the predictions. Additionally, they can be exploited to reveal hidden patterns in the overall model behaviour. This type of interpretability approach can be broken into more specific categories: i.e., counterfactual and feature-based explanations.
To support the discussion of counterfactual explainability in graphs - see Section 2 in the manuscript - we provide the reader with an overall description of a \textit{counterfactual}. A counterfactual explanation for a prediction highlights the smallest change to the feature values that changes the prediction to a predefined output \cite{molnar2020interpretable}. Here, we rely on the counterfactual formalisation provided in \cite{guidotti2022counterfactual} (see Definition \ref{def:general_counterfactual}). 

\begin{definition}\label{def:general_counterfactual}
Given a classifier $\Phi$ that outputs the decision $\Phi(x)$ of the instance $x$, a counterfactual explanation consists of an instance $x'$ such that the decision $\Phi(x') \neq \Phi(x)$ and such that the difference between $x$ and $x'$ is minimal. 
\end{definition}

In this context, we assume that the classifier $\Phi$ - also called \textit{Oracle} - is a black-box model and the instances $x$ and $x'$ consist of a set of features $\mathcal{F}_x$ and $\mathcal{F}_{x'}$, respectively. Thus, counterfactual explanations belong to the family of example-based explanations \cite{aamodt1994case}. Differently from counterfactuals, other example-based explanations \cite{molnar2020interpretable} do not provide any insight into the way an instance $x$ needs to mutate itself in order to change its prediction $\Phi(x)$. For example, let us consider that the customer $x$ requests a bank loan which the intelligent system $\Phi$ rejects. A counterfactual explanation can reveal that a hypothetical customer $x'$ would have the loan accepted, where $x'$ is identical to $x$ but with a yearly income of $\$50,000$ instead of $\$45,000$ and without any debt left. The hypothetical customer $x'$ is a counterfactual example, and the counterfactual explanation consists of the income of $\$50,000$. Thus, according to \cite{guidotti2022counterfactual}, a counterfactual explanation is a set of counterfactual examples: i.e., $\{x_1',\dots,x_h'\}$. Moreover, a counterfactual explainer is defined as follows:

\begin{definition}
A counterfactual explainer is a function $f_k$ that takes in input a classifier $\Phi$, a set $X$ of known instances, and a given instance of interest $x$. Hence, $f_k(x,\Phi, X)$ returns a set $\{x_1',\dots,x_h'\}$ (s.t. $h \leq k$) of valid counterfactual examples where $k$ is the number of counterfactuals required.
\end{definition}

\subsection{Factual vs counterfactual explanations:}\label{sub:factual_counter} Differently from counterfactual explanations, there are other techniques that base their explainability on the answer to the following question: \textit{what were the dominant features that contributed to the outcome $\Phi(x)$ produced on the examined instance $x$?} These techniques are usually dubbed as \textit{factual} or \textit{attribute-based} explanations \cite{yuan2021explainability}. In this scenario, the explanation quantifies the impact of each feature on the outcome. Taking into consideration the previous example, the factual explanation for the loan being rejected could be that the yearly income is low and the debt left prior to the loan is high. Hence, income and debt are the features that mostly contributed to the rejection of the requested loan. Like counterfactual, most factual explainers also rely on comparing the examined instance $x$ to one or more counterfactual inputs - referred as baselines. However, the role of counterfactuals here is to break apart the relative importance of features rather than to identify new instances with favorable predictions. For instance, SHAP \cite{lundberg2017unified} operates by considering counterfactuals that eliminate features and note the marginal effect on the prediction. In other words, we note the change in the outcome when a feature is eliminated. We repeat the process for all combinations of features while computing a weighted average of the marginal effect on $\Phi(x)$.

\section{Datasets adopted in the literature}\label{sec:datasets_app}
\subsection{Synthetic datasets}
\begin{figure}[ht]
  \centering
  \includegraphics[width=0.5\textwidth]{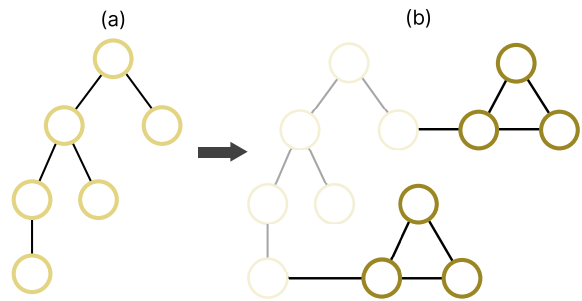}
  \caption{Example of Tree-Cycles dataset instances. Instance (a) belongs to class 0; meanwhile, instance (b) belongs to class 1.}
  \label{fig:tree-cycles}
\end{figure}

The following are datasets containing synthetically generated graphs:

\noindent
\textit{Tree-Cycles} \cite{ying2019gnnexplainer} has an 8-level balanced binary tree base structure, to which 80 six-node cycle motifs are added and attached to random vertices. This method was generalised in \cite{prado2022gretel}, where each instance has a random tree as a base structure, cycle motifs are added to each instance, and the resulting graph is binary-classified based on the presence of a cycle. The user can control the number of instances, the number of vertices per instance, and the number of connecting edges. Figure \ref{fig:tree-cycles} shows a toy example from this dataset where the cycle motif added to the tree is attached to a single vertex to prevent creating additional cycles. This method is similar to what \cite{ying2019gnnexplainer} proposed but with the addition of minimal counterfactual explainability.

\noindent
\textit{Tree-Grid} \cite{ying2019gnnexplainer} is a vertex classification dataset that follows the same idea as Tree-Cycles, except that 3-by-3 grid motifs are attached to the base tree graph in place of the cycle motif.

\noindent
\textit{Tree-Infinity} \cite{prado2022gretel}  follows the same approach as the Tree-Cycles dataset. However, instead of incorporating cycles, it features an infinity symbol attached to the primary graph. 

The motivation behind creating the Tree-Infinity dataset is to overcome the issue in Tree-Cycles, where oracles learn to count vertices and determine if the input is cyclic.

\noindent\textit{BA-Shapes} \cite{ying2019gnnexplainer} is a vertex classification dataset consisting of a base Barabási-Albert (BA) graph with 300 nodes and 80 house-structured motifs attached to random nodes. The resulting graph is further modified with random edges. The motifs have nodes in the top, middle, and bottom of the house or outside of it.

\noindent\textit{BA-2motifs} \cite{luo2020parameterized} is a graph classification dataset comprising 800 BA graphs with either "house" or five-node cycle motifs attached. The graphs are binary classified based on their attached motifs.

\noindent\textit{BA-Community} \cite{ying2019gnnexplainer} is a vertex classification dataset that combines two BA-Shapes graphs. Vertex features are sampled from two Gaussian distributions, one for each BA-Shapes graph. Vertices are classified into 8 classes based on their community memberships and structural roles.

\subsection{Real world datasets}
Besides the synthetic datasets provided above, the literature relies on real datasets. The real datasets can be divided into three domains: i.e., -omics, molecular, and social networks.

\noindent
\textit{Autism Spectrum Disorder} (\textit{ASD}) \cite{abrate2021counterfactual} is a graph classification dataset focused on children below nine years of age. The dataset includes 49 individuals with ASD and 52 Typically Developed (TD) individuals as the control group. The data is obtained using functional magnetic resonance imaging (fMRI), where vertices represent brain Regions of Interest (ROI) and edges represent co-activation between two ROIs.

\noindent\textit{Attention Deficit Hyperactivity Disorder} (\textit{ADHD}) \cite{abrate2021counterfactual} is a graph classification dataset from USC Multimodal Connectivity Database (USCD) \cite{brown2012ucla}, obtained using fMRI. The dataset includes 190 individuals with ADHD and 330 Typically Developed (TD) individuals as the control group. Each instance is a graph with vertices representing brain ROIs and edges representing co-activation between two regions.

\noindent\textit{Blood-Brain Barrier Permeation} (\textit{BBBP}) \cite{wellawatte2022model} is a molecular graph classification dataset for predicting blood-brain barrier permeation. The dataset comprises 2053 molecules selected from publications discussing BBB penetration, of which only 1970 were used for modeling. The excluded compounds exceeded a molecular weight of $600 Da$. The dataset has 1570 instances capable of crossing the Blood-Brain Barrier and 483 instances that cannot.

\noindent\textit{HIV Activity Prediction} \cite{wellawatte2022model} is a molecular graph classification dataset based on the ability of compounds to inhibit HIV. The Drug Therapeutics Program (DTP) prepared the dataset for AIDS antiviral screening, containing over 40,000 compounds.

\noindent\textit{Ogbg-molhiv} \cite{hu2020open} is a molecular graph classification dataset with 41,127 instances. Here, each vertex represents an atom, and each edge is a chemical bond.

\noindent\textit{Mutagenicity} \cite{kazius2005derivation} is a molecular graph classification dataset that identifies toxic substructures contributing to mutagenicity, a property that reduces a compound's potential as a marketable drug. The dataset contains 4,337 molecules classified as mutagenic or non-mutagenic based on the presence of these substructures. On average, each instance contains 30.32 vertices and 30.77 edges.

\noindent\textit{NCI1} \cite{wale2008comparison} is a molecular dataset for graph classification, containing bio-assay data of 4,110 chemical compounds on rodents for studying their carcinogenicity properties. The compounds are categorised as positive or negative for cell lung cancer. The average vertices and edges per instance are 29.87 and 32.30, respectively.

\noindent\textit{TOX21} \cite{kersting2016benchmark} is a molecular dataset collection for graph classification that addresses the issue of chemical toxicity. The dataset contains over 10,000 compounds, including environmental chemicals and approved drugs, divided into multiple datasets corresponding to specific assays. Each instance is binary classified into toxic or non-toxic, providing a means to evaluate environmental chemicals and develop new medicines.

\noindent\textit{ESOL} \cite{wu2018moleculenet} is a small molecular dataset for graph regression tasks. It contains water solubility data for 1,128 compounds represented by SMILES strings, which encode chemical structures without 3D coordinates. Models trained on this dataset estimate solubility directly from these structures.

\noindent\textit{Proteins} \cite{borgwardt2005protein} is a molecular dataset composed by $1113$ graph instances with $39.06$ nodes and $72.82$ edges in average. Graphs in the dataset are classified into enzymes and non-enzymes.

\noindent\textit{Davis} \cite{davis2011comprehensive} contains the drug-target binding affinity of 442 target proteins and 72 drugs. Given a drug-target pair, the smaller the kinase disassociation constant bioactivity, the higher the interaction affinity between the chemical compound and the protein kinase.

\noindent\textit{PDBBind} \cite{wang2004pdbbind} aims to provide a comprehensive collection of experimentally measured binding affinity data for all biomolecular complexes deposited in the Protein Data Bank (PDB). The current release (2020 version) provides binding affinity data for a total of $23,496$ biomolecular complexes in PDB, including protein-ligand ($19,443$), protein-protein ($2,852$), protein-nucleic acid ($1,052$), and nucleic acid-ligand complexes ($149$).

\noindent\textit{CiteSeer} \cite{giles1998citeseer} is a citation network for vertex classification with 3,312 scientific publications classified into six classes and 4,732 links indicating citation relationships. Each publication is a binary word vector of 3,703 unique words.

\noindent\textit{IMDB-M} \cite{yanardag2015deep} contains film collaboration networks. A vertex represents an actor or actress, and an edge connects two vertices when they appear in the same movie.

\noindent\textit{CORA} \cite{mccallum2000automating} consists of 2708 scientific publications classified into one of seven classes. The citation network consists of 5429 links. Each publication in the dataset is described by a binary word vector indicating the absence/presence of the corresponding word from the dictionary. The dictionary consists of 1433 unique words.

\noindent\textit{Musae-Facebook} \cite{rozemberczki2019multiscale} is a page-page graph of verified Facebook sites. Nodes represent official Facebook pages while the links are mutual likes between sites. Node features are extracted from the site descriptions that the page owners created to summarise the purpose of the site. This graph was collected through the Facebook Graph API in November 2017 and restricted to pages from 4 categories that are defined by Facebook. These categories are politicians, governmental organizations, television shows, and companies. The task related to this dataset is multi-class node classification for the 4 site categories.

\noindent \textit{LastFM} \cite{rozemberczki2020characteristic} is a social network of users from Asian (e.g., the Philippines, Malaysia, Singapore) countries. Vertices represent users of the music streaming service LastFM and links among them are friendships. The dataset is collected in March 2020 via available API calls. The classification task related to this dataset is to predict the home country of users given the social network and artists liked by them.

\noindent \textit{Yelp} \cite{wang2019neural} is adopted from the 2018 edition of the Yelp challenge. It contains user ratings and comments for local businesses such as restaurants and bars, viewed as items. Hence, one can build a bipartite graph of user and item vertices, where the edges represent a review or a rating that a particular user has posted for a specific item.

\section{Evaluation tools for GCE}

Having a systematic evaluation mechanism for explainability methods is of paramount importance for further development and extension of the GCE research field. Evaluation tools help researchers standardise their experiments and make it feasible and easier to compare their proposed methods with the state-of-the-art leading to more consolidated claims regarding their predictions/explanations. Most of the methods enlisted in Section 3.3 rely only on some of the metrics in Section 4.2. Nevertheless, they tend to perform their experiments on different datasets, thus hindering the replicability of their performances on other data. The surveys presented in Table 1 of the main material try to propose reproducible evaluation frameworks, respectively, SACE \cite{guidotti2022counterfactual} and Extensible Counterfactual Explainers (ECE) \cite{artelt2019computation}. Notice that Verma et al. \cite{verma2020counterfactual} abstain from providing an open-source framework to solidify their contribution to the literature.

In GCE, GRETEL \cite{prado2022gretel,prado2023developing} is the only dedicated tool that bridges the gap of the lack of extensibility and reproducibility of the evaluation frameworks proposed in the literature. Besides briefly describing GRETEL's components in Sec. \ref{sec:gretel_components}, we discuss its main advantages according to a reproducible and extensible point-of-view (see Sec. \ref{sec:gretel_advantages}).

\subsection{GRETEL: Design principles and core components}\label{sec:gretel_components}

\begin{figure}[!t]
  \centering
  \includegraphics[width=\textwidth]{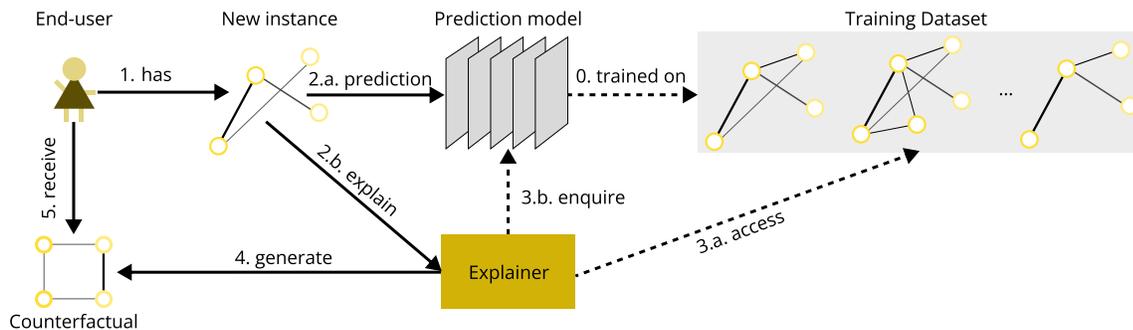}
  \caption{Typical workflow for the counterfactual explanation of an instance classified by a black-box model. The steps (lines) are ordered according to the reported numbers. The prediction model could be trained on a training dataset (step 0). The dashed lines depict that the steps are optional whether the explainer is data-agnostic (step 3.a) and model-agnostic (step 3.b).}
   \label{fig:exai-workflow}
\end{figure}

As mentioned above, GRETEL\footnote{\url{https://github.com/MarioTheOne/GRETEL}} is the only available framework that is dedicated to the evaluation of GCE methods. Here, we discuss its design principles and core components. According to \cite{prado2022gretel}, the evaluation of a GCE follows the workflow depicted in Figure \ref{fig:exai-workflow}. For visualisation purposes, we illustrate a classification scenario, but the setup is general for any learning task. Here, we illustrate the explanation of a new graph instance given by an end-user. Notice that, for the correct function of the explainer, we assume that the black-box prediction model has been trained on a dataset of graph data (step 0). The explainer - based on a particular post-hoc explainability method - can access the training data\footnote{In privacy-preservation scenarios, the explainer does not have access to the raw training data.} (step 3.a) and enquire the prediction model over the outcome for the new instance (step 3.b). The new instance is passed to the prediction model (step 2.a) and the explainer (step 2.b), which generates an explanation (step 4) to pass to the user (step 5).

GRETEL follows an intuitive workflow and aids future researchers in performing exhaustive sets of evaluations. It adopts an object-oriented approach where abstract classes constitute the core of the framework. Hence, GRETEL comes with a built-in factory design pattern that helps future researchers leverage its intrinsic extensibility into forming their ad-hoc explainability scenarios. By default, GRETEL comes with its configurations, already-trained models, and the used datasets. These add-ons to the original repository allow the generation of a dataset or train a model on the fly for reproducibility purposes, hence, mitigating the efforts needed to evaluate new settings.

The GRETEL's main components, according to the original paper, are:
\begin{itemize}
    \item The \textbf{Oracle} depicts a generic interface that provides extensive interactions with the employed prediction models. Besides providing basic logic to keep track of the number of calls made to the oracle, it provides the possibility to save/load the model on the fly.
    \item The \textbf{Explainer} represents the base class extended by ad-hoc explanation methods. Given an oracle $\Phi$ and an instance $G$, it provides its counterfactual explanation $\mathcal{E}_\Phi(G)$. For further experiments, the authors suggest the reader to extend this class and write their implementation accordingly.
    \item The \textbf{EvaluationMetric} is an abstract class useful to define specific evaluation metrics for the quality of the explainer. Again, researchers can implement their metrics and use the ones provided by default within the framework's scope.
    \item The \textbf{Evaluator} is responsible for carrying out the evaluation of a specific explainer. Given an explainer $\Xi$, an oracle $\Phi$, a set of graph instances $\mathcal{G}$, and a set of evaluation metrics $\mathcal{M}$, it collects all the performances of $\Xi$ for each instance $G \in \mathcal{G}$ for all metrics $M \in \mathcal{M}$ according to the predictions $\Phi(G)$.
\end{itemize}
The components mentioned above constitute only part of GRETEL's ecosystem. Therefore, we refer the reader to \cite{prado2022gretel} for a complete overview of the framework. Moreover, notice that GRETEL
can be extended in several ways, but the three fundamental use cases are the following:
\begin{itemize}
    \item a data scientist wants to evaluate a brand new Explainer on an existing Datasets using an established set of Metrics;
    \item a company wants to evaluate, using existing Metrics, which is the best Explainer to adopt on its private Datasets, thus in its business scenario;
    \item a data scientist or a company wants to test their new Metric on the existing Explainers and Datasets to capture unexplored aspects of the quality of the explanations;
\end{itemize}
\noindent We point the reader to \cite{prado2023developing} for a guide on how to use GRETEL encompassing the previously listed use-cases.

\subsection{Reproducibility and beyond with GRETEL}\label{sec:gretel_advantages}

\begin{table}[!t]
\centering
\caption{Relationship between the data used and the analysis (including predictions and evaluations) performed in the experimental process \cite{national2019reproducibility}. Notice that reproducibility lies on the similar-data-similar-analysis category.}
\label{tab:reproducible_vs_replicable}
\resizebox{.5\textwidth}{!}{%
\begin{tabular}{@{}ll|cc@{}}
\toprule
\multicolumn{2}{l}{\multirow{2}{*}{}}                 & \multicolumn{2}{c}{Data} \\ \cmidrule(l){3-4} 
\multicolumn{2}{l}{} & Similar  & Different     \\ \midrule
\multicolumn{1}{c}{\multirow{2}{*}{Analysis}} & \multicolumn{1}{l}{Similar} & Reproducible & Replicable \\ \cmidrule(l){2-4} 
\multicolumn{1}{c}{} & \multicolumn{1}{l}{Different} & Robust   & Generalisable \\ \bottomrule
\end{tabular}%
}
\end{table}
One of the challenges for research in machine learning is to ensure that the proposed methods and their corresponding results are sound and reliable. Specifying hypotheses, running experiments, analysing results, and drawing conclusions is the foundational process of scientific inquiry. For findings to be valid, it is essential that the experiments be repeatable and yield the results and conclusions as originally proposed \cite{pineau2021improving}. While experiments in fields like biology, physics, or sociology are made in the natural/social world, experiments in computer science are computer-designed and human-built, entailing a far easier reproducibility task than the other fields enlisted. Yet, researchers have difficulties reproducing the work of others \cite{henderson2018deep}. In particular, 70\% of researchers failed to reproduce another paper's results in the journal Nature \cite{baker20161}. According to Pineau et al.\cite{pineau2021improving}, the reasons behind the gap of reproducible experiments in machine learning, where experiments are used to train and make predictions on observed data, are the following:
\begin{itemize}
    \item Impossibility to access the training dataset with the same data distribution as in the original papers.
    \item Shortage of details in explaining the model or the training procedure.
    \item Unpublished code necessary to run the experiments. In some cases, the published code can also contain errors hindering the reproducibility of the experimental process in the original paper.
    \item Under-reporting of the evaluation metrics used to assess the performances of the prediction model.
    \item Improper usage of significance statistical tests to analyse results.
    \item Selective reporting and avoidance of adaptive overfitting.
    \item Overclaiming of results leads to wrong conclusions beyond the evidence presented.
\end{itemize}
Besides tackling all the above challenges for reproducibility in the literature, GRETEL provides a systematic way of using already established benchmarks and methods in graph learning. Moreover, it exploits the philosophy of object-oriented software engineering that allows users to extend certain components of the framework to contribute to the Open Science movement \cite{vicente2018open} for future research.

Reproducibility encompasses the re-execution of an experiment using the same data and the same analytical tools, including the prediction model and the same evaluation framework \cite{tatman2018practical}. We invite the reader to notice the subtle difference between reproducibility and replicability \cite{national2019reproducibility}. In detail, replicability entails obtaining consistent results across studies aimed at answering the same scientific question, each with its own data. In other words, reproducibility involves the original data and code, whereas replicability involves new data collections and similar prediction methods used by previous studies. Inspired by \cite{national2019reproducibility,pineau2021improving}, in Table \ref{tab:reproducible_vs_replicable}, we represent the relationship between the data used and the analysis performed on it. The analysis reported in the table includes the prediction model and the evaluation framework adopted. 
Although we are interested in reproducibility, we argue that GRETEL satisfies all the criteria depicted in Table \ref{tab:reproducible_vs_replicable}.
As an initial commit of the repository, GRETEL contains the implementation of some counterfactual explainers presented in Sec. \ref{sec:methods}, the majority of the datasets of Sec. \ref{sec:datasets_app}, and the majority of the metrics of Sec. \ref{sec:evaluation_metrics}. We invite the reader to note that GRETEL is periodically maintained and updated with new state-of-the-art methods/datasets/metrics. Hence, future researchers can use the explainers and evaluation benchmarks to reproduce the experiments presented by the authors in \cite{prado2022gretel}. To promote the reproducibility of these experiments, GRETEL permits and makes available the saved models and provides an on-the-fly load of their weights which can be used to make the same predictions as made in the original paper (\textbf{reproducibility}).

In particular, GRETEL follows a naturally extensible object-oriented philosophy divided into components that work as standalone modules. To this end, GRETEL permits the end-user to provide their implementations of all of its modules, giving sprout to a plug-and-play scenario. Moreover, besides enlisting benchmarking datasets, the framework allows the generation of synthetic graph data based on specific patterns (e.g., cycle creation, random trees, infinity-shaped graphs). This feature aids in replicating the results according to the employed metrics while maintaining the same oracles and explainers (\textbf{replicability}). 

GRETEL gives the user the possibility to extend the base class of oracles and explainers while maintaining the same datasets and evaluation metrics, thus satisfying the \textbf{robustness} relationship between data and analysis. Finally, researchers can use GRETEL in terms of boilerplate code generators. In other words, they can extend the base classes of oracles and explainers for their ad-hoc scenario and generate synthetic datasets or load their own to perform original experiments (\textbf{generalisability}). More specifically, according to the dimensions in Table \ref{tab:reproducible_vs_replicable}, GRETEL provides the following advantages:
\begin{itemize}
    \item \textit{Reproducibility} - GRETEL comes with configuration files that allow the end-user to rerun the experiments provided in the original paper.
    \item \textit{Replicability} - In the configuration file, the user can add new paths pointing to other datasets and run the same experiments as in the paper. Additionally, the user can generate synthetic datasets and evaluate the performances of the explainers on them.
    \item \textit{Robustness} - Besides permitting the extension of explainers/oracles, GRETEL decouples the functioning of its components from hard-coded parameters. The configuration file takes care of the correct parametrisation of the various components. Furthermore, the parameter combination produces different explainers/oracles while maintaining the same code as provided by default.
    \item \textit{Generalisability} - Notice that the entire framework is extensible and personalisable to the user's explanation scenario. To aid the Open Science movement, GRETEL provides the community with Jupyter Notebook tutorials on how to extend each of its components. Finally, to shrink training time consumption, GRETEL is capable of loading model weights on-the-fly if they are provided within the project's workspace, thus, enabling the community to verify the performances reported in the original work.
\end{itemize}

Taking into consideration the observations made above, GRETEL is the only GCE framework in the explainability literature that satisfies all the categories in Table \ref{tab:reproducible_vs_replicable} following compartmentalised and modular software engineering design patterns. We invite the reader to take extra care with specifying the paths to their model weight files and dataset files in their file system when trying to reproduce the experiments provided in GRETEL's online repository. Therefore, absolute path files are used instead of relative ones, which is a clear disadvantage when changing machines or even operative systems\footnote{Absolute paths require maintenance when changing the directory of a specific file. They also require the maintainer to double-check whether the home directory has changed or not. Additionally, when changing operative systems, path separators might be different, which can lead to further intricacies and tediousness in specifying them every time the running environment changes.}. Nevertheless, the configuration files provided help avoid hard-coding the parameters of the explainers and oracles. In this way, GRETEL has the advantage to specify different parameters (e.g., graph edit distance as the minimality evaluation of the explainer). Additionally, the configuration file gives the user the possibility to specify also the strategy of generating synthetic datasets and the embedding mechanism used by the oracle. Among other advantages, this highly tailored feature of customising the experiments into bespoke evaluation scenarios provides the framework with an enhanced robust feature for future usage. Finally, the authors include tutorials\footnote{\url{https://github.com/MarioTheOne/GRETEL/blob/main/examples/tutorial.mp4?raw=true}} on how to extend each of the framework's components and provide evaluations on toy datasets with on-the-fly loaded model weights, hence contributing even more to the Open Science movement described above.
\begin{table}[!ht]
\centering
\caption{The hyperparameters chosen for each SoA method compared in the main manuscript.}
\label{tab:hyperparamters}
\resizebox{\textwidth}{!}{%
\begin{tabular}{@{}llll@{}}
\toprule
      & Tree-Cycles                      & ASD                              & BBBP                             \\ \midrule
OBS   & $\mathcal{D}_{inst}(G,G')$ = GED & $\mathcal{D}_{inst}(G,G')$ = GED & $\mathcal{D}_{inst}(G,G')$ = GED \\ \midrule
DDBS  & $\mathcal{D}_{inst}(G,G')$ = GED & $\mathcal{D}_{inst}(G,G')$ = GED & $\mathcal{D}_{inst}(G,G')$ = GED \\ \midrule
MACCS & $\mathcal{D}_{inst}(G,G')$ = GED & $\mathcal{D}_{inst}(G,G')$ = GED & $\mathcal{D}_{inst}(G,G')$ = GED \\ \midrule
CLEAR &
  \begin{tabular}[c]{@{}l@{}}$|V| = 32$\\ classes = 2\\ epochs = 600\\ h\_dim = 16\\ z\_dim = 16\\ dropout = 0.1\\ disable\_u = False\\ learning rate = 1e-3\\ weight decay = 1e-5\\ average graph pooling\\ $\alpha = 5$\\ $\beta_x = 10$\\ $\beta_{adj} = 10$\\ feature\_dim = 2\\ $\lambda_{sim} = 2$\\ $\lambda_{kl} = 1$\\ $\lambda_{cfe} = 1$\end{tabular} &
  \begin{tabular}[c]{@{}l@{}}$|V| = 116$\\ classes = 2\\ epochs = 600\\ h\_dim = 16\\ z\_dim = 16\\ dropout = 0.1\\ disable\_u = False\\ learning rate = 1e-3\\ weight decay = 1e-5\\ average pooling\\ $\alpha = 5$\\ $\beta_x = 10$\\ $\beta_{adj} = 10$\\ feature\_dim = 2\\ $\lambda_{sim} = 2$\\ $\lambda_{kl} = 1$\\ $\lambda_{cfe} = 1$\end{tabular} &
  \begin{tabular}[c]{@{}l@{}}$|V| = 269$\\ classes = 2\\ epochs = 5\\ h\_dim = 16\\ z\_dim = 16\\ dropout = 0.1\\ disable\_u = False\\ learning rate = 1e-3\\ weight decay = 1e-5\\ average graph pooling\\ $\alpha = 5$\\ $\beta_x = 10$\\ $\beta_{adj} = 10$\\ feature\_dim = 2\\ $\lambda_{sim} = 2$\\ $\lambda_{kl} = 1$\\ $\lambda_{cfe} = 1$\end{tabular} \\ \midrule
CF$^2$ &
  \begin{tabular}[c]{@{}l@{}}$|V| = 32$\\ $\alpha = 0.6$\\ $\lambda = 500$\\ epochs = 100\\ learning rate = 1e-3\\ weight decay = 1e-5\\ batch\_size = $25$\\ $\gamma = 0.5$\\ feature\_dim = 8\end{tabular} &
  \begin{tabular}[c]{@{}l@{}}$|V| = 116$\\ $\alpha = 0.6$\\ $\lambda = 500$\\ epochs = 100\\ learning rate = 1e-3\\ weight decay = 1e-5\\ batch\_size = $5$\\ $\gamma = 0.5$\\ feature\_dim = 8\end{tabular} &
  \begin{tabular}[c]{@{}l@{}}$|V| = 269$\\ $\alpha = 0.6$\\ $\lambda = 500$\\ epochs = 100\\ learning rate = 1e-3\\ weight decay = 1e-5\\ batch\_size = $101$\\ $\gamma = 0.5$\\ feature\_dim = 8\end{tabular} \\ \midrule
MEG &
  \begin{tabular}[c]{@{}l@{}}num\_input = 1024\\ epochs = 10\\ environment = Tree-Cycles\\ action encoder = Tree-Cycles\\ batch size = 1\\ learning rate = 1e-4\\ replay\_buffer\_size = 10000\\ steps\_per\_episode = 1\\ update\_intervals = 1\\ $\gamma = 0.95$\\ polyak = 0.995\\ num\_counterfactuals = 10\end{tabular} &
  \begin{tabular}[c]{@{}l@{}}num\_input = 1024\\ epochs = 10\\ environment = BBBP (fp\_len=1024, fp\_rad=2)\\ action encoder = MorganBitFingerprint (fp\_len=1024, fp\_rad=2)\\ batch size = 1\\ learning rate = 1e-4\\ replay\_buffer\_size = 10000\\ steps\_per\_episode = 1\\ update\_intervals = 1\\ $\gamma = 0.95$\\ polyak = 0.995\\ num\_counterfactuals = 10\end{tabular} &
  \begin{tabular}[c]{@{}l@{}}num\_input = 13456\\ epochs = 10\\ environment = ASD\\ action encoder = identity\\ batch size = 1\\ learning rate = 1e-4\\ replay\_buffer\_size = 10000\\ steps\_per\_episode = 1\\ update\_intervals = 1\\ $\gamma = 0.95$\\ polyak = 0.995\\ num\_counterfactuals = 10\end{tabular} \\ 
\bottomrule
\end{tabular}%
}
\end{table}

\section{Hyperparameter choice and evaluation details}
In the following Sec. \ref{sec:exclusion_reasons}, we provide the reader with the exclusion reasons for some methods from the empirical evaluation of Sec. \ref{sec:gretel_exp}. Even if we are far from reaching an ideal Open Science, we conducted the empirical evaluation comparing 10 different methods, including three newly introduced baselines (shown in Table \ref{tab:soa_performance}). Finally, in Sec. \ref{sec:hyperparameters}, we describe the hyperparameters used in the empirical evaluation (see Sec. \ref{sec:gretel_exp}).

\subsection{Reasons to exclude methods for evaluation}\label{sec:exclusion_reasons}
The methods of \cite{sun2021preserve,cai2022probability,liu2021multi} were not included in the empirical evaluation since they do not have a public code repository. CMGE \cite{wu2021counterfactual} has not been included, although it contains a public repository, since the authors obfuscate, probably for privacy or NDA reasons, the way they preprocess the original dataset. Notice that CMGE works on medical reports (free-text) and transforms them into graphs of entity-relationships. This transformation requires a transformer-based model that requires tokens generated from free-text. In Sec. \ref{sec:gretel_exp}, we evaluate graph datasets that do not require any kind of preprocessing. RCExplainer \cite{bajaj2021robust} has a free repository in Huawei’s marketplace. While integrating RCExplainer on GRETEL, we found out that the original code is a framework per se referring to libraries specifically developed within Huawei. These libraries are only downloadable as already-compiled, thus hindering us to effectively re-code RCExplainer in a reproducible fashion. Notice that RCExplainer does not work if it is decoupled from these libraries. MACDA \cite{nguyen2022explaining} is not included since its underlying oracle does drug-target prediction leading toward an unfair and misaligned comparison with other SoA methods. We also noticed that adapting MACDA to graph classification instead of drug-target predictions would be similar to MEG due to the reinforcement learning policy of generating counterfactuals. We leave this for further investigation. We do not include GCFExplainer \cite{huang2023global} since it does model-level explainability and would be misaligned with the other instance-level explanation methods. Finally, CF-GNNExplainer \cite{lucic2022cf} is not included since it can only work with vertex classification. Again, we emphasise that this survey is designed to provide the reader with a high level of performance of the SoA methods in graph classification. Testing all SoA methods in different classification tasks is left for future work.

\subsection{Hyperparameters}\label{sec:hyperparameters}
Here, we present the hyperparameters used for each of the SoA methods compared in Sec. \ref{sec:gretel_exp}. Notice that we do not perform any hyperparameter optimisation for the SoA and use default parameters - as described in the original papers - where applicable. Table \ref{tab:hyperparamters} shows the used hyperparameters for all the datasets evaluated in Sec. \ref{sec:gretel_exp} (i.e., Tree-Cycles, ASD, and BBBP). Notice that GRETEL's GitHub\footnote{\url{https://github.com/MarioTheOne/GRETEL}} contains the configuration files for each method that reproduces the experiments in Sec. \ref{sec:gretel_exp}.
\end{appendix}



\end{document}